\documentclass[10pt,twocolumn,letterpaper]{article}

\usepackage{pdfpages}
\usepackage{multido}

\usepackage{iccv}
\usepackage{times}
\usepackage{epsfig}
\usepackage{graphicx}
\usepackage{amsmath}
\usepackage{amssymb}
\usepackage{booktabs}
\usepackage{multirow}
\usepackage[table,dvipsnames]{xcolor}

\usepackage{overpic}
\usepackage{tikz}

\usepackage[pagebackref=true,breaklinks=true,colorlinks,bookmarks=false]{hyperref}

\iccvfinalcopy 

\def\iccvPaperID{9265} 

\newcommand{\notsosmall}{\fontsize{10.5pt}{12pt}\selectfont}

\ificcvfinal\fi

\begin{document}

\title{Learning Depth Estimation for Transparent and Mirror Surfaces}

\author{ 
    Alex Costanzino$^*$ \hspace{1.5cm} Pierluigi Zama Ramirez$^*$ \hspace{1.5cm} Matteo Poggi$^*$ \\ 
    Fabio Tosi \hspace{1.5cm} Stefano Mattoccia \hspace{1.5cm} Luigi Di Stefano \\
    \notsosmall CVLAB, Department of Computer Science and Engineering (DISI)\\
    \notsosmall University of Bologna, Italy\\
    {\tt\small \{alex.costanzino, pierluigi.zama, m.poggi, fabio.tosi5\}@unibo.it}
}

\maketitle
\def\thefootnote{*}\footnotetext{\emph{These authors contributed equally to this work}.}
\ificcvfinal\thispagestyle{empty}\fi

\begin{abstract}
   Inferring the depth of transparent or mirror (ToM) surfaces represents a hard challenge for either sensors, algorithms, or deep networks. We propose a simple pipeline for learning to estimate depth properly for such surfaces with neural networks, without requiring any ground-truth annotation. We unveil how to obtain reliable pseudo labels by in-painting ToM objects in images and processing them with a monocular depth estimation model. These labels can be used to fine-tune existing monocular or stereo networks, to let them learn how to deal with ToM surfaces. Experimental results on the Booster dataset show the dramatic improvements enabled by our remarkably simple proposal. 
\end{abstract}



\section{Introduction}
In our daily lives, we often interact with several objects of various appearances. Among them are those made of transparent or mirror surfaces (ToM), ranging from the glass windows of buildings to the reflective surfaces of cars and appliances. These might represent a hard challenge for an autonomous agent leveraging computer vision to operate in unknown environments. Specifically, among the many tasks involved in Spatial AI, accurately estimating depth information on these surfaces remains a challenging problem for both computer vision algorithms and deep networks \cite{zamaramirez2022booster}, yet necessary for proper interaction with the environment in robotic, autonomous navigation, picking, and other application fields. 
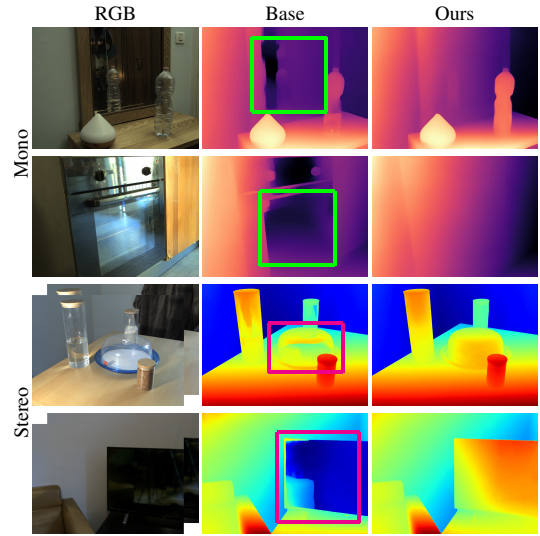
\begin{figure}[t]
    \centering
    \setlength{\tabcolsep}{1pt}
    \scalebox{0.85}{
    \begin{tabular}{cccc}
         & \small RGB & \small Base & \small Ours\\
         \multirow{2}{*}{\small \rotatebox{90}{Mono}} &
         \begin{overpic}[width=0.31\linewidth]{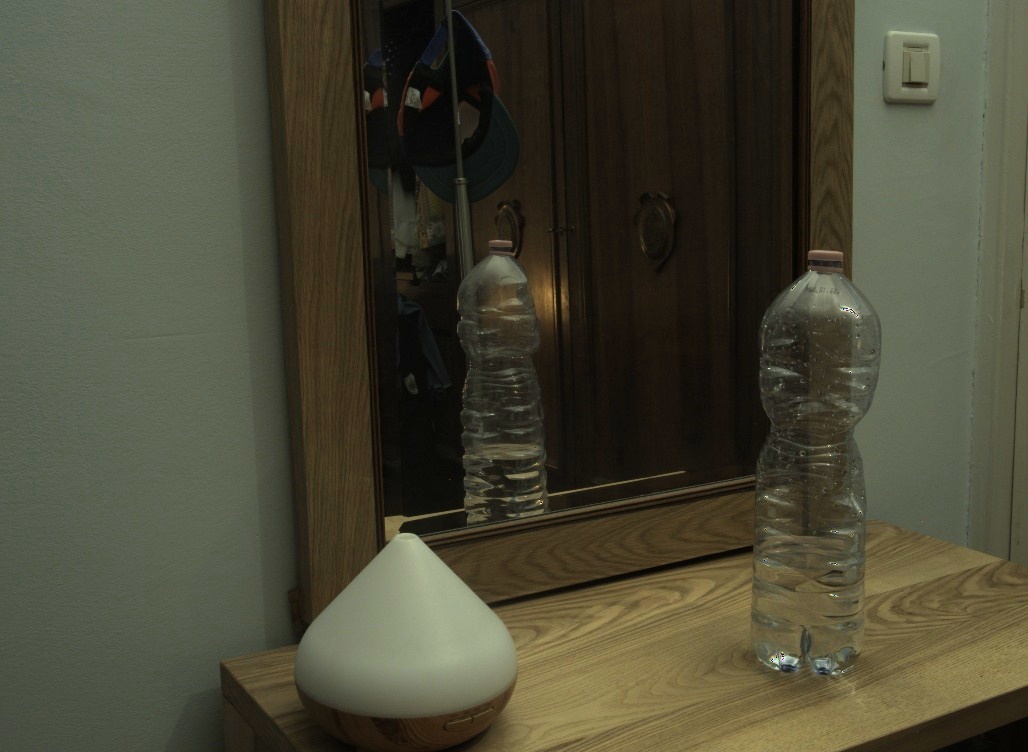}
         \end{overpic} 
         & 
         \linethickness{1.5pt}
         \begin{overpic}[width=0.31\linewidth]{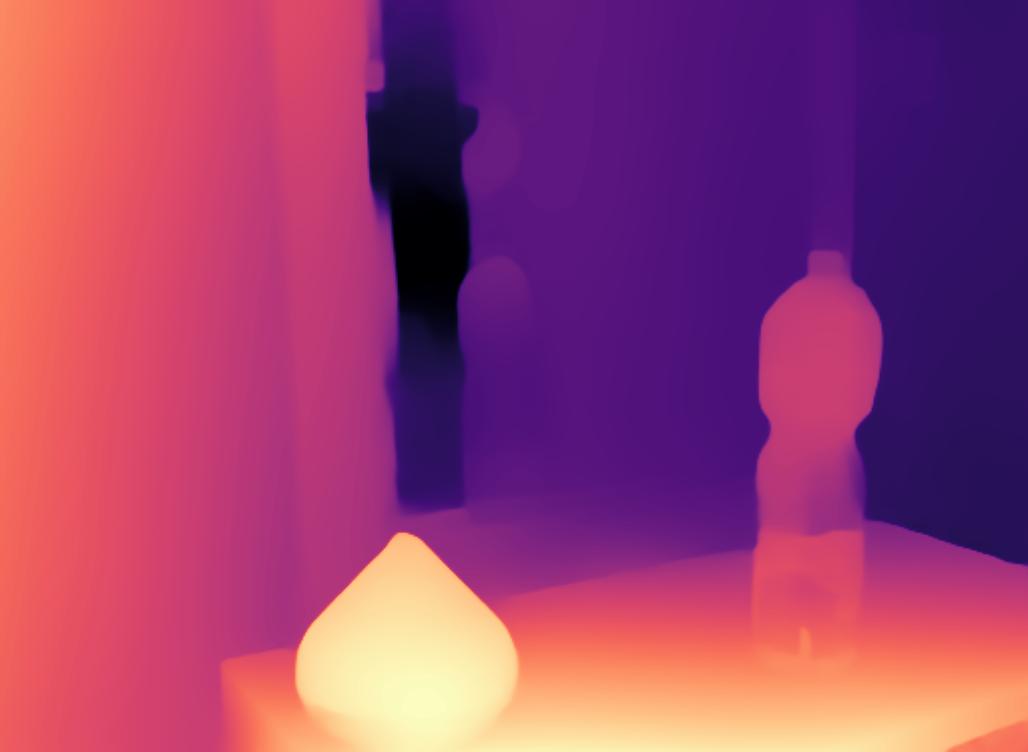}
         \put(30,22){\color{green}\line(0,2){45}}
         \put(30,67){\color{green}\line(2,0){45}}
         \put(75,22){\color{green}\line(0,2){45}}
         \put(30,22){\color{green}\line(2,0){45}}
         \end{overpic} &
         \begin{overpic}[width=0.31\linewidth]{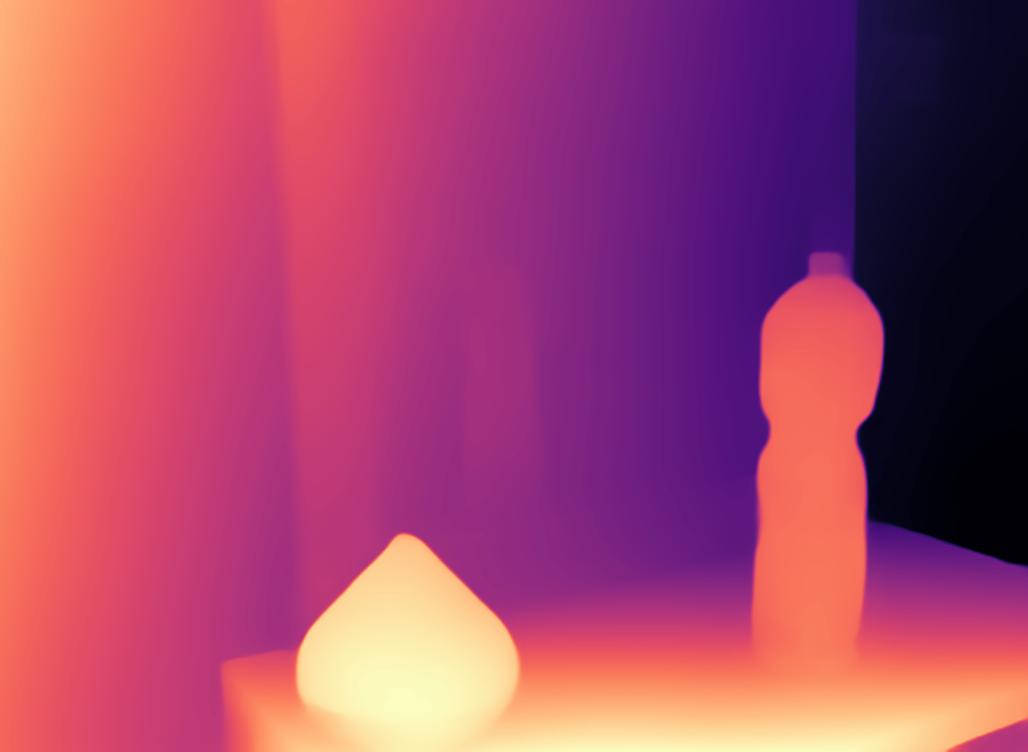}
         \end{overpic} \\
         & \begin{overpic}[width=0.31\linewidth]{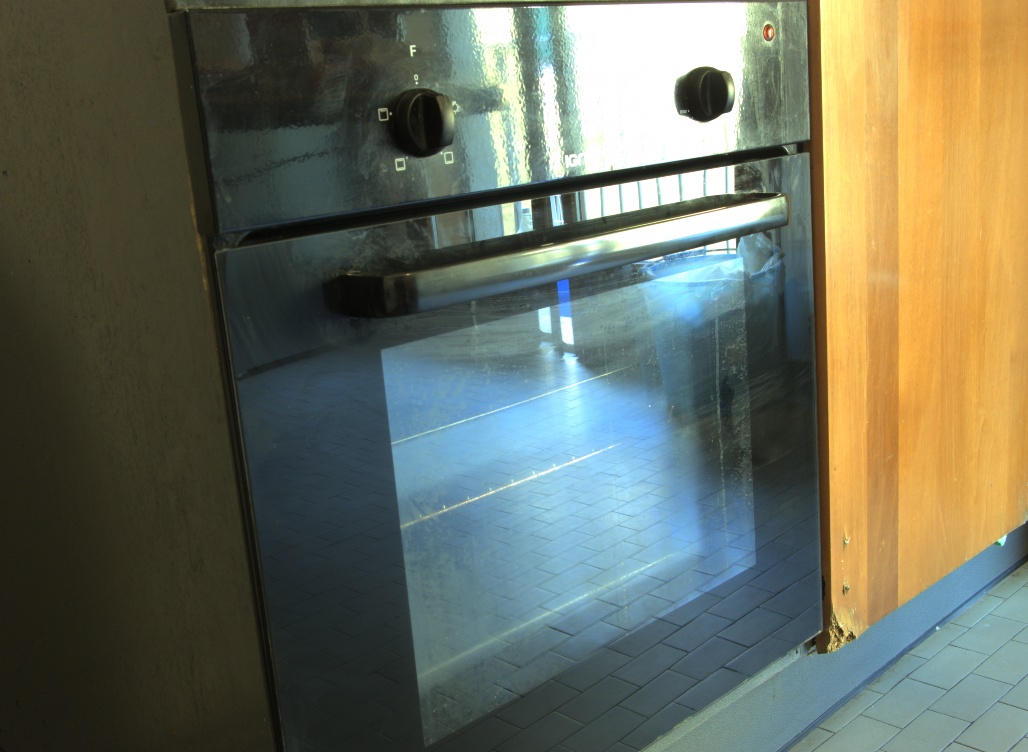}
         \end{overpic} &  
         \linethickness{1.5pt}
         \begin{overpic}[width=0.31\linewidth]{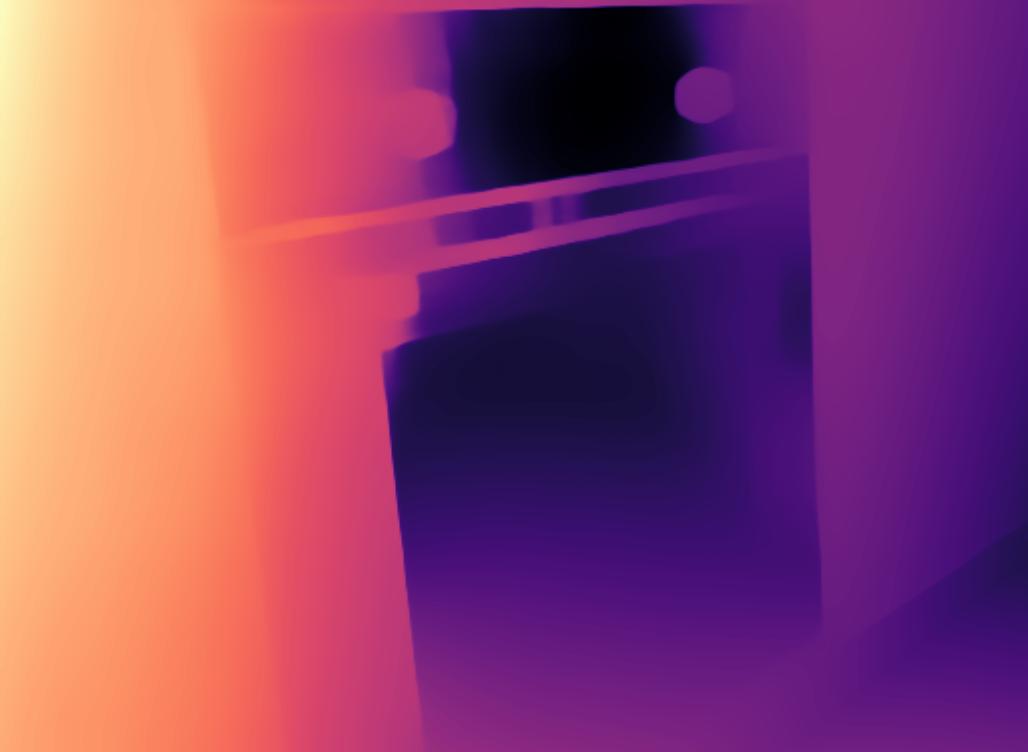}
         \put(35,7){\color{green}\line(0,2){45}}
         \put(35,52){\color{green}\line(2,0){45}}
         \put(80,7){\color{green}\line(0,2){45}}
         \put(35,7){\color{green}\line(2,0){45}}
         \end{overpic} & 
          \begin{overpic}[width=0.31\linewidth]{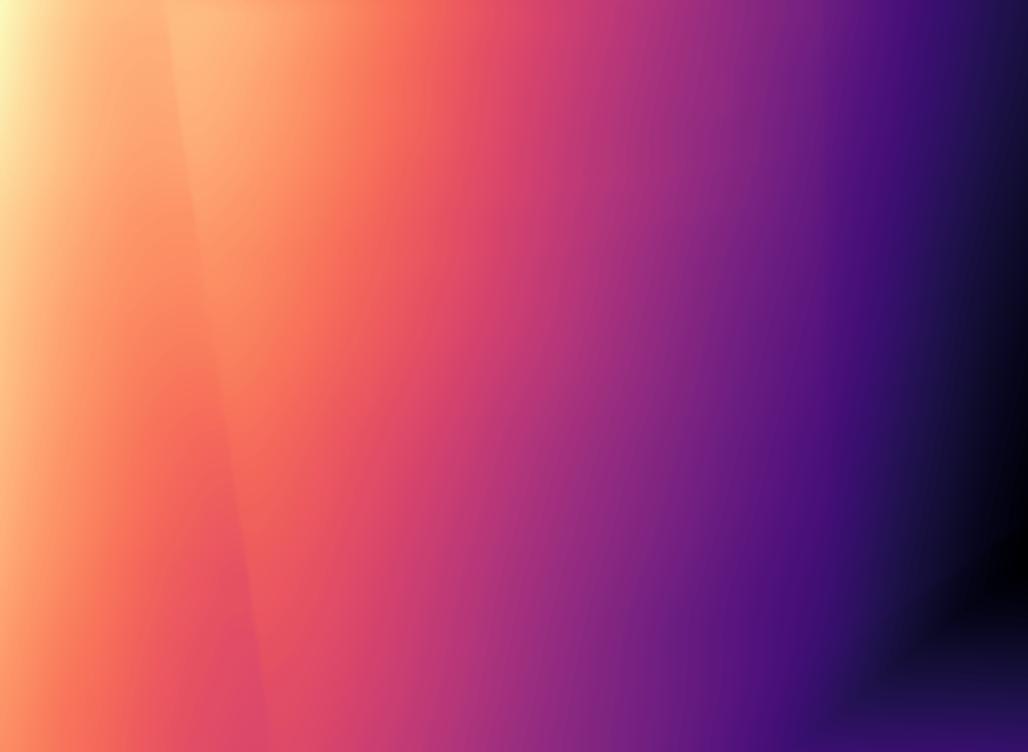}
          \end{overpic} \\
         \multirow{2}{*}{\small \rotatebox{90}{Stereo}} &
         \begin{overpic}[width=0.31\linewidth]{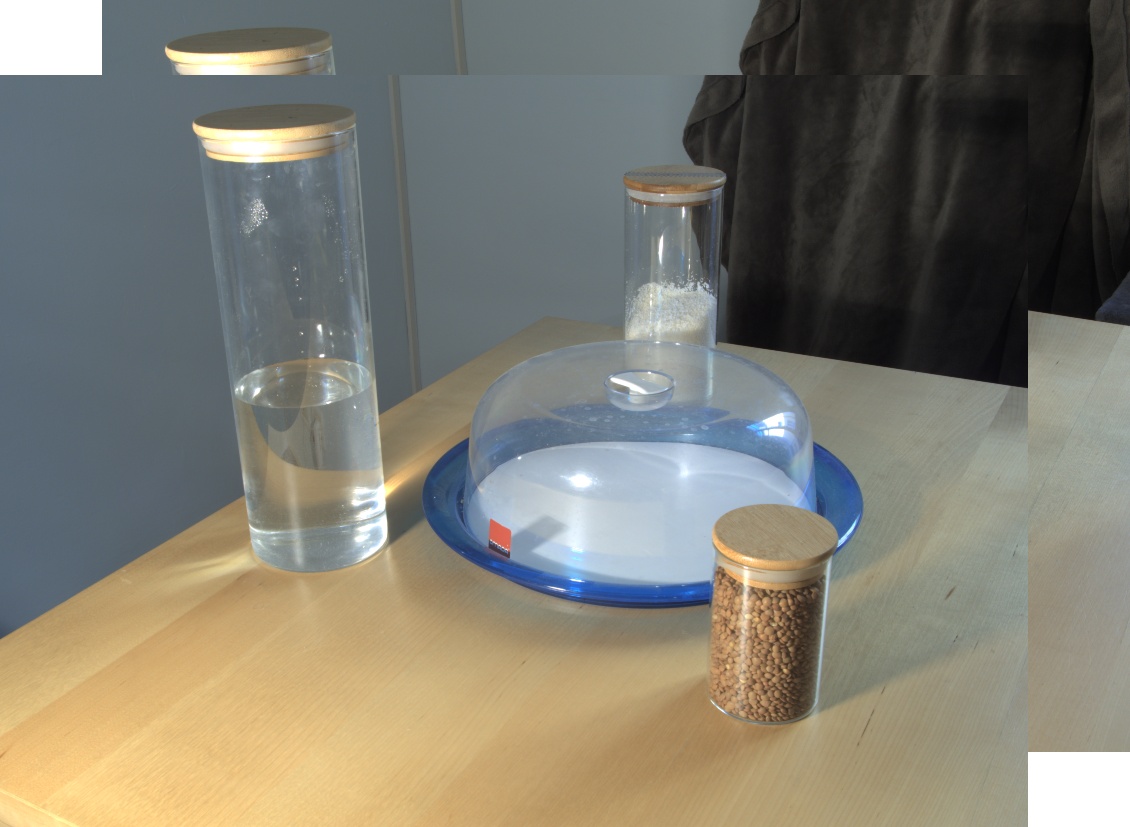}
         \end{overpic} &
         \linethickness{1.5pt}
         \begin{overpic}[width=0.31\linewidth]{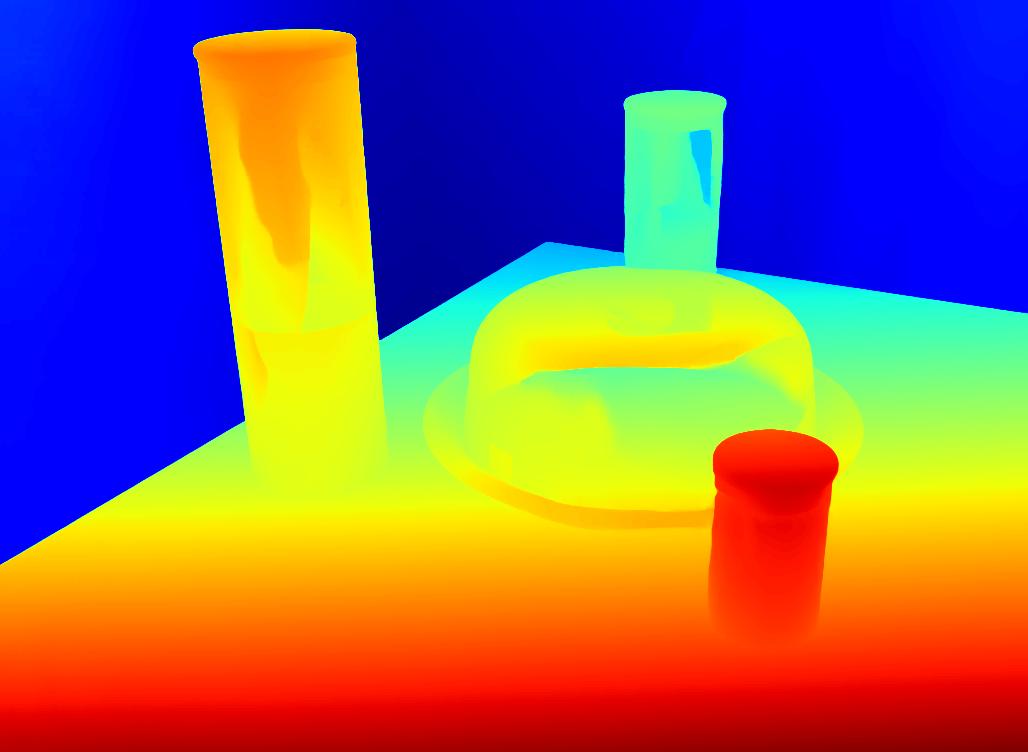}
         \put(40,20){\color{magenta}\line(0,2){30}}
         \put(40,50){\color{magenta}\line(2,0){45}}
         \put(85,20){\color{magenta}\line(0,2){30}}
         \put(40,20){\color{magenta}\line(2,0){45}}
         \end{overpic} &
         \begin{overpic}[width=0.31\linewidth]{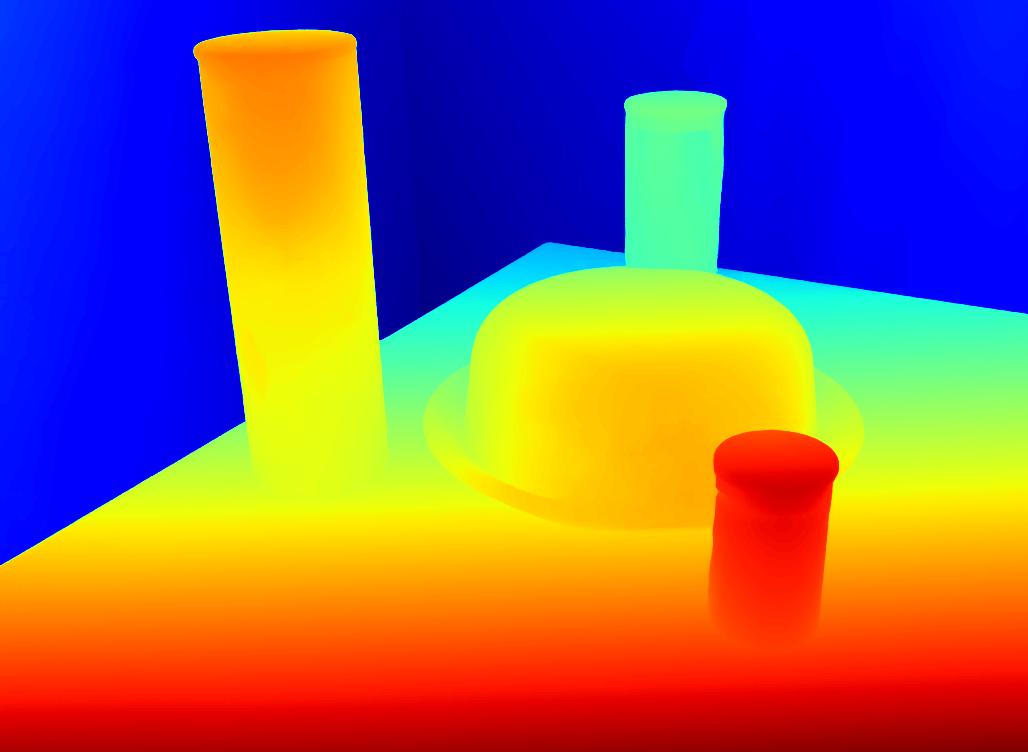}
         \end{overpic} \\
         & \begin{overpic}[width=0.31\linewidth]{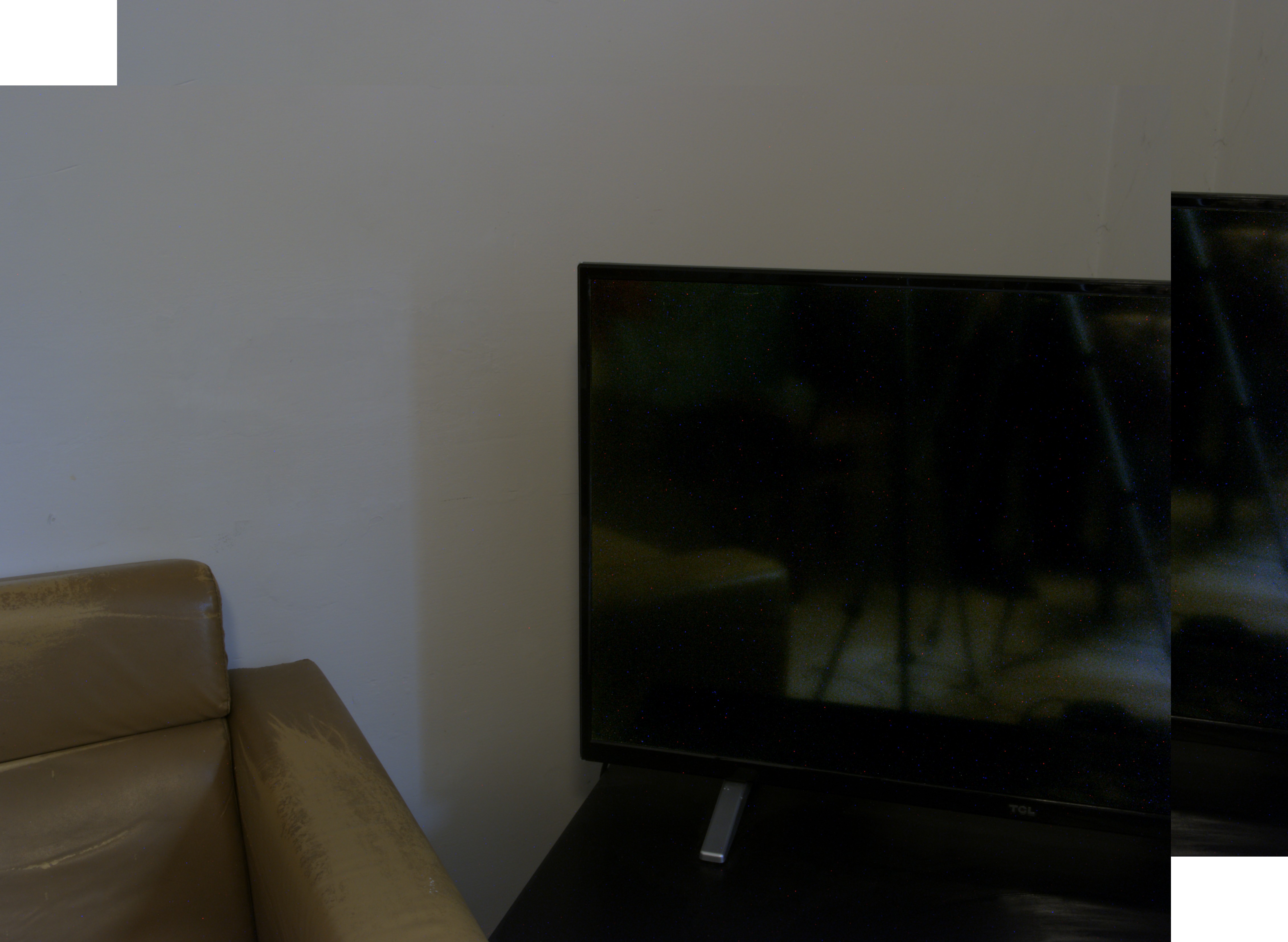}
         \end{overpic} &
         \linethickness{1.5pt}
         \begin{overpic}[width=0.31\linewidth]{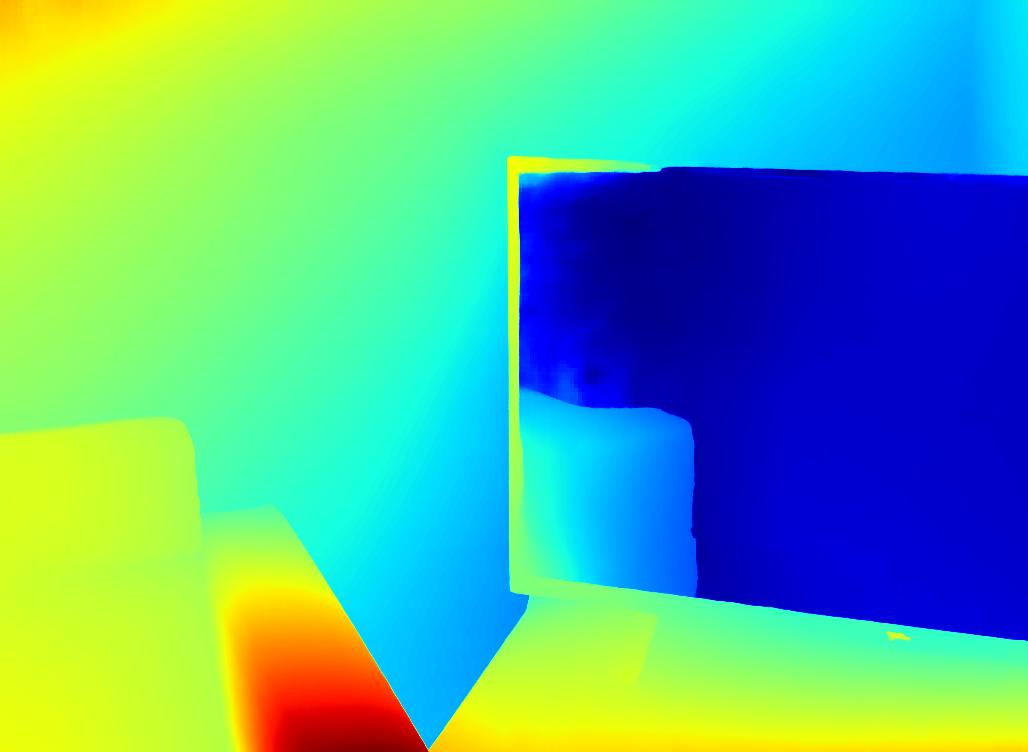}
         \put(45,7){\color{magenta}\line(0,2){55}}
         \put(45,62){\color{magenta}\line(2,0){50}}
         \put(95,7){\color{magenta}\line(0,2){55}}
         \put(45,7){\color{magenta}\line(2,0){50}}
         \end{overpic} &
         \begin{overpic}[width=0.31\linewidth]{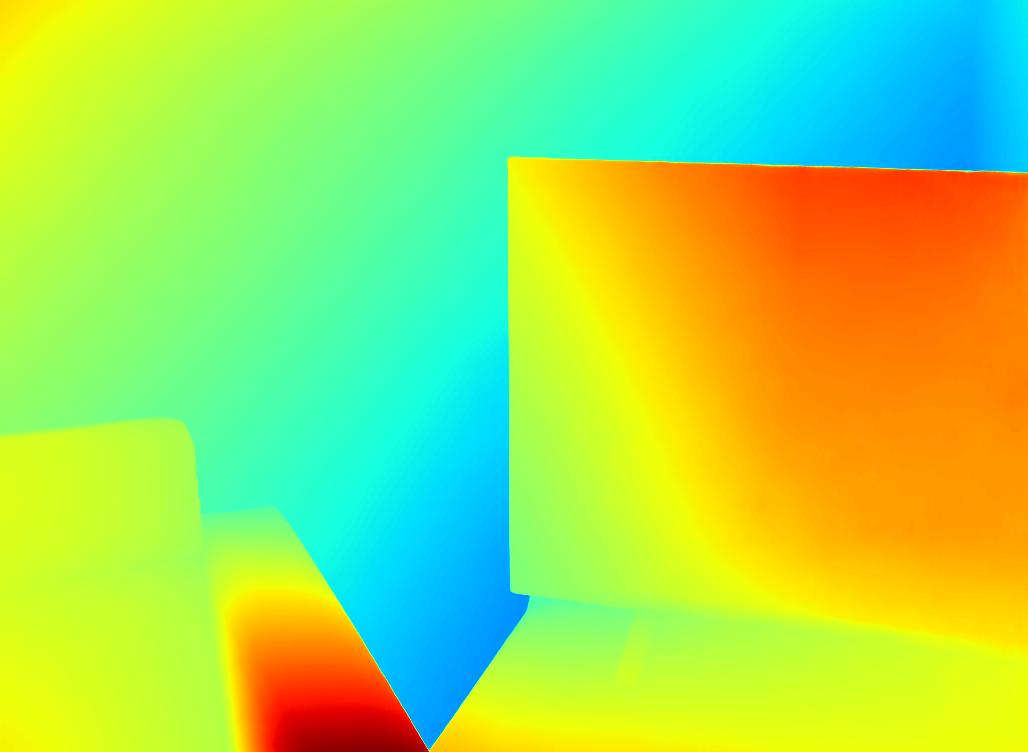}
         \end{overpic} \\
    \end{tabular}}
    \caption{\textbf{Depth estimation on ToM surfaces.} Two examples for both monocular (top) and stereo (bottom) images. In the central column, the depth/disparity maps predicted by DPT \cite{Ranftl2021} and CREStereo \cite{li2022practical} original weights. In the rightmost column, the depth/disparity maps predicted by the models after being fine-tuned by our strategy without exploiting any ground-truth depth.} 
    \label{fig:teaser}
\end{figure}
This difficulty arises because ToM surfaces introduce misleading visual information about scene geometry, which makes depth estimation challenging not only for computer vision systems but even for humans -- e.g., we might not distinguish the presence of a glass door in front of us due to its transparency.
On the one hand, the definition of depth itself might appear ambiguous in such cases: is \textit{depth} the distance to the scene behind the glass door or to the door itself? Nonetheless, from a practical point of view, we argue that the actual definition  depends on the task itself -- e.g., a mobile robot should definitely be aware of the presence of the glass door.
On the other hand, as humans can deal with this through experience, depth sensing techniques based on deep learning, e.g., monocular  \cite{Ranftl2022,Ranftl2021} or stereo \cite{lipson2021raft,li2022practical} networks, hold  the potential to address this challenge given sufficient training data \cite{zamaramirez2022booster}.

Unfortunately, light reflection and refraction over ToM  surfaces violate also the working principles of most active depth sensors, such as Time-of-Flight (ToF) cameras or devices projecting structured-light patterns. This  has two practical consequences:  i) it makes active sensors unsuited to deal with ToM objects in real-world applications, and ii) prevents the use of these sensors for collecting and annotating data to train deep neural networks to deal with ToM objects. As evidence of this, very few datasets featuring transparent objects provide ground-truth depth annotations, which have been obtained through very intensive human intervention \cite{zamaramirez2022booster}, graphical engines \cite{sajjan2020clear}, or based on  the availability of CAD models \cite{chen2022clearpose} for ToM objects.

In short, accurately perceiving the presence (and depth) of ToM objects represents an open challenge for both sensing technologies and deep learning frameworks.
Purposely, this paper proposes a simple yet effective strategy for obtaining training data and, thereby, dramatically boosting the accuracy of learning-based depth estimation frameworks dealing with ToM surfaces. Driven by the observation that ToM objects alone are responsible for misleading recent monocular networks \cite{Ranftl2022,Ranftl2021}, which would otherwise generalize well to most unseen environments, we argue that \textit{replacing} them with equivalent, yet opaque objects would allow restoring an environment layout in which such networks could accurately estimate the depth of the scene. To this end, we mask ToM objects in images by in-painting them with  arbitrary uniform colors.
Then, we employ a monocular depth network to generate a \textit{virtual} depth map out of the modified image. By repeating this process on a variety of images featuring ToM objects, we can easily and effectively annotate a dataset and then use it to train the same monocular network used to distill labels, which will now process the not-in-painted images. 
As a result, the trained monocular network will learn to handle ToM objects, producing consistent depth even in their presence. 

Our main contributions can be resumed as follows:

\begin{itemize}
    \item We propose a simple yet very effective strategy to deal with ToM objects. We trick a monocular depth estimation network by replacing ToM objects with  virtually textured ones, inducing it to hallucinate their depths.

    \item We introduce a processing pipeline for fine-tuning a monocular depth estimation network to deal with ToM objects. Our pipeline exploits the network itself to generate virtual depth annotations and requires only segmentation masks delineating ToM objects -- either human-made or predicted by other networks \cite{xie2020segmenting,Yang_2019_ICCV} -- thus getting rid of the need for any depth annotations.

    \item We show how our strategy can be extended to other depth estimation settings, such as stereo matching. Our experiments on the Booster dataset \cite{zamaramirez2022booster} prove how monocular and stereo networks dramatically improve their prediction on ToM objects after being fine-tuned according to  our methodology.
\end{itemize}
Fig. \ref{fig:teaser} highlights some specific regions where monocular (top) and stereo (bottom) models struggle (middle column), and how they learn to handle ToM surfaces thanks to our strategy (rightmost column).

The project page is available at \url{https://cvlab-unibo.github.io/Depth4ToM/}.



\begin{figure*}[t]
    \centering
    \includegraphics[width=0.8\linewidth]{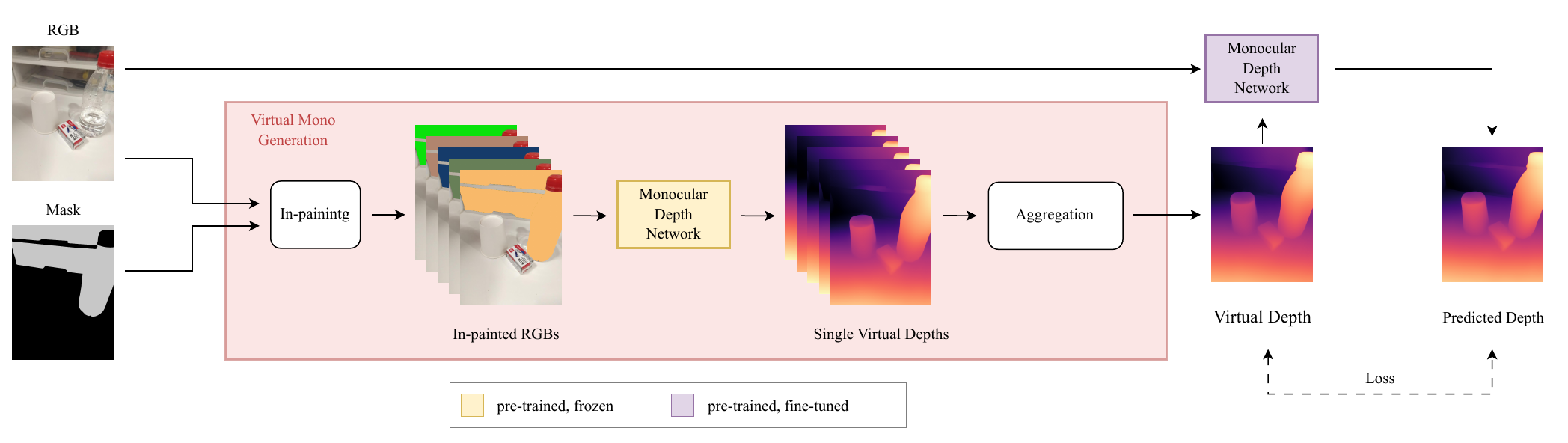}
    \caption{\textbf{Monocular Distillation pipeline.} Given an RGB and a segmentation mask, we in-paint pixels belonging to transparent and mirror surfaces with a random uniform color and process these augmented images with a pre-trained monocular network. The obtained virtual depths are aggregated to obtain a pseudo-labeled dataset for fine-tuning the network itself.}
    \label{fig:framework_mono}
\end{figure*}

\section{Related Work} 


\textbf{Monocular Depth Estimation.} Early methods used CNNs for pixel-level regression \cite{garg2016unsupervised, monodepth17}. More recent approaches such as AdaBins \cite{bhat2021adabins}, DPT \cite{Ranftl2021}, and MiDaS \cite{Ranftl2022} use adaptive bins and vision transformers for depth regression and leverage large-scale depth training by mixing multiple datasets.
Self-supervised methods use view synthesis for image reconstruction, where predicted depth is combined with known or estimated camera pose to establish correspondences between adjacent images, exploiting either stereo pairs \cite{garg2016unsupervised, monodepth17} or monocular videos \cite{zhou2017unsupervised, monodepth2}.
%
Recent works aim to improve the robustness of the photometric loss based on SSIM and L1 \cite{zhao2020monocular} by incorporating photometric uncertainty \cite{yang2020d3vo, poggi2020uncertainty}, feature descriptors \cite{zhan2018unsupervised, shu2020feature, spencer2020defeat}, 3D geometric constraints \cite{mahjourian2018unsupervised}, proxy supervision \cite{watson2019self, tosi2019learning}, optical flow \cite{yin2018geonet, tosi2020distilled}, or adversarial losses \cite{aleotti2018generative, pilzer2018unsupervised}.
Others propose architecture changes as in \cite{zhao2022monovit, guizilini20203d, pillai2019superdepth, johnston2020self, gonzalezbello2020forget}.
Except for some works that address non-Lambertian depth estimation using depth completion approaches and sparse depth measurements from active sensors \cite{choi2021selfdeco, sajjan2020clear}, to the best of our knowledge, we are not aware of any previous single-view depth estimation network that can handle ToM surfaces.  

\textbf{Stereo Matching.} Traditional algorithms \cite{scharstein2002taxonomy} utilize handcrafted features to estimate a disparity map \cite{Secaucus_1994_ECCV, hirschmuller2007stereo, yang2008stereo, yang2010constant, liang2011hardware, taniai2014graph, kolmogorov2004energy, boykov2001fast}. Then, deep learning methods replaced traditional matching cost computation, as demonstrated in \cite{zbontar2016stereo}, and, eventually,  end-to-end approaches became the most effective solution for disparity estimation. These networks can be mainly categorized into 2D and 3D architectures, with the former adopting an encoder-decoder design \cite{mayer2016large, Pang_2017_ICCV_Workshops, Liang_2018_CVPR, saikia2019autodispnet, song2018edgestereo, yang2018segstereo, yin2019hierarchical, Tankovich_2021_CVPR} and the latter building a feature cost volume from extracted features on the image pair \cite{Kendall_2017_ICCV, chang2018psmnet, khamis2018stereonet, zhang2019ga, cheng2019learning, cheng2020hierarchical, duggal2019deeppruner, yang2019hierarchical, wang2019anytime, guo2019group, Shen_2021_CVPR}. A thorough review of these works can be found in \cite{poggi2021synergies}. Recent papers exploit iterative refinement paradigms \cite{lipson2021raft, li2022practical} or rely on  Vision Transformers \cite{li2021revisiting,guo2022context}. However, due to its inherently ill-posed nature, dealing with non-Lambertian surfaces, such as ToM objects, remains a very challenging problem for any kind of existing stereo approach.  

\textbf{Non-Lambertian Object Perception.} 
Due to the relevance of dealing with ToM objects, some recent datasets focus on them.
Trans10K \cite{xie2020segmenting} and MSD \cite{Yang_2019_ICCV} consist of over 10\,000 and 4\,000 real in-the-wild images of transparent objects and mirrors, respectively. Both datasets provide manually annotated segmentations of ToM materials, though none of them provide depth labels.
Others provide depth annotations:
ClearPose \cite{chen2022clearpose} includes over 350\,000 labeled real-world RGB-D frames of 63 household objects.
ClearGrasp \cite{sajjan2020clear} consists of over 50\,000 synthetic RGB-D images of transparent objects, as well as real-world test benchmark with 286 RGB-D images.
In addition, Booster \cite{zamaramirez2022booster} focuses on stereo matching, providing  high-resolution depth labels and stereo pairs acquired in indoor scenes with specular and transparent surfaces.
TOD \cite{Liu_2020_CVPR} contains 15 transparent objects, labeled with relevant 3D keypoints, comprising 48\,000 stereo and RGBD images.
StereOBJ-1M \cite{liu2021stereobj} also deals with stereo vision, but focuses on pose estimation for ToM objects and do not provides depth ground truths.
%
Obtaining depth labels for these kinds of datasets is expensive, challenging, and time-consuming since it requires either CAD models for ToM objects \cite{chen2022clearpose}, painting such objects in the scene \cite{sajjan2020clear,zamaramirez2022booster,Liu_2020_CVPR} or relies on a complex multi-camera setup \cite{whelan2018reconstructing}.
In contrast, our proposal effectively sidesteps these challenges, by demonstrating that monocular and stereo networks can learn to deal with these objects in the absence of depth annotations.


\section{Method}


Our goal is to generate depth annotations for images featuring ToM objects in a cheap and scalable manner. This allows for training deep networks to properly estimate their depth as the distance of the closest surface in front of the camera, rather than the distance of the scene content refracted/reflected through it.
Our strategy is simple yet dramatically effective and relies on the availability of recent pre-trained monocular depth estimation models \cite{Ranftl2022, Ranftl2021}, which are capable of strong generalization across a variety of  scenes though struggling to deal with ToM surfaces.
Based on the above state of affairs, we argue that ToM objects are often the sole elements harming the reliability of recent pre-trained  monocular depth estimation networks. Therefore, by virtually replacing these objects with textured artifacts that resemble their very same shapes, the monocular model may be possibly tricked and induced into estimating the depth of an opaque object, ideally placed at the very same spot in the scene. This methodology can be realized   by delineating ToM  objects, through manual annotations or a segmentation network, masking them from the image and then in-painting virtual textures within the masked areas.
On the one hand, since a proper detection of ToM objects is crucial to our methodology, manual labeling indisputably results in the most accurate choice, though it comes with significant annotation costs. On the other hand, relying on a segmentation network would alleviate this cost: 
one would need some initial human annotations for training, but this would then allow to segment a large number of images for free.
Unfortunately, the overall effectiveness of our methodology would be inevitably affected by the accuracy of the trained segmentation model. However, we reckon that annotating images with segmentation masks requires, definitely,  a vastly lower effort compared to depth annotation \cite{zamaramirez2022booster,liu2021stereobj}. 
Hence, we settled on exploring both the aforementioned approaches.  

The reader may argue that, as a consequence of our intuition, training a depth network to deal with ToM objects might be unnecessary -- indeed, it would be sufficient to segment and in-paint such objects at deployment time before estimating depth. However, we retort that such a methodology would rely heavily on the actual accuracy of the model trained to segment ToM objects, which is not granted to generalize. Moreover, it would add non-negligible computational cost -- i.e., the inference by a second network.
On the contrary, an offline training or fine-tuning procedure allows for exploiting human-made annotation -- if available -- and, potentially, enable the trained network to learn how to properly estimate depth on ToM surfaces and to get rid of the second network, as well as design advanced strategies for other depth estimation frameworks, e.g. deep stereo networks.
Our experiments will highlight that the former strategy results ineffective, while we achieve a large boost in accuracy by fine-tuning depth models with our approach.

In the remainder, we describe our methodology to deal with ToM objects.
Given a dataset of images $\mathcal{I}$, our pipeline sketched in Fig. \ref{fig:framework_mono} builds as follows: 
i) surface labeling, ii) in-painting and distillation, and iii) fine-tuning of the depth network on virtual labels. Additionally, we show how it can be revised to fine-tune also deep stereo networks.

\begin{figure}[t]
    \centering
    \setlength{\tabcolsep}{1pt}
    \scalebox{1}{
    \begin{tabular}{ccccc}
         \scriptsize \textbf{\textit{RGB}} & 
         \scriptsize \textbf{\textit{Mask GT}} & 
         \scriptsize \textbf{\textit{Base}} & 
         \scriptsize \textbf{\textit{N=1 (Gray)}} & 
         \scriptsize \textbf{\textit{N=5}} \\
         
         \includegraphics[width=0.18\linewidth]{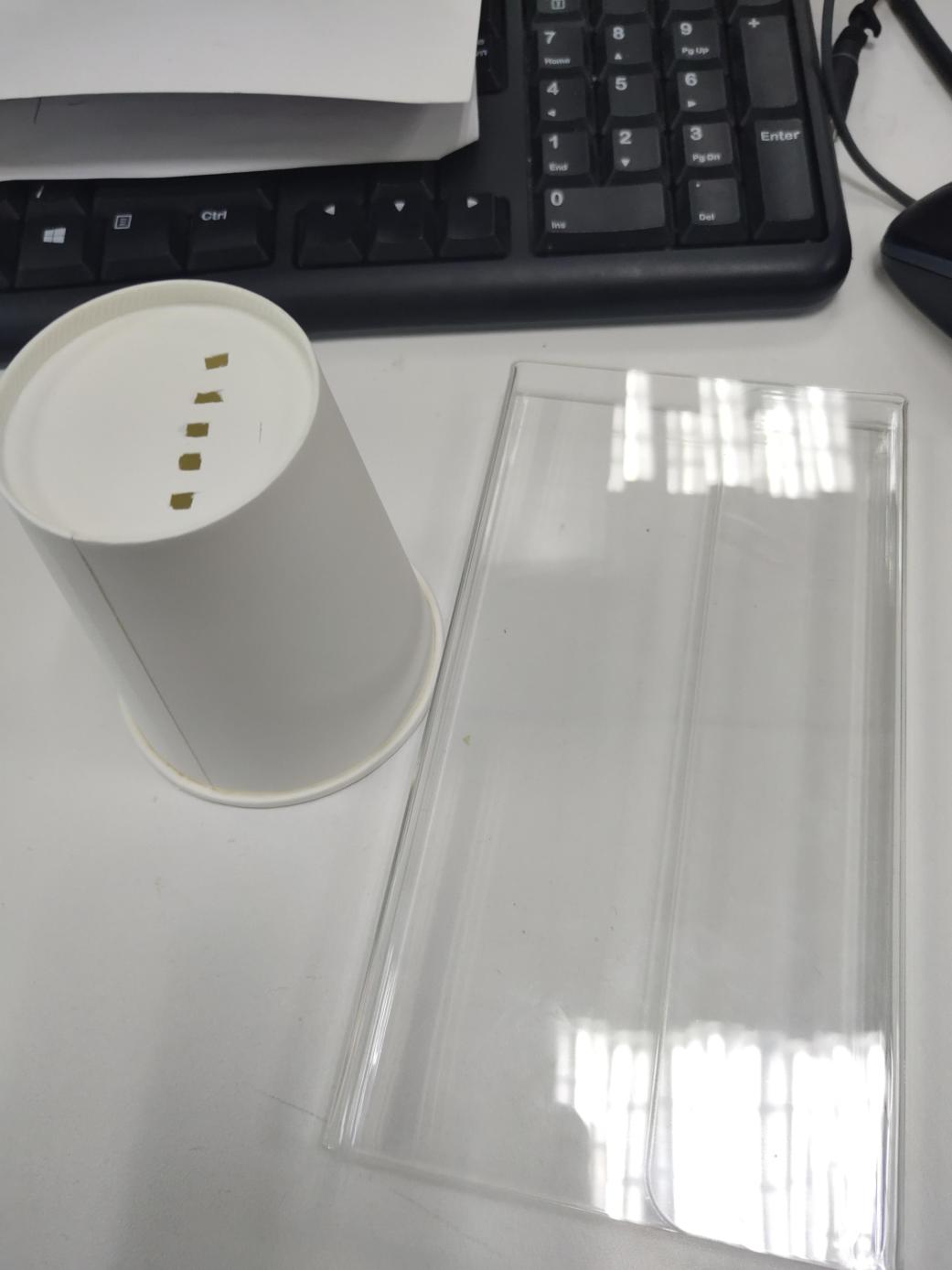} &
         \includegraphics[width=0.18\linewidth]{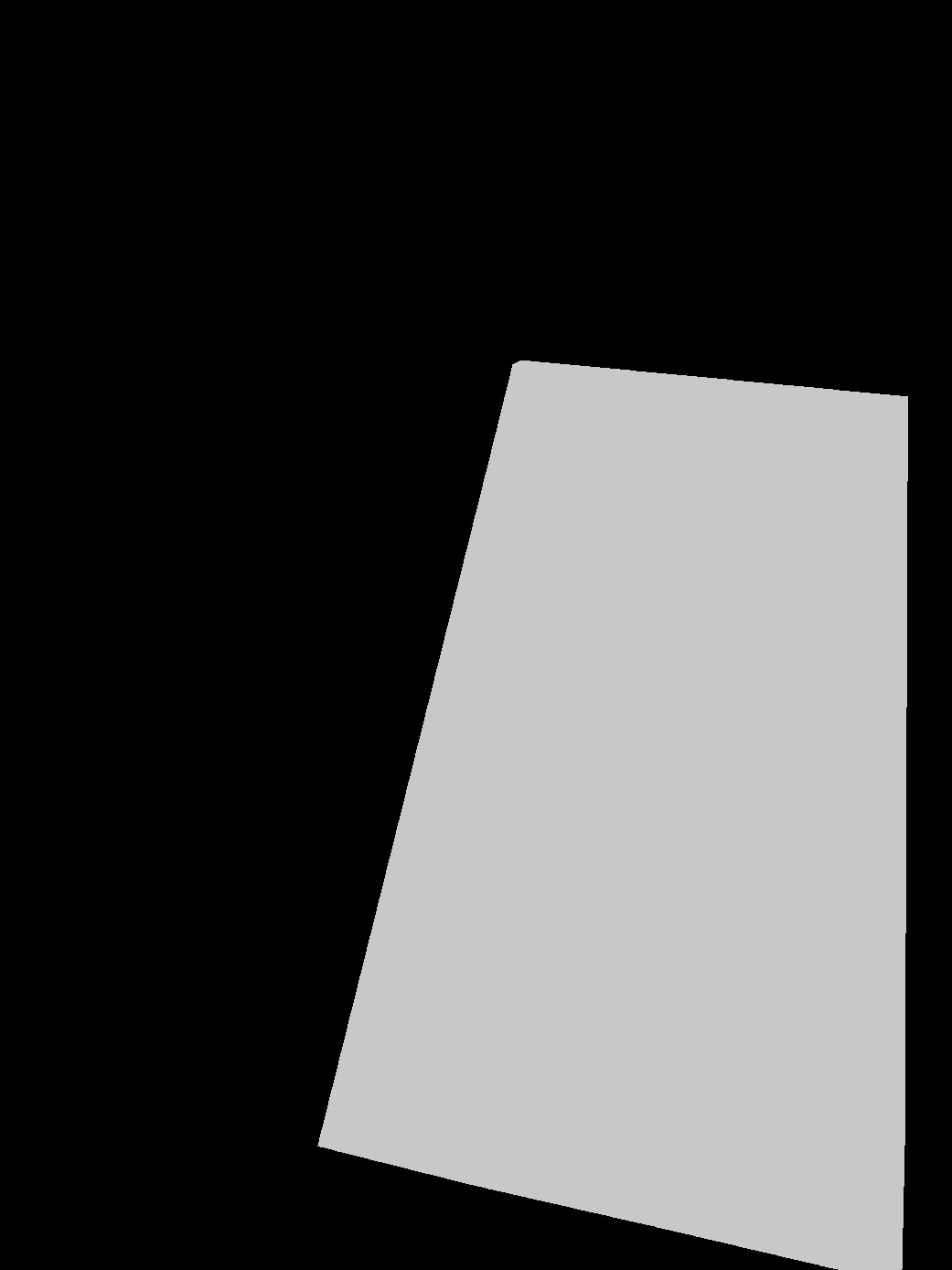} & 
         \includegraphics[width=0.18\linewidth]{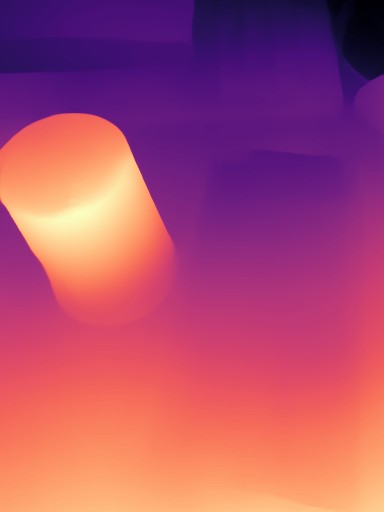} & 
         \includegraphics[width=0.18\linewidth]{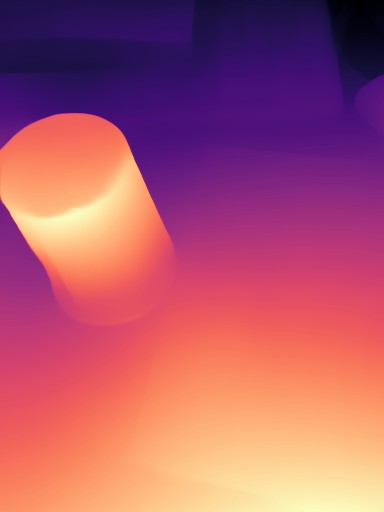} & 
         \includegraphics[width=0.18\linewidth]{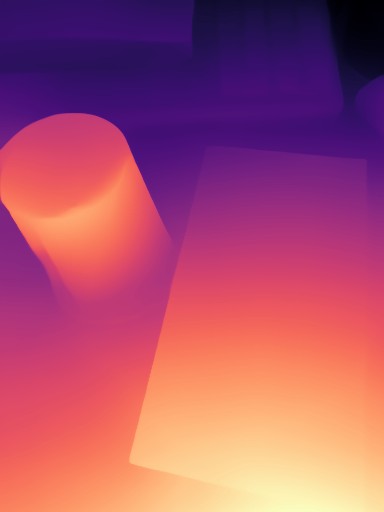} \\
    \end{tabular}}
    \caption{\textbf{Virtual depth generation alternatives.} From left to right: RGB, ground-truth segmentation, DPT predictions on the RGB image, on the gray-masked input, and the median of five predictions on images masked with random colors.}
    \label{fig:ablation_depth_proxy_mono}
\end{figure}

\textbf{Surface Labeling.} For any image $I_k \in \mathcal{I}$, we produce a segmentation mask $M_k$ classifying each pixel $p$ as

\begin{equation}
    M_k(p) = \begin{cases}
    1 & \text{if } I_k(p) \in \text{ToM surfaces } \\
    0 & \text{Otherwise} \\ 
    
\end{cases}
\end{equation}
by labeling pixels as either 1 or 0 if they belong to a ToM surface or not, respectively.
Such a segmentation mask can be  obtained either through manual annotation or by means of a segmentation network $\Theta$ as $M_k = \Theta(I_k)$.

\textbf{In-painting and Distillation.} Given an image $I_k$ and its corresponding segmentation mask $M_k$, we generated an augmented image $\tilde{I}_k$ applying an in-painting operation to replace the pixels belonging to ToM objects with a color $c$:

\begin{equation}
    \tilde{I}_k(p) = \begin{cases}
    c & \text{if } M_k(p) = 1 \\ 
    I_k & \text{otherwise } \\
\end{cases}
\end{equation}
Then, a virtual depth $\tilde{D}_k$ for image $I_k$ is obtained by forwarding $\tilde{I}_k$ to a monocular depth network $\Psi$ as $\tilde{D}_k = \Psi(\tilde{I}_k)$.
Colors are randomly sampled for every single frame $I_k$. However, depending on the image content, certain colors might result ineffective and increase the scene ambiguity -- e.g., by in-painting  white pixels into  a transparent object located in front of a white wall. To discourage these occurrences, we sample a set of $N$ custom colors $c_i, i \in [0, N-1]$, and  in-paint $I_k$ using each of these custom colors, so as to generate a set of $N$ augmented images $\tilde{I}^i_k$. Then, we obtain the final, \textit{Virtual Depth} $\tilde{D}_k$ by computing the per-pixel median between the $N$ depth maps 

\begin{equation}
    \tilde{D}^*_k = \text{med} \Bigl\{ \Psi(\tilde{I}^i_k), i \in [0,N-1] \Bigr\}
\end{equation}

As depicted in Fig. \ref{fig:ablation_depth_proxy_mono}, in some cases, the in-painted color might be similar to the background -- e.g., the transparent object disappears when a single gray mask is used -- while it is visible by aggregating multi-color in-painting.

\begin{table*}[t]
\centering
\scalebox{0.55}{
\begin{tabular}{ccccc}
 & MiDaS \cite{Ranftl2021} && DPT \cite{Ranftl2022} \\
 \begin{tabular}{ll}
 \toprule
 \\
 Category & Method \\
 \midrule
 \cellcolor{pink}All & Base \\
 \cellcolor{pink}All & Virtual Depth N=1 \\
 \cellcolor{pink}All & Virtual Depth N=5 \\
 \bottomrule
 \toprule
 \cellcolor{blue!25}ToM & Base \\
 \cellcolor{blue!25}ToM & Virtual Depth N=1 \\
 \cellcolor{blue!25}ToM & Virtual Depth N=5 \\
 \bottomrule
 \toprule
 \cellcolor{YellowOrange}Other & Base \\
 \cellcolor{YellowOrange}Other & Virtual Depth N=1 \\
 \cellcolor{YellowOrange}Other & Virtual Depth N=5 \\
 \bottomrule
 \end{tabular}
 &
 \begin{tabular}{rrrrr | rrr}
 \toprule
 $\delta$ $<$ 1.25 &  $\delta$ $<$ 1.20 & $\delta$ $<$ 1.15 & $\delta$ $<$ 1.10 & $\delta$ $<$ 1.05 & MAE & Abs. Rel & RMSE \\
 $\uparrow$ (\%) & $\uparrow$ (\%) & $\uparrow$ (\%) & $\uparrow$ (\%) & $\uparrow$ (\%) & $\downarrow$ (mm) & $\downarrow$ & $\downarrow$ (mm) \\
 \midrule
94.56 & 91.72 & 85.68 & 74.00 & \textbf{50.12} & 90.82 & \textbf{0.07} & 120.51 \\
\textbf{96.08} & 93.23 & 87.68 & 75.23 & 49.26 & 83.88 & \textbf{0.07} & 112.03 \\
96.04 & \textbf{93.51} & \textbf{87.93} & \textbf{75.70} & 49.36 & \textbf{82.98} & \textbf{0.07} & \textbf{110.52} \\ 
 \bottomrule
 \toprule
87.44 & 83.40 & 72.71 & 59.63 & 36.28 & 122.33 & 0.12 & 140.31 \\
\textbf{94.11} & \textbf{91.99} & \textbf{84.12} & 68.40 & 41.17 & 76.69 & 0.09 & 86.46 \\
93.87 & 91.64 & 83.66 & \textbf{68.76} & \textbf{43.65} & \textbf{76.65} & \textbf{0.08} & \textbf{86.01} \\
 \bottomrule
 \toprule
 94.57 & 91.81 & 85.99 & 74.01 & \textbf{50.28} & 91.08 & \textbf{0.07} & 119.86 \\ 
 95.62 & 92.50 & 86.77 & 74.63 & 48.76 & 88.30 & \textbf{0.07} & 116.78 \\
 \textbf{95.66} & \textbf{92.93} & \textbf{87.31} & \textbf{75.42} & 49.06 & \textbf{86.49} & \textbf{0.07} & \textbf{114.48} \\
 \bottomrule
 \end{tabular}
 & &
 \begin{tabular}{rrrrr | rrr}
 \toprule
 $\delta$ $<$ 1.25 &  $\delta$ $<$ 1.20 & $\delta$ $<$ 1.15 & $\delta$ $<$ 1.10 & $\delta$ $<$ 1.05 & MAE & Abs. Rel & RMSE \\
 $\uparrow$ (\%) & $\uparrow$ (\%) & $\uparrow$ (\%) & $\uparrow$ (\%) & $\uparrow$ (\%) & $\downarrow$ (mm) & $\downarrow$ & $\downarrow$ (mm) \\
 \midrule
96.79 & 94.45 & 89.71 & 79.00 & 56.26 & 75.35 & 0.06 & 100.68 \\ 
97.59 & 95.88 & 92.06 & 82.75 & 60.14 & 64.99 & \textbf{0.05} & 85.96 \\
\textbf{98.43} & \textbf{96.74} & \textbf{92.86} & \textbf{83.42} & \textbf{60.18} & \textbf{62.46} & \textbf{0.05} & \textbf{82.06} \\
 \bottomrule
 \toprule
92.77 & 88.77 & 80.98 & 62.46 & 37.70 & 113.14 & 0.10 & 136.28 \\
96.00 & 93.67 & 88.88 & 75.79 & 45.26 & 65.58 & 0.07 & 78.24 \\
\textbf{98.94} & \textbf{97.19} & \textbf{92.24} & \textbf{77.52} & \textbf{45.97} & \textbf{57.19} & \textbf{0.06} & \textbf{66.86} \\
 \bottomrule
 \toprule
97.10 & 94.84 & 90.08 & 79.76 & 57.31 & 73.19 & 0.06 & 95.63 \\
97.63 & 95.96 & 92.09 & 83.11 & \textbf{61.01} & 66.10 & \textbf{0.05} & 87.37 \\
\textbf{98.29} & \textbf{96.50} & \textbf{92.57} & \textbf{83.51} & 60.90 & \textbf{64.17} & \textbf{0.05} & \textbf{84.06} \\
 \bottomrule
 \end{tabular}
 \end{tabular}
 }
 \caption{\textbf{Virtual depth distillation by varying $N$.}
 Results on Booster train set at quarter resolution. All networks use the official weights \cite{Ranftl2021, Ranftl2022} without further training. Different masking strategies are applied to the RGB input image. Best results in \textbf{bold}.}
 \label{tab:ablation_mask_generalization}\vspace{-0.2cm}
\end{table*}
\begin{figure}[t]
    \centering
    \includegraphics[width=0.85\linewidth]{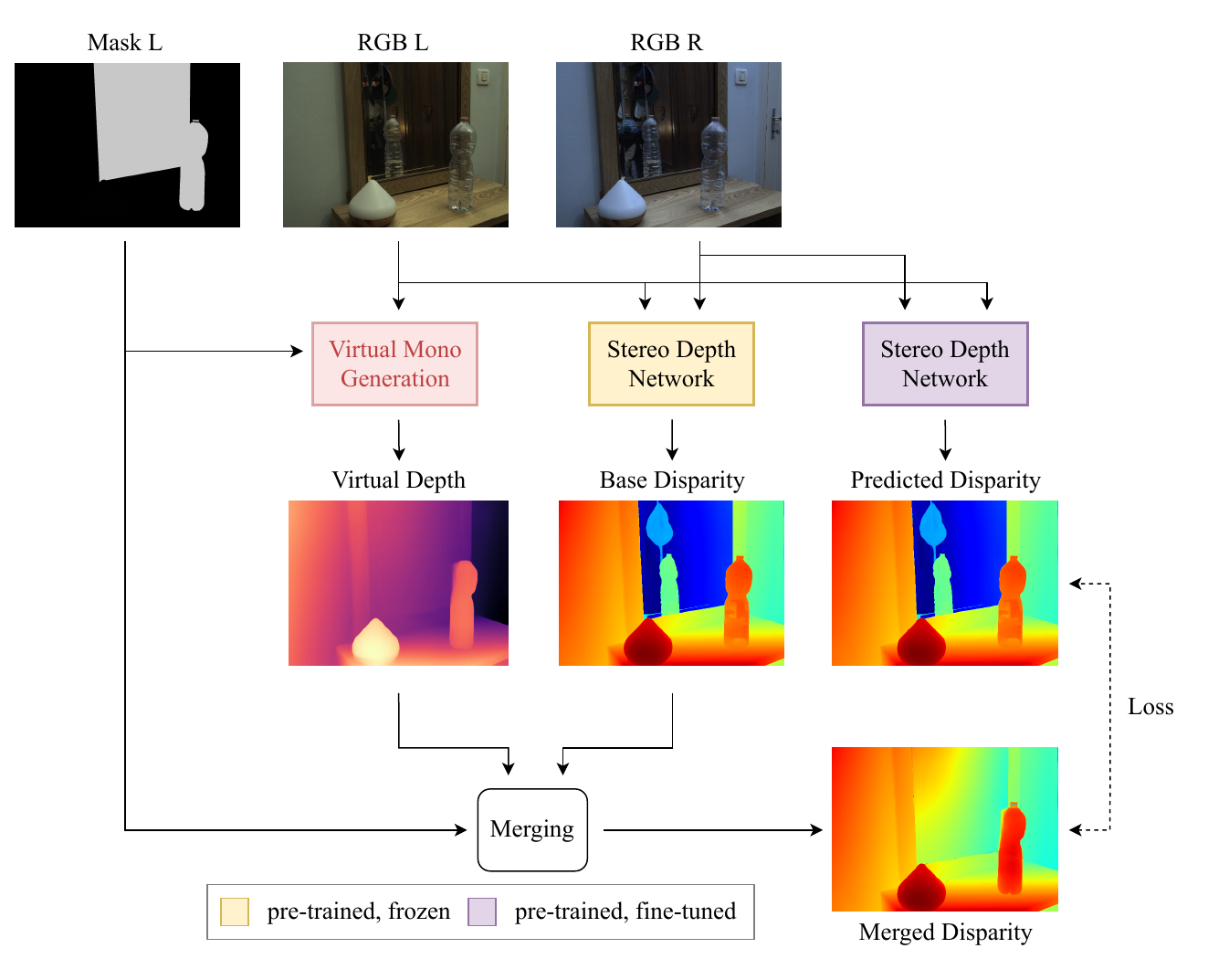}
    \caption{\textbf{Stereo Distillation pipeline.} 
    Given a stereo pair and a segmentation mask for the left image, we merge predictions from a pre-trained stereo network with virtual depths obtained by a monocular network with our strategy. We merge the monocular or stereo maps, by taking values belonging to either ToM or other surfaces from mono or stereo, respectively. These final merged depth labels are used to fine-tune the original stereo network.}
    \label{fig:framework_stereo}
\end{figure}

\textbf{Fine-Tuning on Virtual Labels.} The steps outlined so far allow for labeling a dataset $\mathcal{I}$ with virtual depth labels that are not influenced by the ambiguities of ToM objects. Then, our newly annotated dataset can be used to train or fine-tune a depth estimation network, thereby enabling it to handle the aforementioned difficult  objects robustly.
Specifically, during training, the original images $I_k$ are forwarded to the network, and the predicted depth $\hat{D}_k$ is optimized with respect to the distilled virtual ground-truth map $\tilde{D}^*_k$ obtained from in-painted images.
This simple pipeline can dramatically improve the accuracy of monocular  depth estimation networks when dealing with ToM objects, as we will show in our experiments.

\textbf{Extension to Deep Stereo.} Our pipeline can be adapted to fine-tune deep stereo models as well, as shown in Fig. \ref{fig:framework_stereo}. Again, we argue that state-of-the-art stereo architectures \cite{lipson2021raft,li2022practical} already expose outstanding generalization capabilities while struggling with ToM objects, due to the task of matching pixels belonging to non-Lambertian surfaces being inherently ambiguous.
Consequently, we exploit a monocular depth estimation network to obtain virtual depth annotations solely for these objects. Given a dataset $\mathcal{S}$ consisting of stereo pairs $(L_k,R_k)$, we distill virtual depth labels $\tilde{D}^*_k$ from $L_k$ and triangulate them into disparities $\tilde{d}^*_k$ according to extrinsic parameters of the stereo rig. Then, we predict a \textit{Base} disparity map $d_k$ by forwarding $(L_k,R_k)$ to the stereo network we aim to fine-tune. Eventually, we replace the disparities for ToM objects with those from $d_k$ according to $M_k$, this latter produced over $L_k$ this time. Formally, this operation namely \textit{Merging}, is defined as:

\begin{equation}
\label{eq:merging}
    d_k(p) = \begin{cases}
    d_k(p) & \text{if } M_k(p) = 0 \\ 
    \alpha_k\hat{d}^*_k(p) + \beta_k & \text{otherwise } \\
\end{cases}
\end{equation}
with $\alpha_k,\beta_k$ being scale and shift factors, as monocular predictions are up to an unknown scale factor. Following \cite{Ranftl2022}, $\alpha_k,\beta_k$ are estimated through Least Square Estimation (LSE) regression over $d_k$ for pixels not belonging to any ToM object, i.e., having $M_k(p)=0$:

\begin{equation}
\label{eq:rescaling}
    (\alpha_k,\beta_k) = \text{arg}\min_{\alpha,\beta} \sum_{p| M_k(p)=0} \Big( \alpha \hat{d}^*_k(p) + \beta - d_k(p) \Big)^2
\end{equation}



\begin{table*}[t]
\centering
\scalebox{0.55}{
\begin{tabular}{ccccc}
 & MiDaS \cite{Ranftl2021} && DPT \cite{Ranftl2022} \\
 \begin{tabular}{ll}
 \toprule
 \\
 Category & Method \\
 \midrule
 \cellcolor{pink}All & Base \\
 \cellcolor{pink}All & Ft. Base \\
 \cellcolor{pink}All & Ft. Virtual Depth \\
 \bottomrule
 \toprule
 \cellcolor{blue!25}ToM  & Base \\
 \cellcolor{blue!25}ToM  & Ft. Base \\
 \cellcolor{blue!25}ToM  & Ft. Virtual Depth \\
 \bottomrule
 \toprule
 \cellcolor{YellowOrange}Other& Base \\
 \cellcolor{YellowOrange}Other& Ft. Base \\
 \cellcolor{YellowOrange}Other & Ft. Virtual Depth \\
 \bottomrule
 \end{tabular}
 &
 \begin{tabular}{rrrrr | rrr}
 \toprule
 $\delta$ $<$ 1.25 &  $\delta$ $<$ 1.20 & $\delta$ $<$ 1.15 & $\delta$ $<$ 1.10 & $\delta$ $<$ 1.05 & MAE & Abs. Rel & RMSE \\
 $\uparrow$ (\%) & $\uparrow$ (\%) & $\uparrow$ (\%) & $\uparrow$ (\%) & $\uparrow$ (\%) & $\downarrow$ (mm) & $\downarrow$ & $\downarrow$ (mm) \\
 \midrule
 94.56 & 91.72 & 85.68 & 74.00 & 50.12 & 90.82 & \textbf{0.07} & 120.51 \\
94.09 & 91.28 & 85.29 & 73.55 & 49.14 & 93.34 & \textbf{0.07} & 124.54 \\
\textbf{95.07} & \textbf{92.31} & \textbf{86.39} & \textbf{75.20} & \textbf{50.57} & \textbf{88.83} & \textbf{0.07} & \textbf{118.23} \\
 \bottomrule
 \toprule
 87.44 & 83.40 & 72.71 & 59.63 & 36.28 & 122.33 & 0.12 & 140.31 \\
86.30 & 82.96 & 72.84 & 61.04 & 37.91 & 126.71 & 0.12 & 145.69 \\ 
\textbf{91.81} & \textbf{89.12} & \textbf{81.68} & \textbf{70.75} & \textbf{47.85} & \textbf{93.66} & \textbf{0.09} & \textbf{104.80} \\
 \bottomrule
 \toprule
 94.57 & 91.81 & 85.99 & 74.01 & 50.28 & 91.08 & \textbf{0.07} & 119.86 \\ 
94.27 & 91.49 & 85.87 & 73.77 & 49.38 & 92.48 & \textbf{0.07} & 122.46 \\
\textbf{95.33} & \textbf{92.53} & \textbf{86.80} & \textbf{75.43} & \textbf{50.46} & \textbf{89.17} & \textbf{0.07} & \textbf{119.58} \\
 \bottomrule
 \end{tabular}
 & &
 \begin{tabular}{rrrrr | rrr}
 \toprule
 $\delta$ $<$ 1.25 &  $\delta$ $<$ 1.20 & $\delta$ $<$ 1.15 & $\delta$ $<$ 1.10 & $\delta$ $<$ 1.05 & MAE & Abs. Rel & RMSE \\
 $\uparrow$ (\%) & $\uparrow$ (\%) & $\uparrow$ (\%) & $\uparrow$ (\%) & $\uparrow$ (\%) & $\downarrow$ (mm) & $\downarrow$ & $\downarrow$ (mm) \\
 \midrule
 96.79 & 94.45 & 89.71 & 79.00 & 56.26 & 75.35 & 0.06 & 100.68 \\ 
96.83 & 94.60 & 90.14 & 79.46 & 56.89 & 76.00 & 0.06 & 100.98 \\
\textbf{97.99} & \textbf{96.65} & \textbf{93.55} & \textbf{83.94} & \textbf{60.46} & \textbf{64.93} & \textbf{0.05} & \textbf{85.93} \\
 \bottomrule
 \toprule
 92.77 & 88.77 & 80.98 & 62.46 & 37.70 & 113.14 & 0.10 & 136.28 \\
92.69 & 89.49 & 81.12 & 63.62 & 37.95 & 118.84 & 0.11 & 141.27 \\
\textbf{96.68} & \textbf{95.96} & \textbf{92.23} & \textbf{79.96} & \textbf{54.67} & \textbf{70.68} & \textbf{0.06} & \textbf{83.06} \\
 \bottomrule
 \toprule
 97.10 & 94.84 & 90.08 & 79.76 & 57.31 & 73.19 & 0.06 & 95.63 \\
97.15 & 95.02 & 90.97 & 80.77 & 58.05 & 72.45 & 0.06 & 94.07 \\
\textbf{98.07} & \textbf{96.64} & \textbf{93.52} & \textbf{84.30} & \textbf{61.19} & \textbf{64.70} & \textbf{0.05} & \textbf{85.57} \\
 \bottomrule
 \end{tabular}
 \end{tabular}
 }
 \caption{\textbf{Monocular networks fine-tuning - ground-truth segmentation.} Training on all MSD and Trans10K, results on the Booster train set at quarter resolution. All models start from the official weights \cite{Ranftl2021,Ranftl2022}. Best results in \textbf{bold}.}\vspace{-0.3cm}
 \label{tab:finetuning_mono}
\end{table*}

\section{Experimental Settings}


\textbf{Implementation Details.}
We employ MiDaS \cite{Ranftl2022} and DPT \cite{Ranftl2021} as our monocular networks using the official pre-trained weights, 
given their excellent in-the-wild generalization performance. To fine-tune them, we iterate for 20 epochs with batch size 8 and a learning rate of $10^{-7}$ with exponential decay with gamma 0.95. We use random color and brightness and random horizontal flip augmentations. We pad/crop and resize images to match the pre-training resolution, i.e., 384 pixels for the long or short side, preserving aspect ratio with mirror pad or square crop, for MiDaS or DPT, respectively.
We normalize images as the original networks do.
Regarding stereo networks, we employ RAFT \cite{lipson2021raft} and CREStereo \cite{li2022practical}, using the official pre-trained weights, 
since they achieve the top rankings in the Middlebury dataset \cite{scharstein2014high} among published methods.
To fine-tune them, we run 20 epochs, with batch size 2, fixed learning rate $10^{-5}$. Following \cite{zamaramirez2022booster}, we randomly resize images to half or quarter of the original dataset resolution, randomly crop to 456$\times$884 and 448$\times$880 for RAFT and CREStereo respectively, and further randomly scale images and disparities by a factor $\in [0.9, 1.1]$. We assume 22 and 10 iterations during training for RAFT-Stereo and CREStereo, respectively. During testing, we run 32 and 20 iterations.
%
When creating virtual labels with our masking strategy, we fix the random seed of color sampling to 0.

\textbf{Datasets.}
Among the datasets, we selected Trans10K \cite{xie2020segmenting}, MSD\cite{Yang_2019_ICCV}, and Booster\cite{zamaramirez2022booster} as they focus on ToM surfaces and contain images acquired in many realistic environments. Trans10K contains 5\,003, 1\,003, 4\,431 images for the training, validation, and test set, respectively, featuring common transparent objects and  stuff. It provides segmentation masks with pixels categorized into 12 different classes that we collapse into 2 -- ToM (classes 1 to 11) or not. MSD contains 3\,066, and 958 images and binary segmentation masks for the training and test set, respectively, featuring mirrors. Booster contains 228, and 191 images for training and testing, respectively. The dataset provides disparity and segmentation maps for the training set, where the segmentation maps are categorized into 4 classes, which we group into 2
 -- classes 2-3 into ``ToM" category, classes 0-1 into ``Other" materials.
%
We fine-tune on Trans10K and MSD for monocular models and on the Booster training split for stereo networks, without using any depth ground truths.

\begin{table*}[t]
\centering
\scalebox{0.52}{
\begin{tabular}{ccccc}
 & MiDaS \cite{Ranftl2021} && DPT \cite{Ranftl2022} \\
 \begin{tabular}{ll}
 \toprule
 \\
 Category & Method \\
 \midrule
 \cellcolor{pink}All & Base \\
  \cellcolor{pink}All & Virtual Depth (Proxy) \\
 \cellcolor{pink}All & Ft. Virtual Depth (GT) \\
 \cellcolor{pink}All & Ft. Virtual Depth (Proxy) \\
 \bottomrule
 \toprule
 \cellcolor{blue!25}ToM & Base \\
 \cellcolor{blue!25}ToM & Virtual Depth (Proxy) \\
 \cellcolor{blue!25}ToM & Ft. Virtual Depth (GT) \\
 \cellcolor{blue!25}ToM & Ft. Virtual Depth (Proxy) \\
 \bottomrule
 \toprule
 \cellcolor{YellowOrange}Other & Base \\
 \cellcolor{YellowOrange}Other & Virtual Depth (Proxy) \\
 \cellcolor{YellowOrange}Other & Ft. Virtual Depth (GT) \\
 \cellcolor{YellowOrange}Other & Ft. Virtual Depth (Proxy) \\
 \bottomrule
 \end{tabular}
 &
 \begin{tabular}{rrrrr | rrr}
 \toprule
 $\delta$ $<$ 1.25 &  $\delta$ $<$ 1.20 & $\delta$ $<$ 1.15 & $\delta$ $<$ 1.10 & $\delta$ $<$ 1.05 & MAE & Abs. Rel & RMSE \\
 $\uparrow$ (\%) & $\uparrow$ (\%) & $\uparrow$ (\%) & $\uparrow$ (\%) & $\uparrow$ (\%) & $\downarrow$ (mm) & $\downarrow$ & $\downarrow$ (mm) \\
 \midrule
 94.56 & 91.72 & 85.68 & 74.00 & 50.12 & 90.82 & \textbf{0.07} & 120.51 \\
 91.78 & 87.52 & 79.57 & 66.00 & 42.21 & 105.67 & 0.09 & 140.00 \\
94.99 & 92.12 & 86.32 & 74.99 & 49.91 & 88.63 & \textbf{0.07} & 117.40 \\
\textbf{95.00} & \textbf{92.17} & \textbf{86.48} & \textbf{75.46} & \textbf{50.24} & \textbf{88.10} & \textbf{0.07} & \textbf{117.10} \\
 \bottomrule
 \toprule
  87.44 & 83.40 & 72.71 & 59.63 & 36.28 & 122.33 & 0.12 & 140.31 \\
  86.37 & 79.35 & 69.91 & 54.84 & 32.31 & 112.92 & 0.12 & 124.10 \\
\textbf{91.85} & \textbf{89.93} & \textbf{83.25} & \textbf{70.63} & \textbf{47.55} & \textbf{92.05} & \textbf{0.09} & \textbf{103.54} \\
91.12 & 88.36 & 81.18 & 69.15 & 45.02 & 96.33 & \textbf{0.09} & 107.59 \\ 
 \bottomrule
 \toprule
  94.57 & 91.81 & 85.99 & 74.01 & 50.28 & 91.08 & \textbf{0.07} & 119.86 \\ 
  91.80 & 87.47 & 79.41 & 65.86 & 41.95 & 108.09 & 0.09 & 143.64 \\
\textbf{95.27} & 92.30 & 86.60 & 75.43 & 49.95 & 88.81 & \textbf{0.07} & 118.48 \\
\textbf{95.27} & \textbf{92.42} & \textbf{86.84} & \textbf{76.02} & \textbf{50.48} & \textbf{88.07} & \textbf{0.07} & \textbf{117.99} \\
 \bottomrule
 \end{tabular}
 & &
 \begin{tabular}{rrrrr | rrr}
 \toprule
 $\delta$ $<$ 1.25 &  $\delta$ $<$ 1.20 & $\delta$ $<$ 1.15 & $\delta$ $<$ 1.10 & $\delta$ $<$ 1.05 & MAE & Abs. Rel & RMSE \\
 $\uparrow$ (\%) & $\uparrow$ (\%) & $\uparrow$ (\%) & $\uparrow$ (\%) & $\uparrow$ (\%) & $\downarrow$ (mm) & $\downarrow$ & $\downarrow$ (mm) \\
 \midrule
 96.79 & 94.45 & 89.71 & 79.00 & 56.26 & 75.35 & 0.06 & 100.68 \\ 
 93.23 & 89.43 & 81.98 & 68.06 & 43.62 & 98.09 & 0.08 & 128.36 \\
98.09 & \textbf{96.85} & \textbf{93.91} & \textbf{83.50} & \textbf{58.74} & \textbf{65.52} & \textbf{0.05} & \textbf{86.41} \\
\textbf{98.11} & 96.68 & 93.48 & 83.00 & 58.13 & 66.43 & \textbf{0.05} & 87.18 \\
 \bottomrule
 \toprule
 92.77 & 88.77 & 80.98 & 62.46 & 37.70 & 113.14 & 0.10 & 136.28 \\
 89.75 & 85.26 & 75.17 & 56.57 & 31.75 & 110.74 & 0.11 & 123.28 \\
\textbf{96.95} & \textbf{96.26} & \textbf{93.27} & \textbf{81.84} & \textbf{55.49} & \textbf{67.95} & \textbf{0.06} & \textbf{80.88} \\
96.82 & 96.05 & 92.57 & 80.62 & 53.74 & 70.67 & \textbf{0.06} & 83.44 \\
 \bottomrule
 \toprule
 97.10 & 94.84 & 90.08 & 79.76 & 57.31 & 73.19 & 0.06 & 95.63 \\
 93.22 & 89.39 & 82.07 & 68.30 & 43.88 & 98.64 & 0.08 & 129.48 \\
98.19 & \textbf{96.85} & \textbf{93.85} & \textbf{83.70} & \textbf{58.93} & \textbf{65.65} & \textbf{0.05} & \textbf{86.30} \\
\textbf{98.20} & 96.67 & 93.35 & 83.20 & 58.40 & 66.51 & \textbf{0.05} & 87.06 \\
 \bottomrule
 \end{tabular}
 \end{tabular}
 }
 \caption{\textbf{Monocular networks fine-tuning -- ground-truth vs proxy segmentation.}
 Training on only the test set of MSD and Trans10K, results on the Booster train set
at quarter resolution. All models start from the official weights \cite{Ranftl2021, Ranftl2022}. Best results in \textbf{bold}.}
 \label{tab:proxy_vs_gt_mask}
\end{table*}

\textbf{Evaluation Protocol.}
We evaluate the accuracy of the monocular networks using several metrics, including absolute error relative to the ground-truth value (ABS Rel.), the percentage of pixels having the maximum between the prediction/ground-truth and ground-truth/prediction ratios lower than a threshold ($\delta_i$, with $i$ being 1.05, 1.10, 1.15, 1,20, and 1.25), the mean absolute error (MAE) and Root Mean Squared Error (RMSE).
Additionally, we evaluate stereo networks using the metrics defined in Booster \cite{zamaramirez2022booster}, i.e. bad-2, bad-4, bad-6, bad-8, MAE, RMSE. 
Results are reported on all valid pixels (\textit{All}) or for those belonging to either ToM or other objects, in order to assess the impact of our strategy on the different kinds of surfaces.
For any metrics considered for stereo networks, the lower, the better -- annotated with $\downarrow$ in tables. The same applies to metrics used for monocular networks except for $\delta_i$, resulting in the higher, the better -- with $\uparrow$ being reported in tables.
As the predictions by monocular networks are up to an unknown scale factor, we rescale them according to the LSE criterion from \cite{Ranftl2022} defined in Eq. \ref{eq:rescaling}, yet using all valid pixels here.
Monocular networks are evaluated on the Booster training set, while stereo models are evaluated on the Booster test set. As for the latter, results split into  ``ToM" and ``Other" objects have been kindly computed by the Booster authors based on the segmentation classes we defined. 

\section{Experiments}

\begin{figure}[t]
    \centering
    \setlength{\tabcolsep}{1pt}
    
    \begin{tabular}{ccc}
         \scriptsize \textbf{\textit{RGB}} & 
         \scriptsize \textbf{\textit{Mask GT}} & 
         \scriptsize \textbf{\textit{Mask Proxy}} \\
         \includegraphics[width=0.28\linewidth]{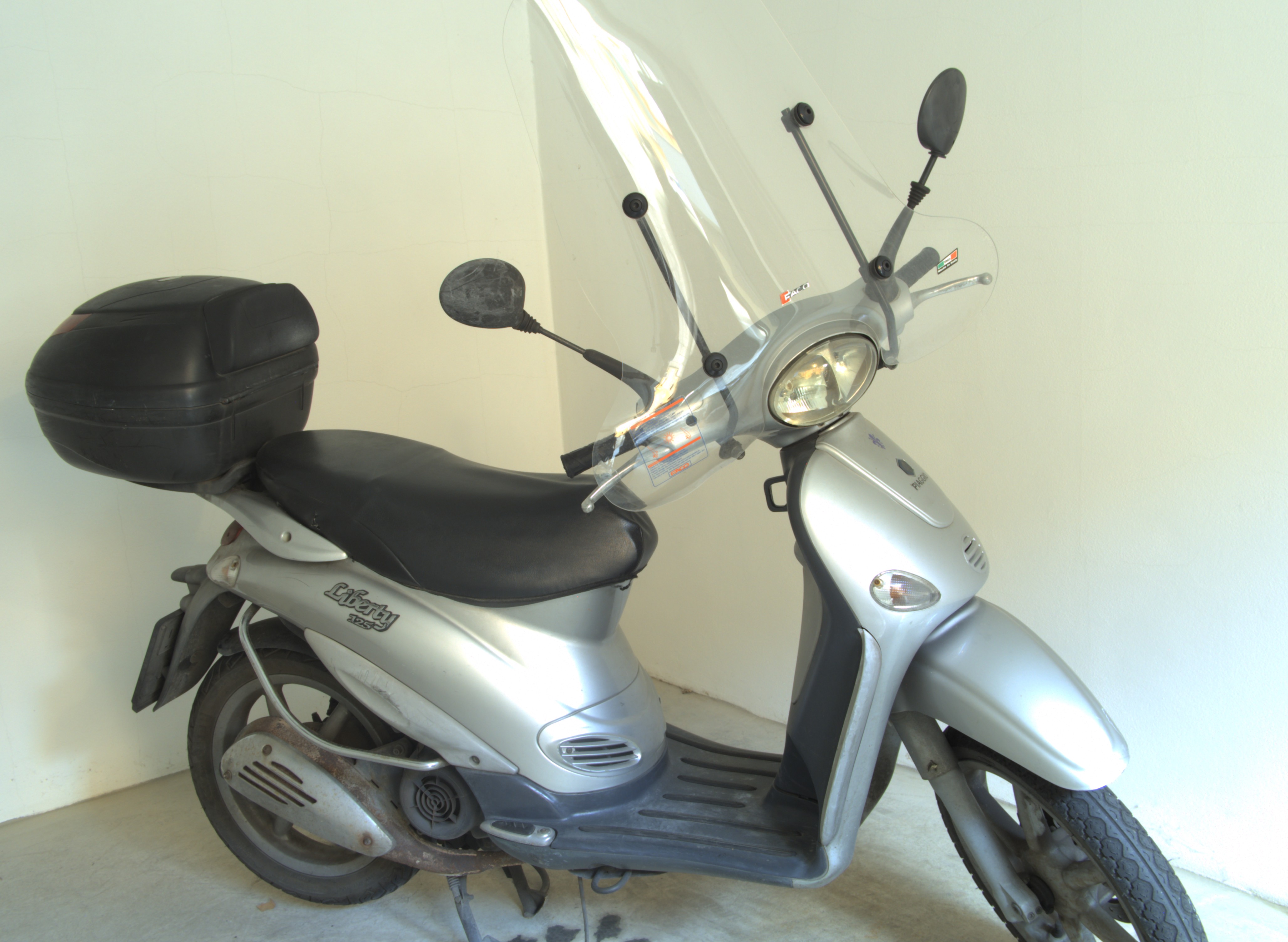} &
         \includegraphics[width=0.28\linewidth]{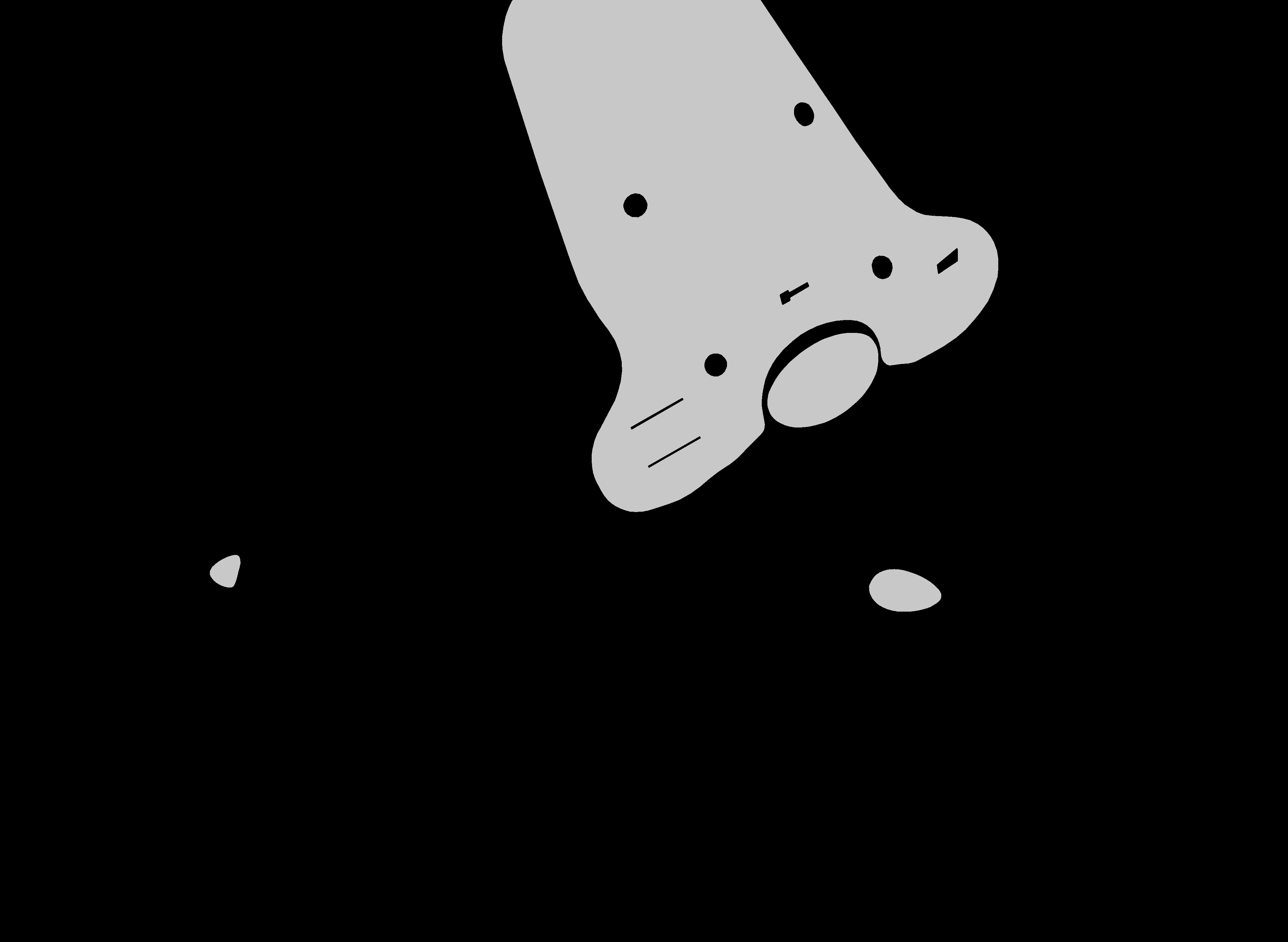} &
         \includegraphics[width=0.28\linewidth]{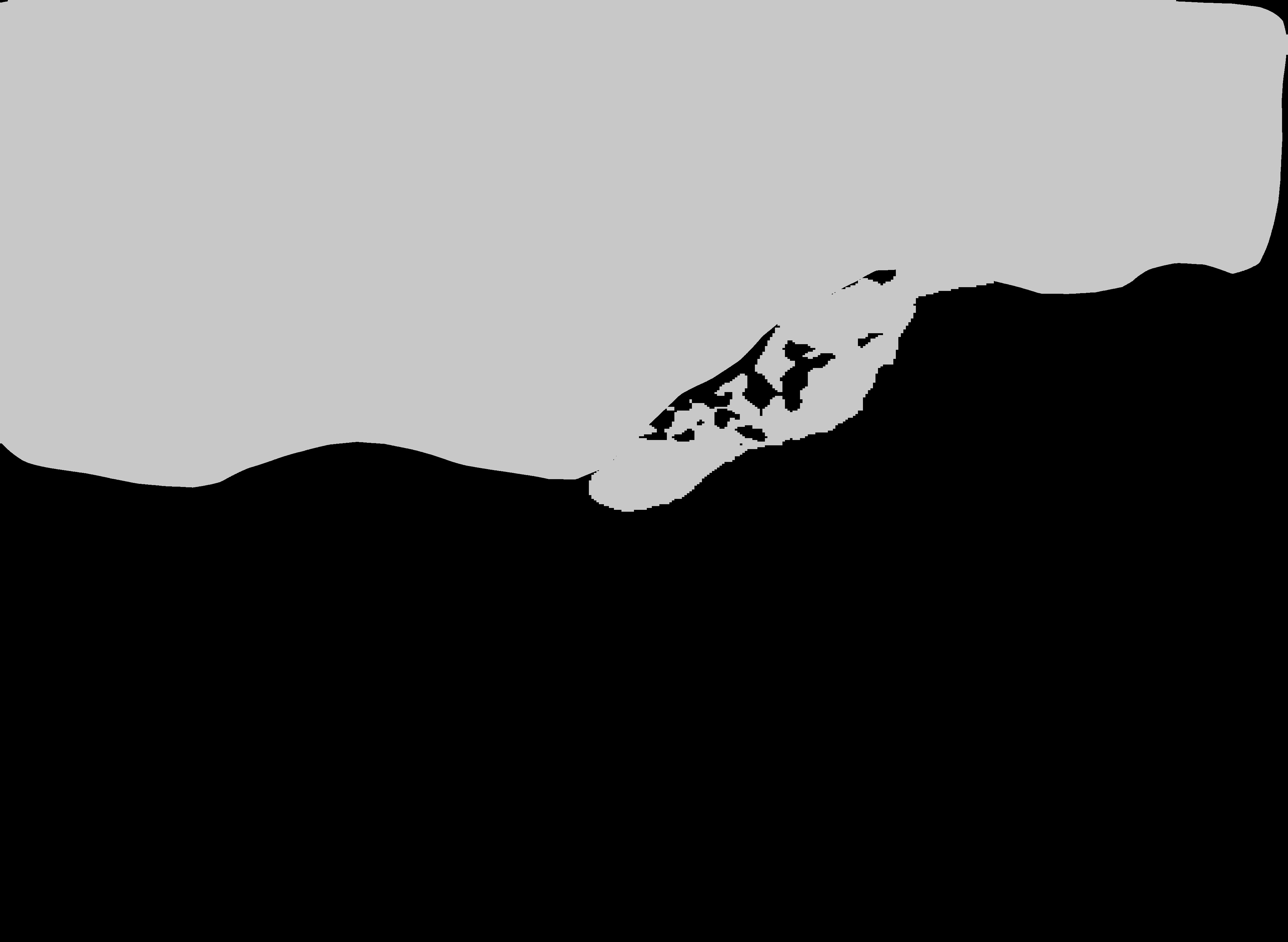} \\
         \scriptsize \textbf{\textit{Base}} &
         \scriptsize \textbf{\textit{Virtual Depth (GT)}} & 
         \scriptsize \textbf{\textit{Virtual Depth (Proxy)}} \\
         \includegraphics[width=0.28\linewidth]{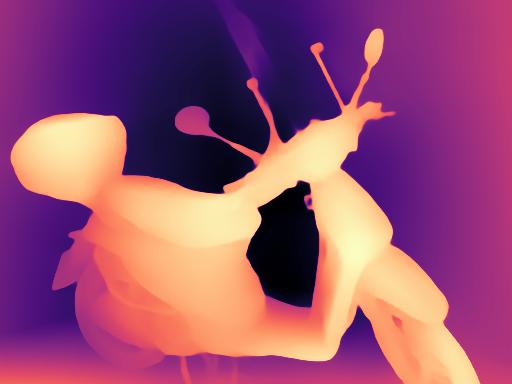} & 
         \includegraphics[width=0.28\linewidth]{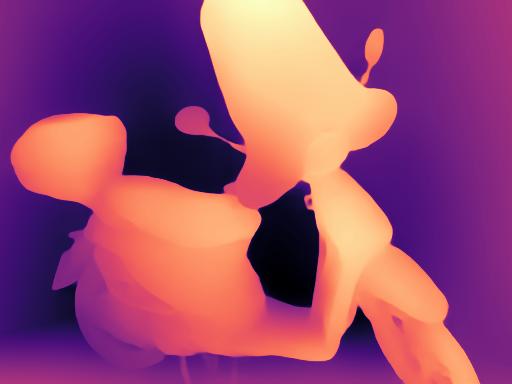} & 
         \includegraphics[width=0.28\linewidth]{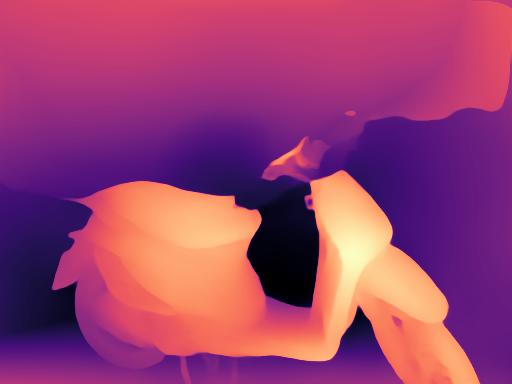} \\
    \end{tabular}    
    \caption{\textbf{Virtual Depth Qualitatives -- GT vs Proxy.} From left to right - Top: RGB, ground-truth and proxy segmentations; Bottom: prediction with DPT on the RGB image prediction with DPT on the median of five predictions by in-painting with either the ground truth or semantic proxy masks on Booster.}
    \label{fig:ablation_semantic_proxy_mono}
\end{figure}

\subsection{Monocular Depth Estimation.}

\textbf{Number of In-Paintings.}
We investigate the quality of the virtual depth labels by varying $N$.
When using $N=1$ we generate a single in-painted image that is forwarded to the monocular network, while with $N=5$ we generate virtual depths from 5 masked images with different colors which are then aggregated by selecting the pixel-wise depth median.
In Tab. \ref{tab:ablation_mask_generalization}, we report the accuracy of depth maps produced by the two strategies, together with those of the \textit{Base} architectures, i.e. without applying any in-painting strategy.
Firstly, with both MiDaS and DPT, both in-painting strategies obtain virtual depths that are much more accurate for ToM regions w.r.t. the \textit{Base} architecture.
Secondly, \textit{N=5} maps yield slightly better results in most metrics, especially when looking at DPT performance.
We ascribe it to the higher robustness of the second strategy.
For the remaining experiments, we fix $N=5$ as we did not observed any further improvement with larger values. 

\textbf{Fine-tuning Results (GT Segmentation).}
In Tab. \ref{tab:finetuning_mono}, we report results on the Booster train set, obtained after fine-tuning MiDaS and DPT on all available data from Trans10K and MSD. In the \textit{Base} row, we report the results of the network using the officially released weights without any further training, and we compare with those in row \textit{Ft. Virtual Depth}, i.e., the results of our method.
We notice that the accuracy on \textit{All} pixels is improved with our approach. In particular, we achieve a significant boost in performances for ToM surfaces, of 4.37, 5.72, 8.97, 11.12 and 11.57\%, 28.67mm, 0.03\%, 35.51mm for MiDaS \cite{Ranftl2022}, and 3.91, 7.19, 11.25, 16.8 and 16.97\%, 42.46mm, 0.04\%, 53.255mm for DPT\cite{Ranftl2022} in the $\delta_{1.25}$, $\delta_{1.20}$, $\delta_{1.15}$, $\delta_{1.10}$, $\delta_{1.05}$, MAE, Abs.Rel, and RMSE, respectively. 
We highlight that, after fine-tuning, the accuracy on \textit{ToM} is only slightly worse than on \textit{Other}.
Moreover, class \textit{Other} metrics are also slightly better, probably because of the enhanced features extracted by the network, which has a better understanding of the scene context.
Finally, we have reported in \textit{Ft. Base} the fine-tuning results obtained by self-training the networks on their own predictions without any in-painting strategy. As expected, without the appropriate virtual depth labels, the networks cannot effectively learn from the new dataset, yielding results comparable to the \textit{Base} architecture.
Experiments on additional datasets are in the supplement.

\begin{table}[t]
\centering
\setlength{\tabcolsep}{3pt}
\scalebox{0.5}{
\begin{tabular}{cccc}
 & RAFT-Stereo \cite{lipson2021raft} & CREStereo \cite{li2022practical} \\
 \begin{tabular}{ll}
 \toprule
 \\
 Category & Method \\
 \midrule
 \cellcolor{pink}All & Base \\
 \cellcolor{pink}All & Ft. Base\\
 \cellcolor{pink}All & Ft. Virtual Depth\\
 \cellcolor{pink}All & Ft. Merged\\
 \bottomrule
 \toprule
 \cellcolor{blue!25}ToM & Base \\
 \cellcolor{blue!25}ToM & Ft. Base\\
 \cellcolor{blue!25}ToM & Ft. Virtual Depth\\
 \cellcolor{blue!25}ToM & Ft. Merged \\
 \bottomrule
 \toprule
 \cellcolor{YellowOrange}Other & Base \\
 \cellcolor{YellowOrange}Other & Ft. Base\\
 \cellcolor{YellowOrange}Other & Ft. Virtual Depth\\
 \cellcolor{YellowOrange}Other & Ft. Merged \\
 \bottomrule
 \end{tabular}
 &
 \begin{tabular}{rrrr | rr}
 \toprule
 bad-2 & bad-4 & bad-6 & bad-8 & MAE & RMSE \\
 $\downarrow$ (\%) & $\downarrow$ (\%) & $\downarrow$ (\%) & $\downarrow$ (\%) & $\downarrow$ (px) & $\downarrow$ (px) \\
 \midrule
17.42 & 13.49 & 11.59 & 10.11 & 4.07 & 8.63 \\
18.42 & 14.02 & 12.00 & 10.47 & 4.32 & 8.91  \\
16.80 & 12.35 & 9.99 & 8.09 & 2.60 & 6.04 \\
\textbf{14.68} & \textbf{9.63} & \textbf{7.36} & \textbf{5.58} & \textbf{1.95} & \textbf{4.58} \\
\bottomrule
 \toprule
56.77 & 44.38 & 38.43 & 33.31 & 13.45 & 16.56 \\
58.57 & 44.42 & 38.00 & 32.99 & 13.84 & 16.47 \\
57.08 & 42.89 & 34.74 & 27.62 & 8.48 & 10.75 \\
\textbf{47.54} & \textbf{30.55} & \textbf{22.81} & \textbf{16.62} & \textbf{5.83} & \textbf{7.43} \\
\bottomrule
 \toprule
8.48 & 5.74 & 4.52 & 3.83 & 1.58 & 3.79 \\
9.20 & 6.20 & 4.89 & 4.14 & 1.70 & 3.98 \\
7.97 & 5.04 & 3.82 & 3.16 & 1.19 & 3.21 \\
\textbf{7.14} & \textbf{4.09} & \textbf{2.88} & \textbf{2.23} & \textbf{0.96} & \textbf{2.64}  \\
\bottomrule
 \end{tabular}
 &
 \begin{tabular}{rrrr | rr}
 \toprule
 bad-2 & bad-4 & bad-6 & bad-8 & MAE & RMSE \\
 $\downarrow$ (\%) & $\downarrow$ (\%) & $\downarrow$ (\%) & $\downarrow$ (\%) & $\downarrow$ (px) & $\downarrow$ (px) \\
 \midrule
15.13 & 10.70 & 8.91 & 7.57 & 3.15 & 7.40 \\
14.66 & 10.16 & 8.41 & 6.99 & 2.88 & 6.72 \\
18.83 & 14.28 & 12.33 & 10.80 & 5.09 & 9.78 \\
\textbf{10.85} & \textbf{6.11} & \textbf{4.39} & \textbf{3.12} & \textbf{1.51} & \textbf{3.62} \\
\bottomrule
 \toprule
51.83 & 37.88 & 32.86 & 28.19 & 12.42 & 15.60 \\
49.89 & 34.89 & 29.40 & 24.63 & 10.60 & 13.43 \\
49.51 & 35.14 & 29.60 & 24.57 & 11.94 & 13.85 \\
\textbf{36.90} & \textbf{20.75} & \textbf{15.38} & \textbf{10.65} & \textbf{5.02} & \textbf{6.69} \\
\bottomrule
 \toprule
8.11 & 4.83 & 3.50 & 2.78 & 1.14 & 2.77 \\
7.01 & 4.20 & 3.15 & 2.53 & 1.00 & 2.56 \\
11.44 & 8.17 & 6.96 & 6.25 & 2.72 & 5.49 \\
\textbf{5.27} & \textbf{2.72} & \textbf{1.81} & \textbf{1.34} & \textbf{0.68} & \textbf{1.69} \\
\bottomrule
 \end{tabular}
 \end{tabular}
 }
 \caption{\textbf{Stereo networks fine-tuning -- ground truth segmentation.}
 Results on Booster test set at quarter resolution. All models start from the official weights \cite{lipson2021raft,li2022practical} and are fine-tuned according to different strategies. Best results in \textbf{bold}.}\vspace{-0.5cm}
 \label{tab:stereo}
\end{table}

\begin{figure}[t]
    \centering
    \setlength{\tabcolsep}{1pt}
    \begin{tabular}{ccc}
         \scriptsize \textbf{\textit{RGB Left}} & 
         \scriptsize \textbf{\textit{Mask GT}} & 
         \scriptsize \textbf{\textit{Virtual Disparity (Stereo)}} \\
         \includegraphics[width=0.280\linewidth]{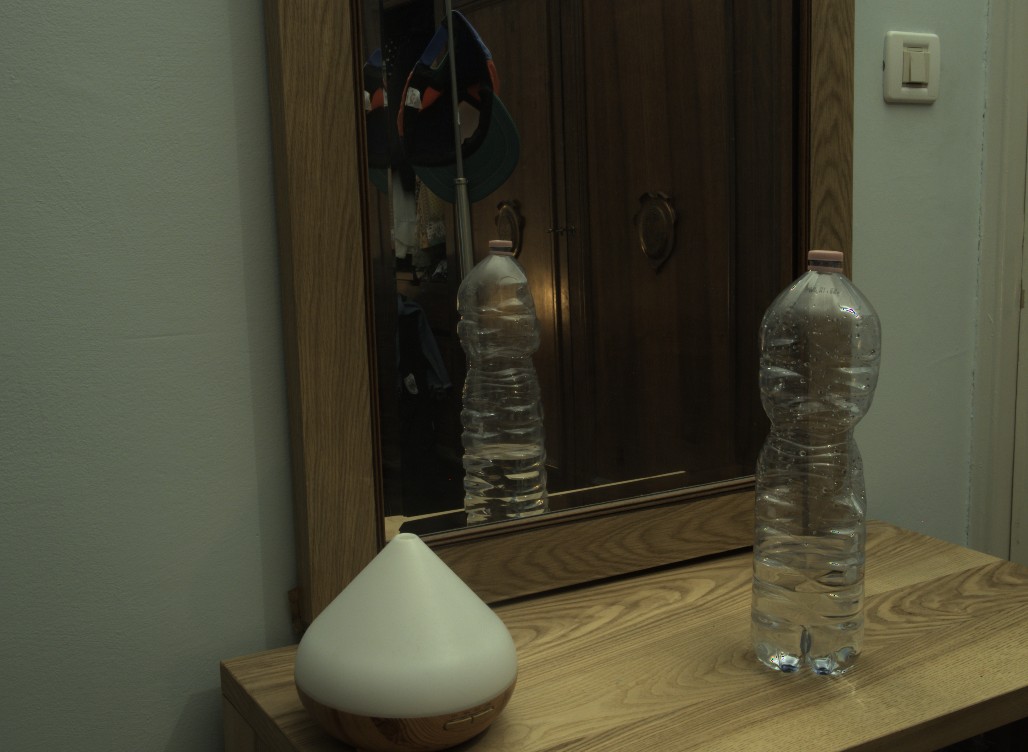} &
         \includegraphics[width=0.280\linewidth]{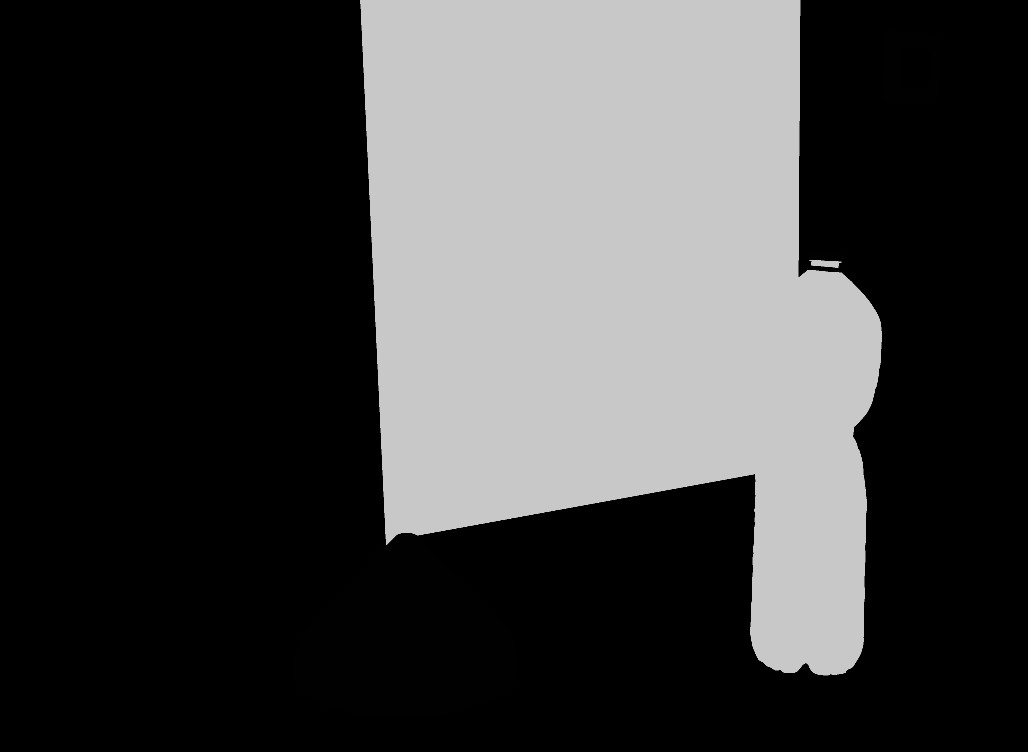} &
         \includegraphics[width=0.280\linewidth]{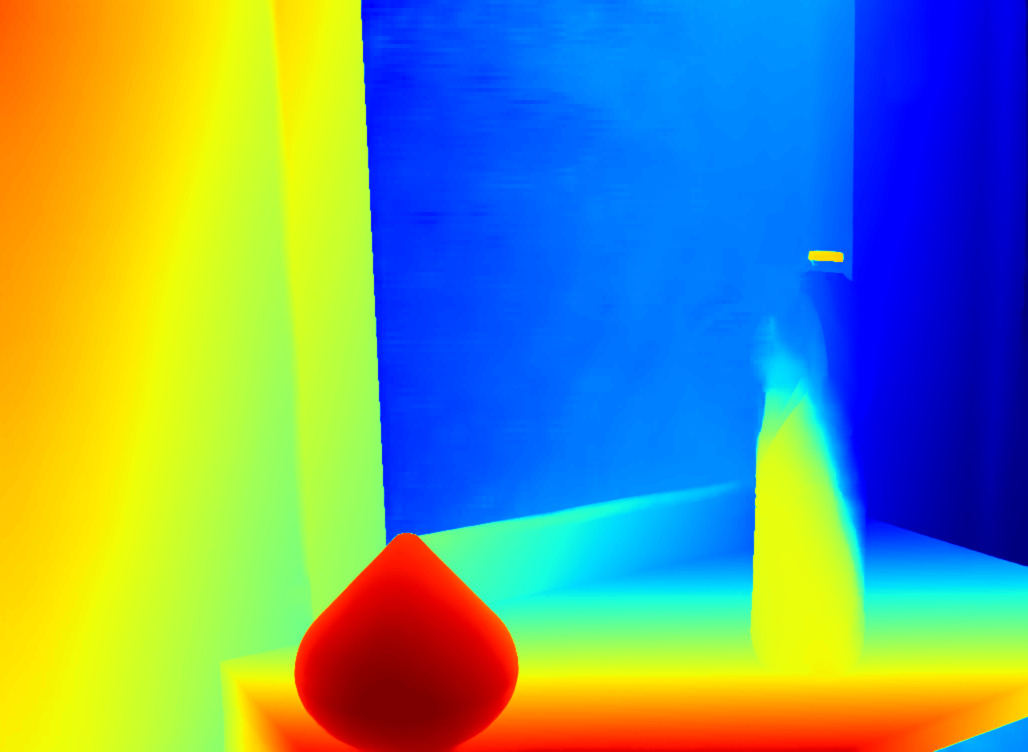}
         \\
         \scriptsize \textbf{\textit{Virtual Depth (Mono)}} & 
         \scriptsize \textbf{\textit{Base Stereo}} & 
         \scriptsize \textbf{\textit{Merged}} \\
         \includegraphics[width=0.280\linewidth]{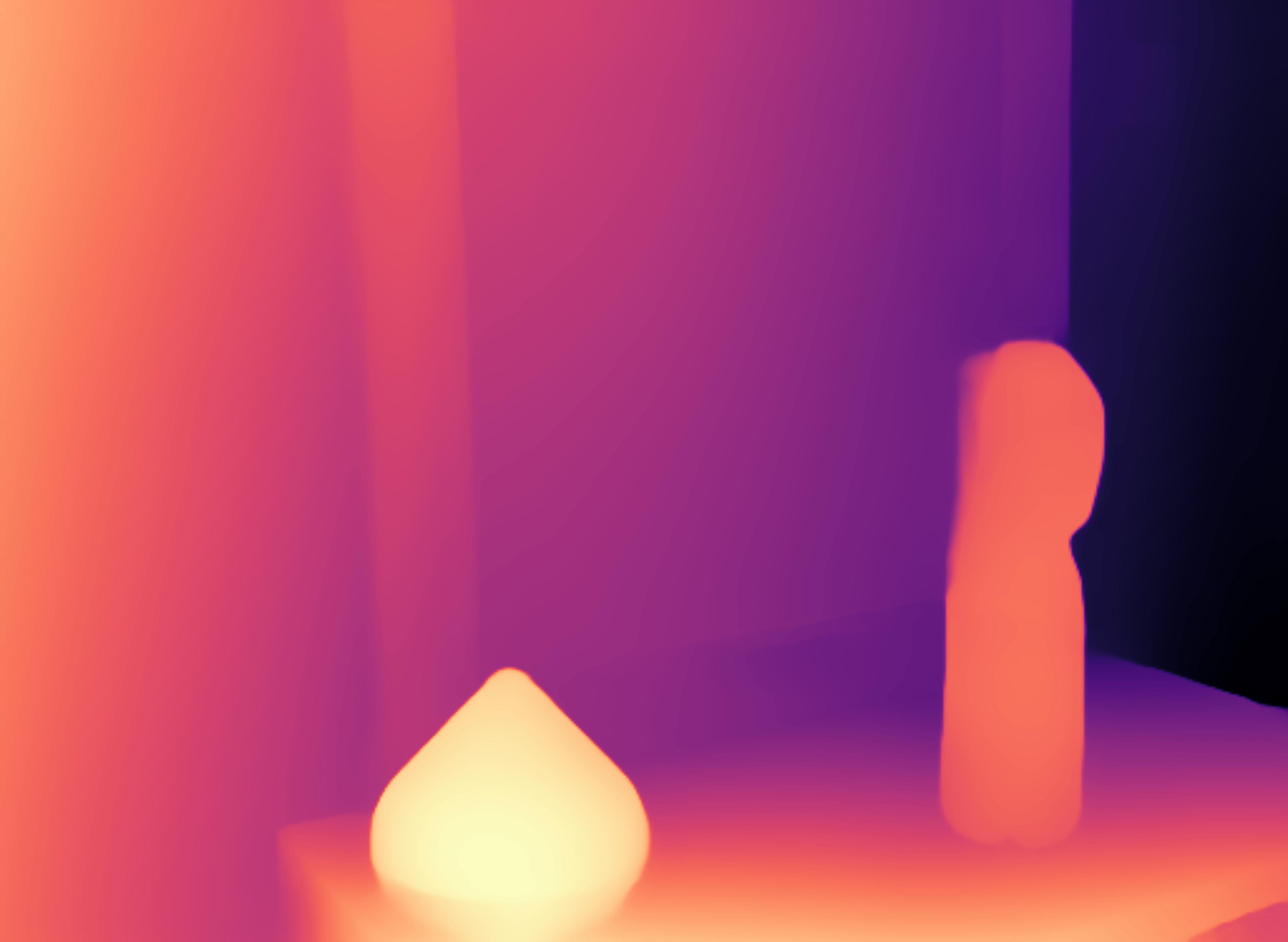} & 
         \includegraphics[width=0.280\linewidth]{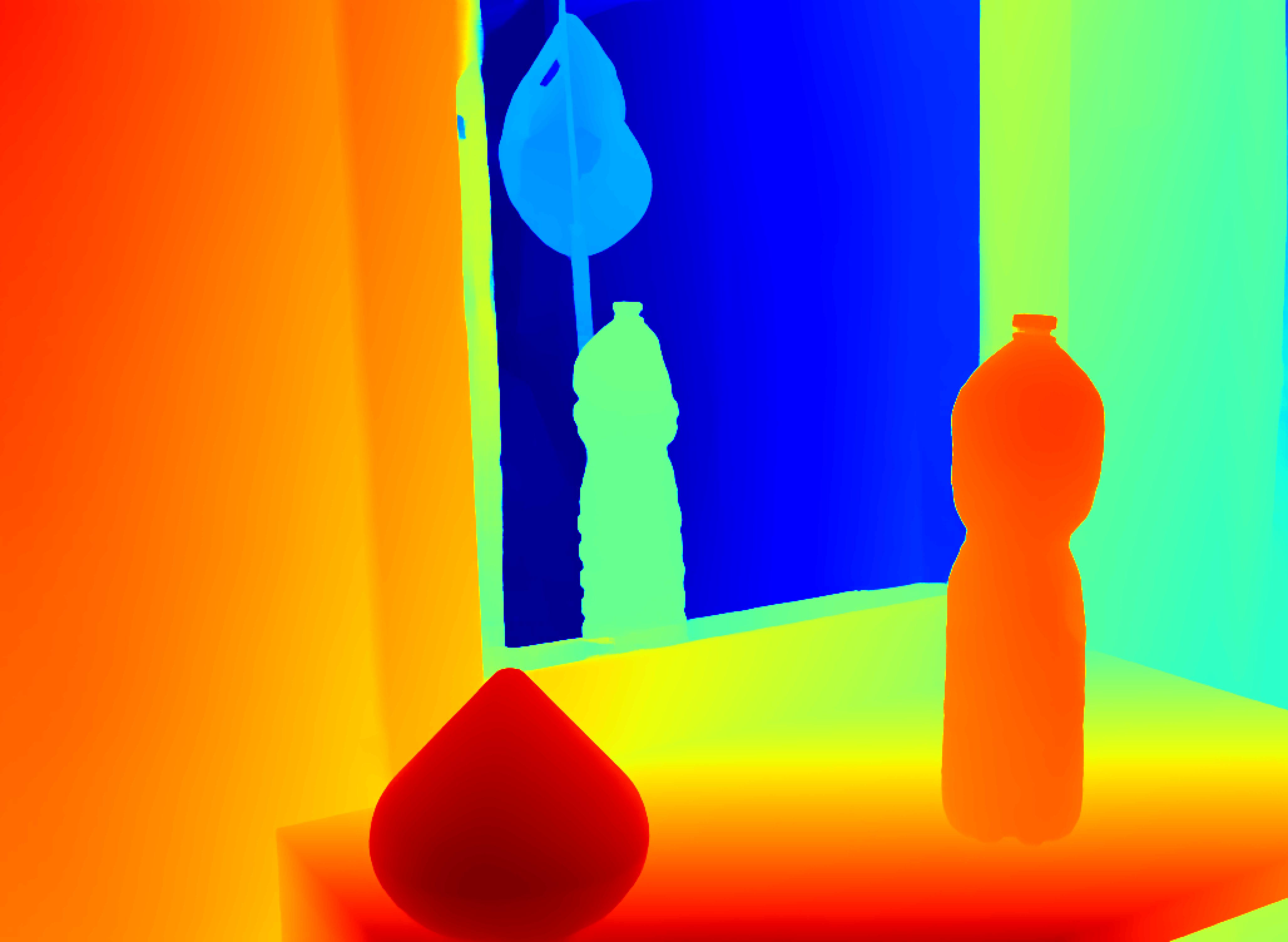} & 
         \includegraphics[width=0.280\linewidth]{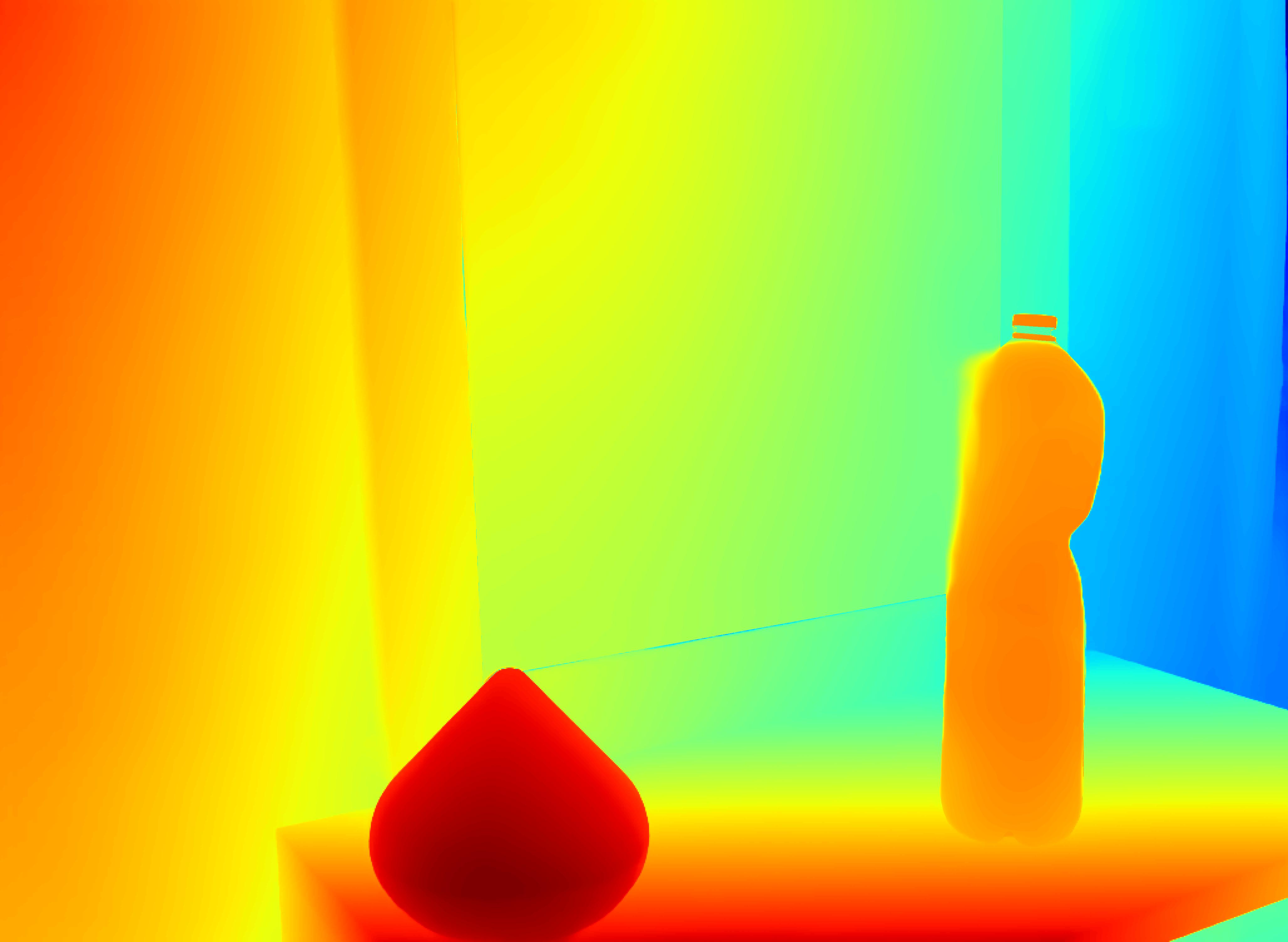} \\
    \end{tabular}
    
    \caption{\textbf{Qualitative comparison -- disparity virtual labels.} On top: left RGB image, ground truth segmentation mask, and virtual disparities by CREStereo processing masked stereo pairs. At bottom: depth by DPT on the in-painted left image, disparities by CREStereo, and their final merged labels.}
    \label{fig:proxy_merge}
\end{figure}

\textbf{Fine-tuning Results (Proxy Segmentation).}
Even though obtaining semantic labels is cheaper than collecting depth ground truths, using the predictions of a segmentation network as proxy semantic annotations would accelerate the dataset collection process. Thus, we investigate the impact of replacing manually annotated masks in our pipeline with the predictions of 
Trans2Seg\cite{xie2020segmenting} and MirrorNet\cite{Yang_2019_ICCV}, pre-trained on the training set of Trans10K and MSD, respectively, on the unseen test set of each dataset. We use weights made available by the authors. 
For a fair comparison, we also re-train again the models exploiting GT segmentations only on the test sets of the two datasets.
Tab. \ref{tab:proxy_vs_gt_mask} highlights that both models, using either GT - \textit{Ft. Virtual Depth (GT)} - or proxy segmentations  - \textit{Ft. Virtual Depth (Proxy)}, achieve much more accurate results compared to the \textit{Base} network. Interestingly, the two networks yield comparable results in the class \textit{Other}, while the one using GTs is slightly better than the other in the class \textit{ToM}, yet still comparable.
Finally, in row \textit{Virtual Depth (Proxy)}, we report the results of our in-painting methodology (i.e., without fine-tuning) but coloring pixels according to the proxy segmentations. We note that performances are even worse than the \textit{Base} method. Indeed, the segmentation network struggles to generalize to the Booster dataset, making the depth model incapable of estimating the correct values, e.g., due to some overextended in-painted ToM areas, as shown in Fig. \ref{fig:ablation_semantic_proxy_mono}.
Yet, depth networks, fine-tuned on the test set of MSD and Trans10K (row \textit{Ft. Virtual Depth (Proxy)}),
generalize properly on Booster.

\begin{table}[t]
    \centering
    \setlength{\tabcolsep}{3pt}
    \scalebox{0.49}{
    \begin{tabular}{cccc}
    & RAFT-Stereo \cite{lipson2021raft} & CREStereo \cite{li2022practical} \\
    \begin{tabular}{ll}
    \toprule
    \\
    Category & Method \\
    \midrule
    \cellcolor{pink}All & Base\\
    \cellcolor{pink}All & Ft. Merged (Proxy) \\
    \cellcolor{pink}All & Ft. Merged (GT)\\
    \bottomrule
    \toprule
    \cellcolor{blue!25}ToM & Base \\
    \cellcolor{blue!25}ToM & Ft. Merged (Proxy) \\
    \cellcolor{blue!25}ToM & Ft. Merged (GT) \\
    \bottomrule
    \toprule
    \cellcolor{YellowOrange}Other & Base \\
    \cellcolor{YellowOrange}Other & Ft. Merged (Proxy) \\
    \cellcolor{YellowOrange}Other & Ft. Merged (GT) \\
    \bottomrule
    \end{tabular}
    &
    \begin{tabular}{rrrr | rr}
    \toprule
    bad-2 & bad-4 & bad-6 & bad-8 & MAE & RMSE \\
    $\downarrow$ (\%) & $\downarrow$ (\%) & $\downarrow$ (\%) & $\downarrow$ (\%) & $\downarrow$ (px) & $\downarrow$ (px) \\
    \midrule
    17.42 & 13.49 & 11.59 & 10.11 & 4.07 & 8.63 \\
    19.56 & 13.53 & 10.53 & 8.28 & 2.51 & 5.37 \\
    \textbf{14.68} & \textbf{9.63} & \textbf{7.36} & \textbf{5.58} & \textbf{1.95} & \textbf{4.58} \\
    \bottomrule
    \toprule
    56.77 & 44.38 & 38.43 & 33.31 & 13.45 & 16.56 \\
    51.85 & 35.32 & 27.17 & 20.88 & 6.56 & 8.04 \\
    \textbf{47.54} & \textbf{30.55} & \textbf{22.81} & \textbf{16.62} & \textbf{5.83} & \textbf{7.43} \\
    \bottomrule
    \toprule
    8.48 & 5.74 & 4.52 & 3.83 & 1.58 & 3.79 \\
    11.95 & 7.36 & 5.31 & 4.15 & 1.41 & 3.41 \\
    \textbf{7.14} & \textbf{4.09} & \textbf{2.88} & \textbf{2.23} & \textbf{0.96} & \textbf{2.64}  \\
    \bottomrule
    \end{tabular}
    &
    \begin{tabular}{rrrr | rr}
    \toprule
    bad-2 & bad-4 & bad-6 & bad-8 & MAE & RMSE \\
    $\downarrow$ (\%) & $\downarrow$ (\%) & $\downarrow$ (\%) & $\downarrow$ (\%) & $\downarrow$ (px) & $\downarrow$ (px) \\
    \midrule
    15.13 & 10.70 & 8.91 & 7.57 & 3.15 & 7.40 \\
    12.51 & 7.33 & 5.16 & 3.56 & \textbf{1.38} & \textbf{3.30} \\
    \textbf{10.85} & \textbf{6.11} & \textbf{4.39} & \textbf{3.12} & {1.51} & {3.62} \\
    \bottomrule
    \toprule
    51.83 & 37.88 & 32.86 & 28.19 & 12.42 & 15.60 \\
    40.63 & 23.38 & 16.55 & 10.81 & \textbf{4.11} & \textbf{5.55} \\
    \textbf{36.90} & \textbf{20.75} & \textbf{15.38} & \textbf{10.65} & {5.02} & {6.69} \\
    \bottomrule
    \toprule
    8.11 & 4.83 & 3.50 & 2.78 & 1.14 & 2.77 \\
    6.51 & 3.65 & 2.52 & 1.88 & 0.82 & 2.00 \\
    \textbf{5.27} & \textbf{2.72} & \textbf{1.81} & \textbf{1.34} & \textbf{0.68} & \textbf{1.69} \\
    \bottomrule
    \end{tabular}
    \end{tabular}
    }
\caption{\textbf{Stereo networks fine-tuning -- ground truth vs proxy segmentation.}
Results on Booster test set at quarter resolution. All models start from the official weights \cite{lipson2021raft,li2022practical} and are fine-tuned according to different strategies. Best results in \textbf{bold}.}\vspace{-0.3cm}
\label{tab:stereo_proxy}
\end{table}
\begin{figure}[t]
    \centering
    \setlength{\tabcolsep}{1pt}
    
    \begin{tabular}{ccc}
         \scriptsize \textbf{\textit{RGB Left}} & 
         \scriptsize \textbf{\textit{Mask GT Left}} & 
         \scriptsize \textbf{\textit{Mask Proxy Left}} \\

         \includegraphics[width=0.28\linewidth]{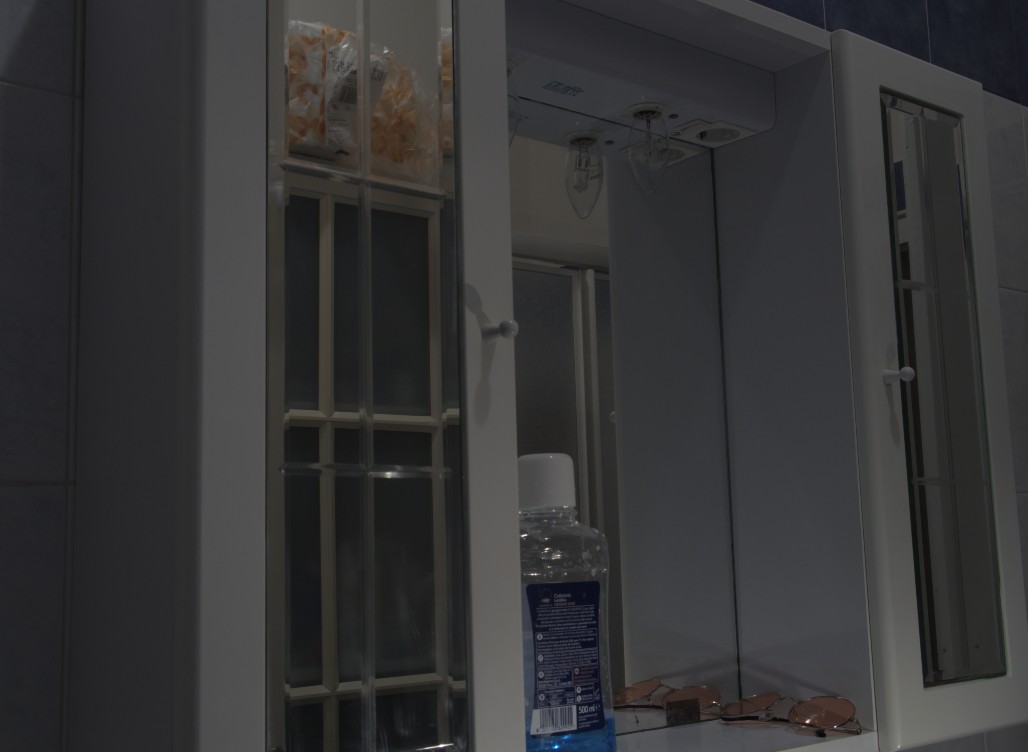} &
         \includegraphics[width=0.28\linewidth]{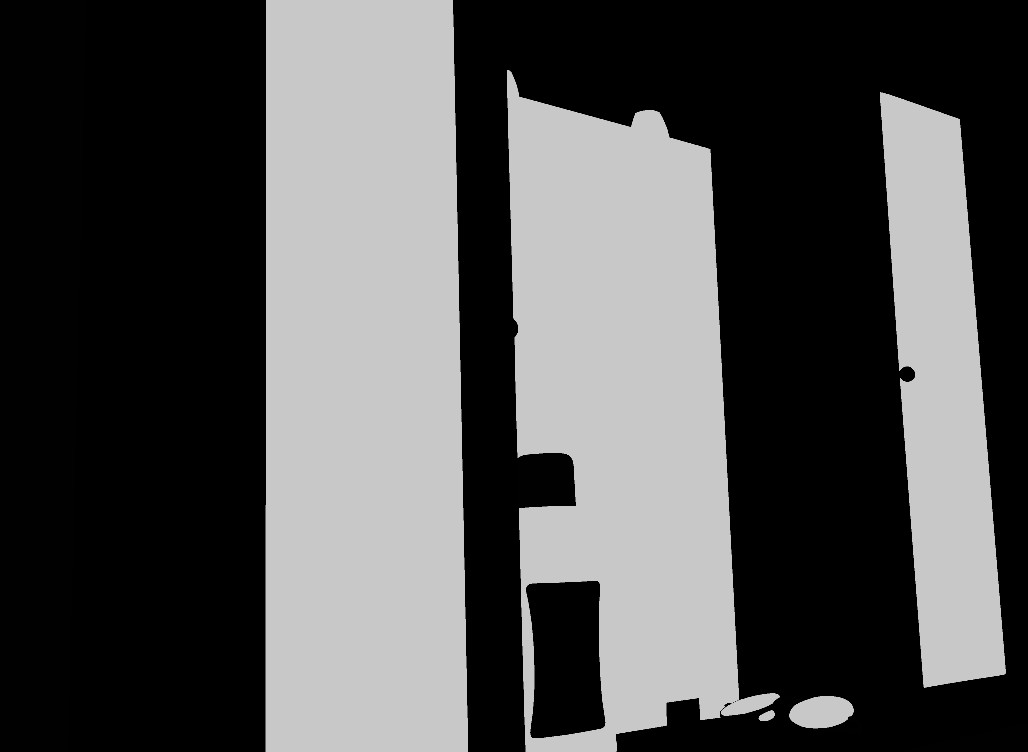} &
         \includegraphics[width=0.28\linewidth]{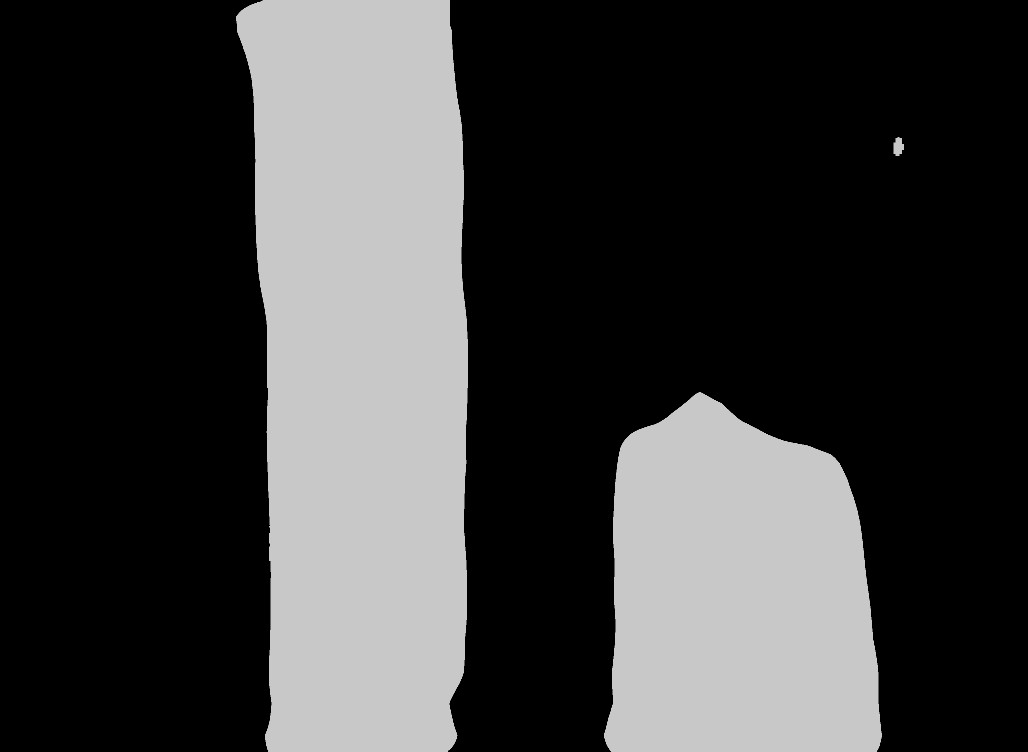} \\
         \scriptsize \textbf{\textit{Base}} & 
         \scriptsize \textbf{\textit{Merged (GT)}} & 
         \scriptsize \textbf{\textit{Merged (Proxy)}} \\
         \includegraphics[width=0.28\linewidth]{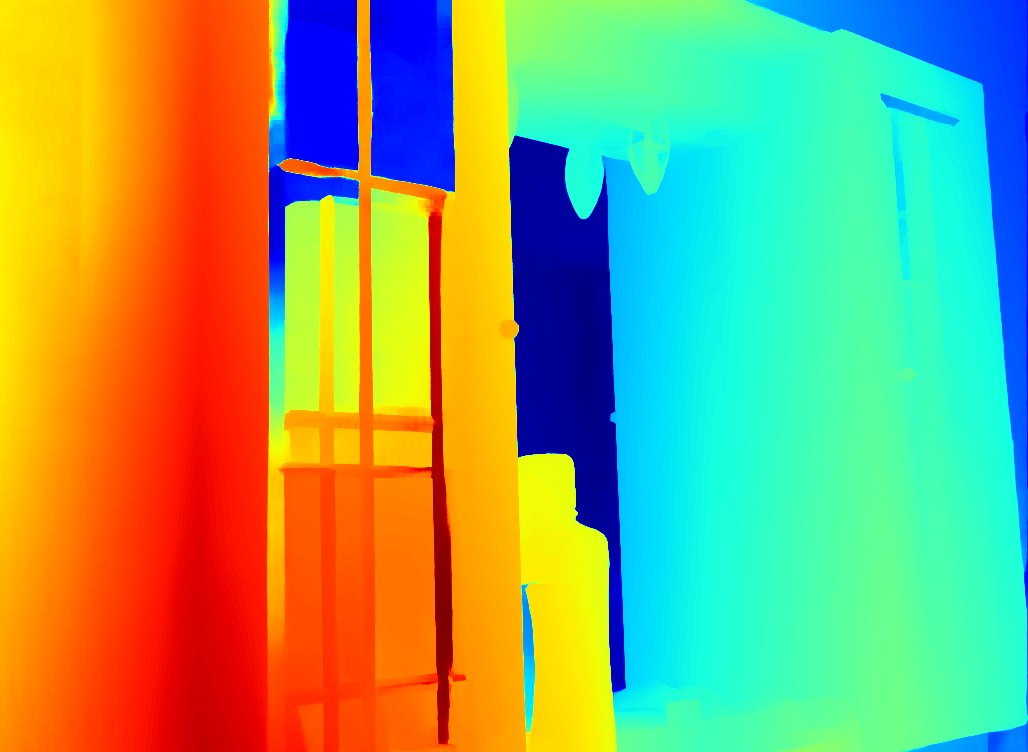} & 
         \includegraphics[width=0.28\linewidth]{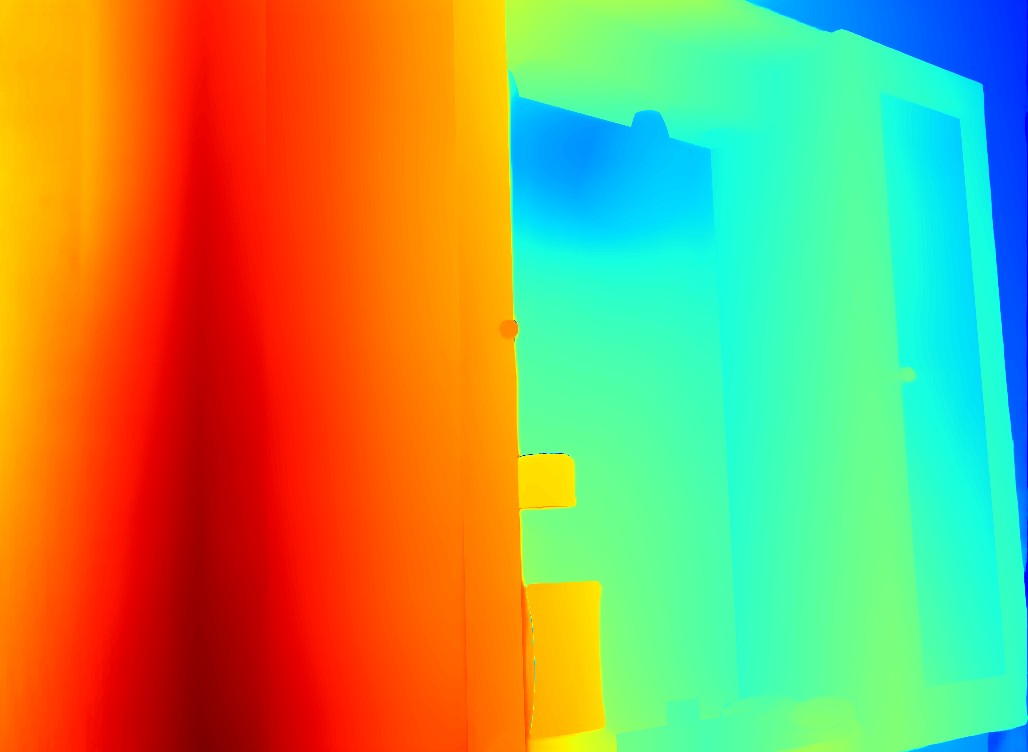} & 
         \includegraphics[width=0.28\linewidth]{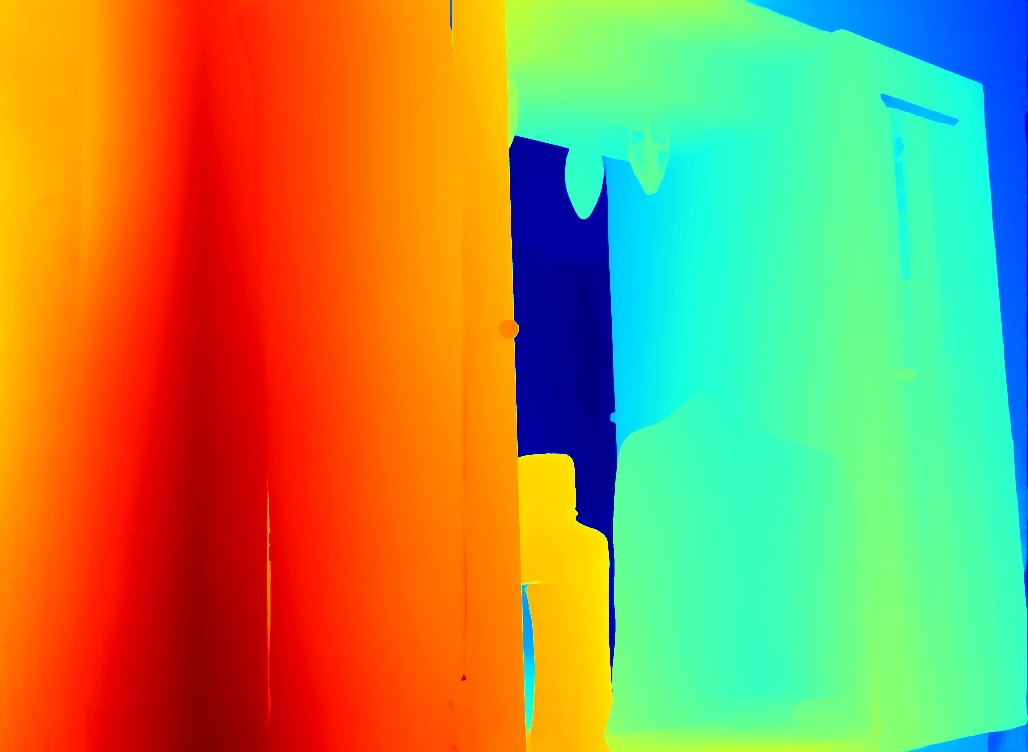} \\
    \end{tabular}    
    \caption{\textbf{Stereo depth merging with GT or Proxy semantic labels.} From left to right: RGB left image, ground-truth semantic mask, proxy semantic mask, prediction by CREStereo on the RGB images, and the final merged labels using either the GT or Proxy segmentation masks.}
    \label{fig:proxy_stereo}
\end{figure}

\begin{figure*}
    \centering
    \renewcommand{\tabcolsep}{5pt}
    \begin{tabular}{ccccccc}
    & \multicolumn{2}{c}{\scriptsize \textbf{\textit{MiDaS}}\cite{Ranftl2022}} & & 
        \multicolumn{2}{c}{\scriptsize \textbf{\textit{DPT}}\cite{Ranftl2021}} \\
        \scriptsize \textbf{\textit{RGB}} & 
        \scriptsize \textbf{\textit{Base}} & 
        \scriptsize \textbf{\textit{Ft (Proxy mask)}} & 
        \scriptsize \textbf{\textit{Ft (GT mask)}} & 
        \scriptsize \textbf{\textit{Base}} & 
        \scriptsize \textbf{\textit{Ft (Proxy mask)}} & 
        \scriptsize \textbf{\textit{Ft (GT mask)}} \\

        \includegraphics[width=0.1\textwidth]{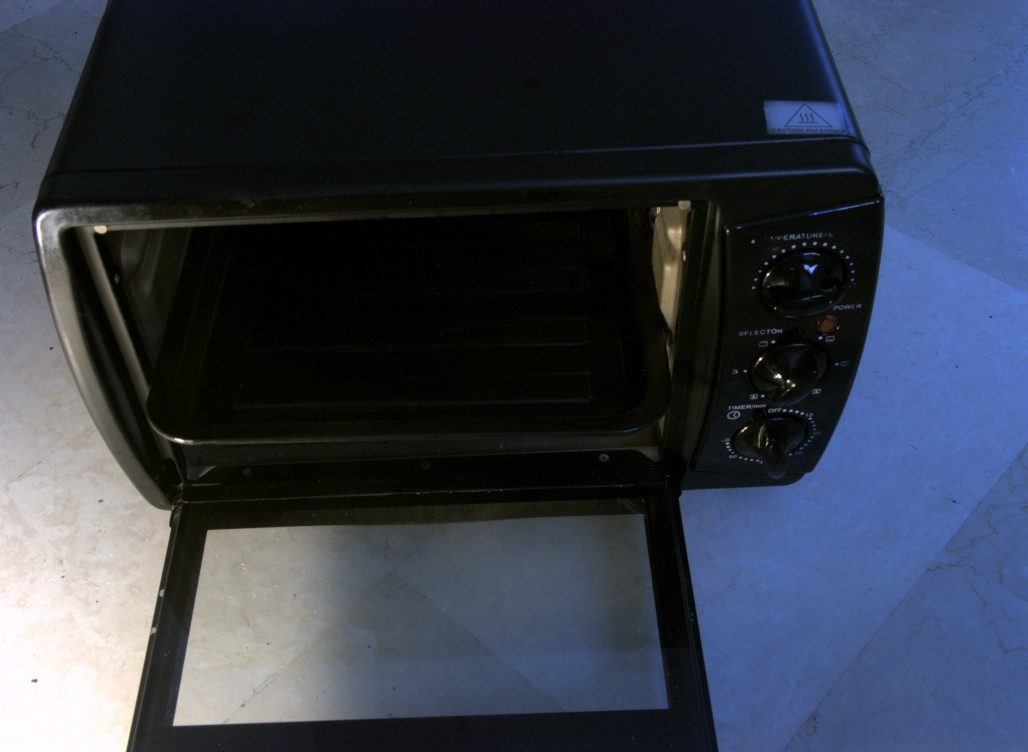} & 
        \includegraphics[width=0.1\textwidth]{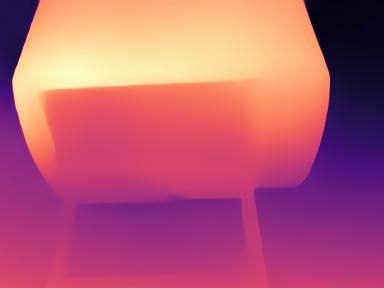} &
        \includegraphics[width=0.1\textwidth]{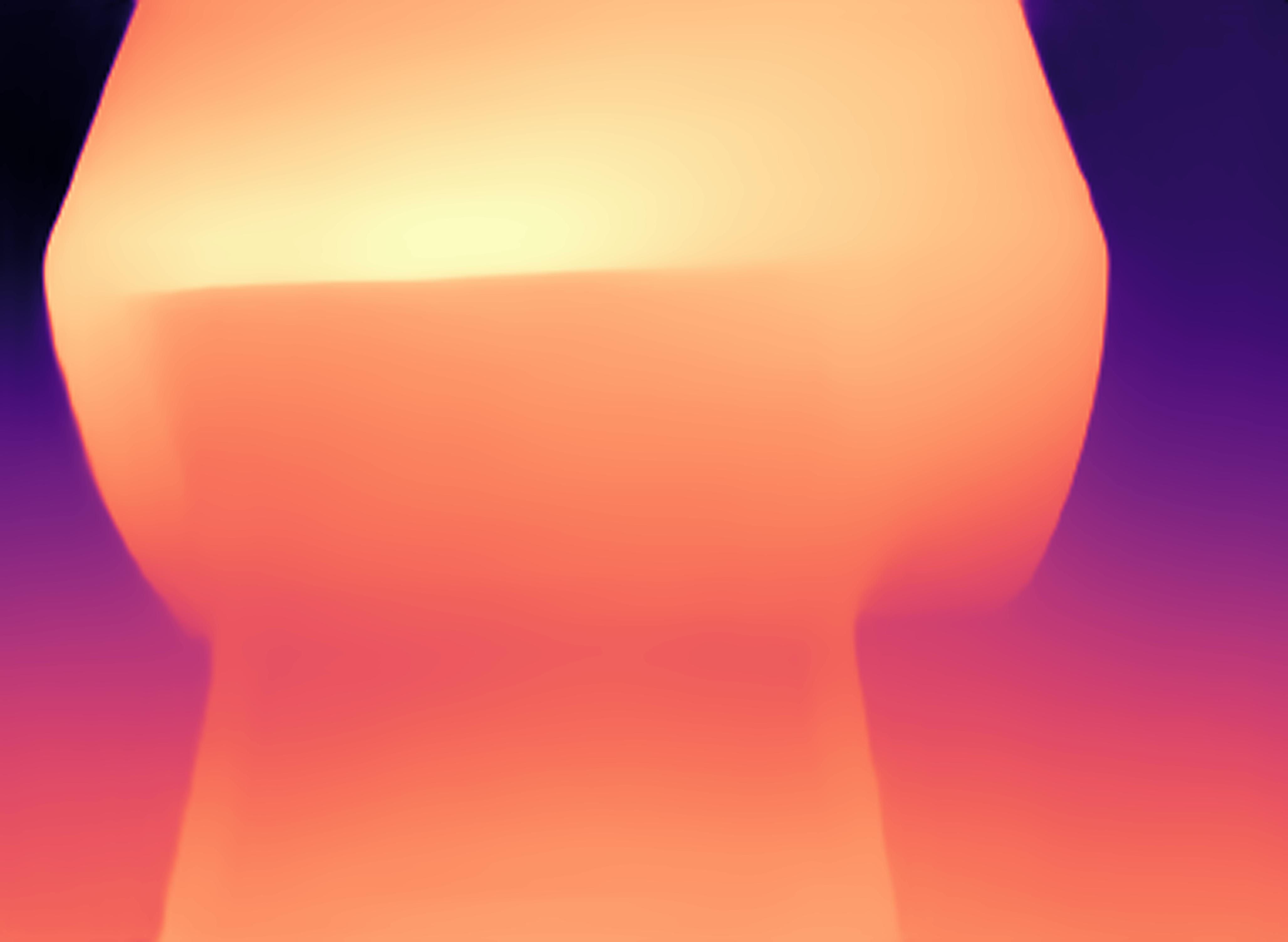} &
        \includegraphics[width=0.1\textwidth]{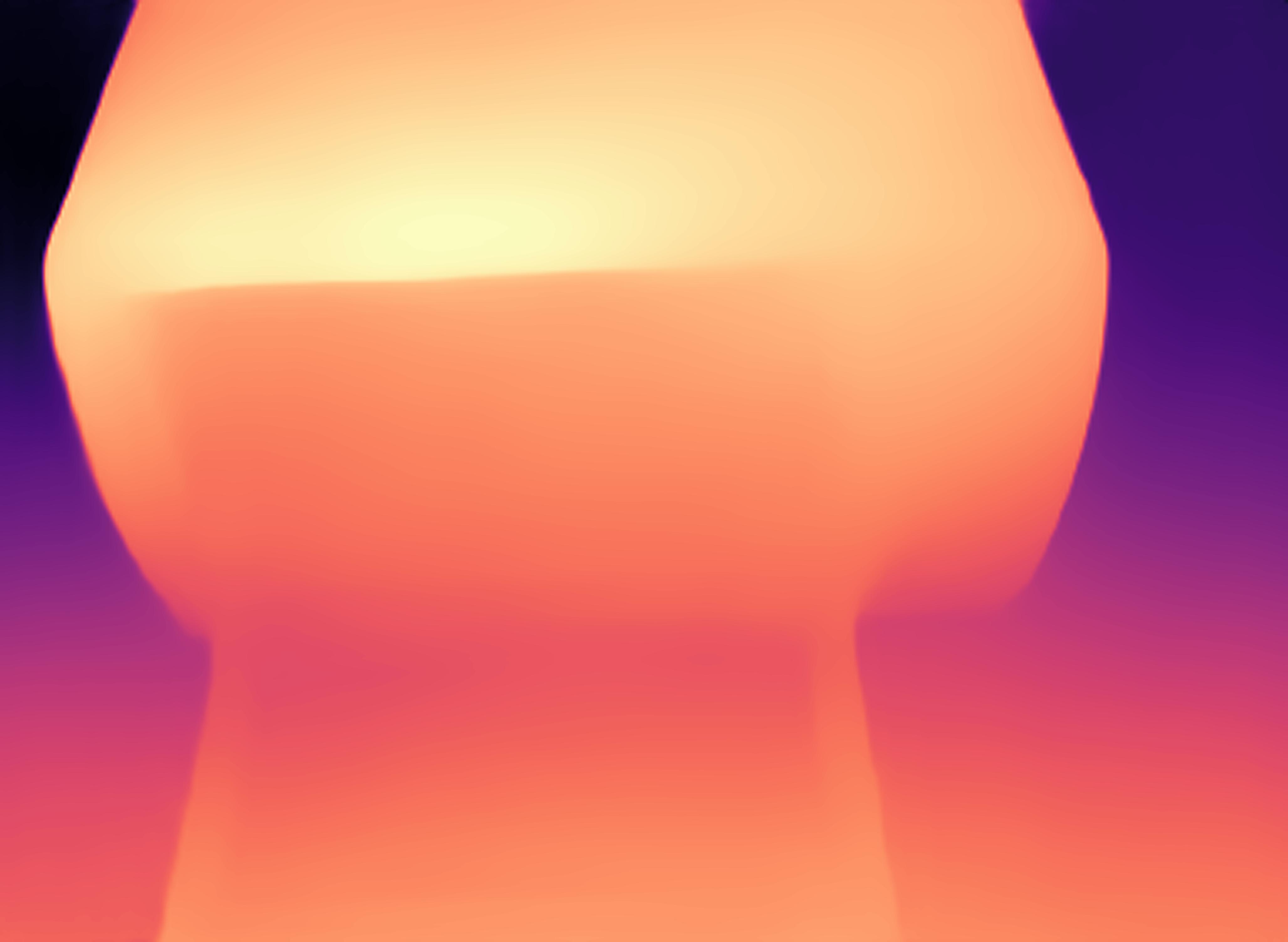} & 
        \includegraphics[width=0.1\textwidth]{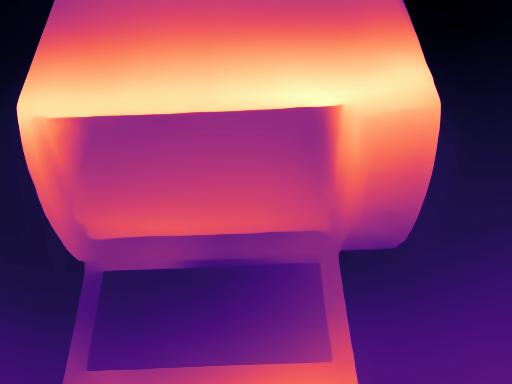} &
        \includegraphics[width=0.1\textwidth]{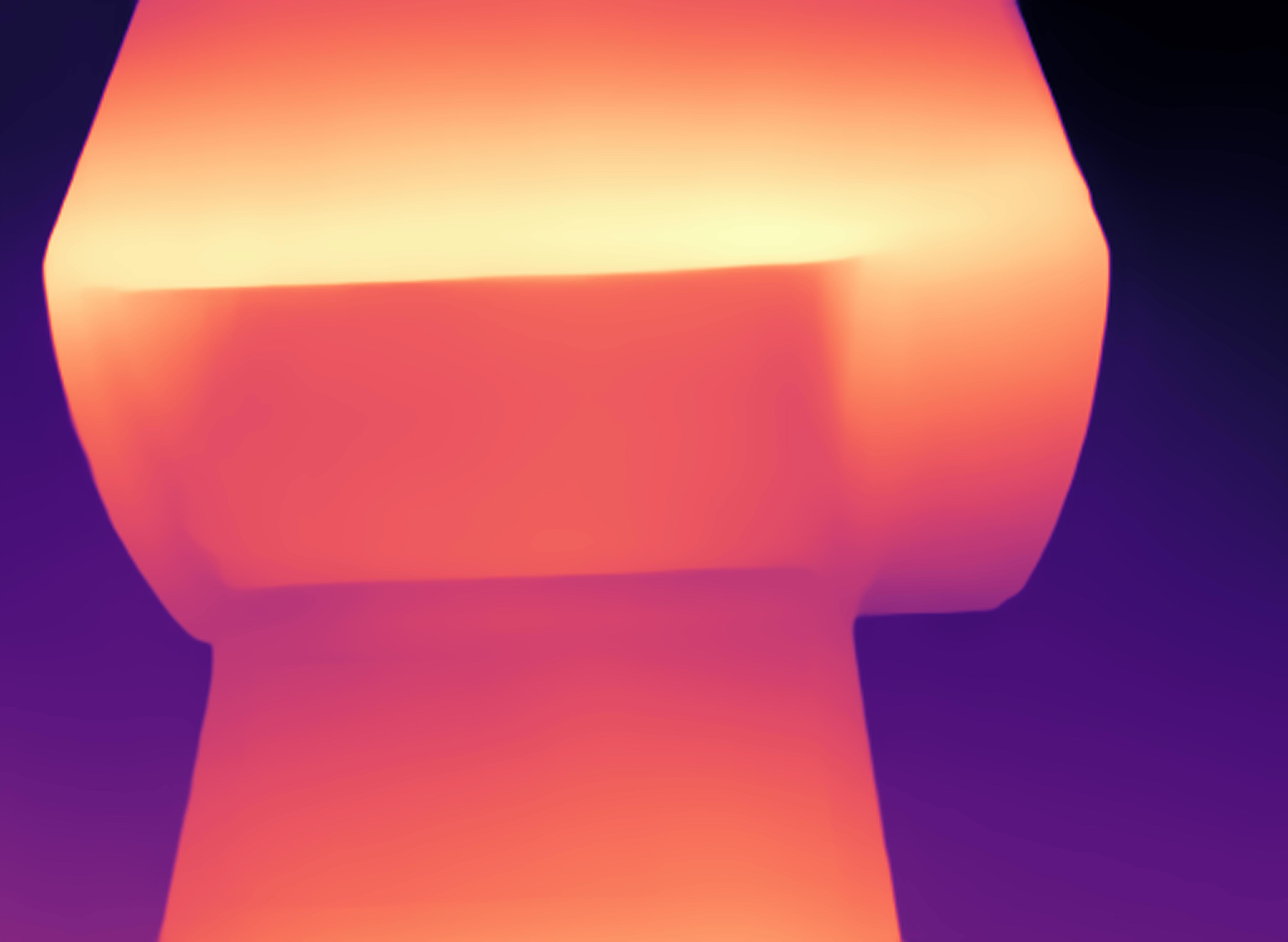} &
        \includegraphics[width=0.1\textwidth]{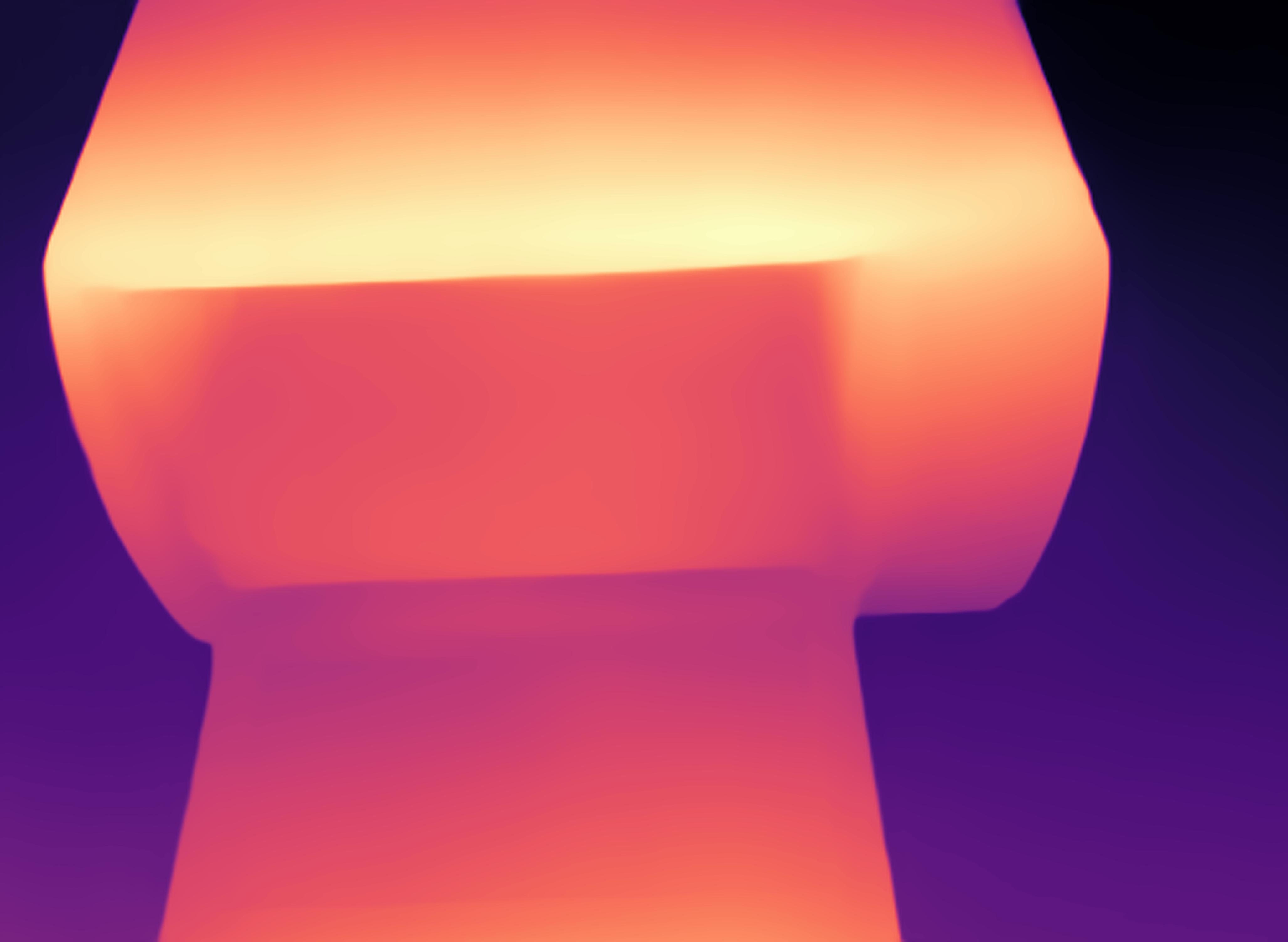} \\

        \includegraphics[width=0.1\textwidth]{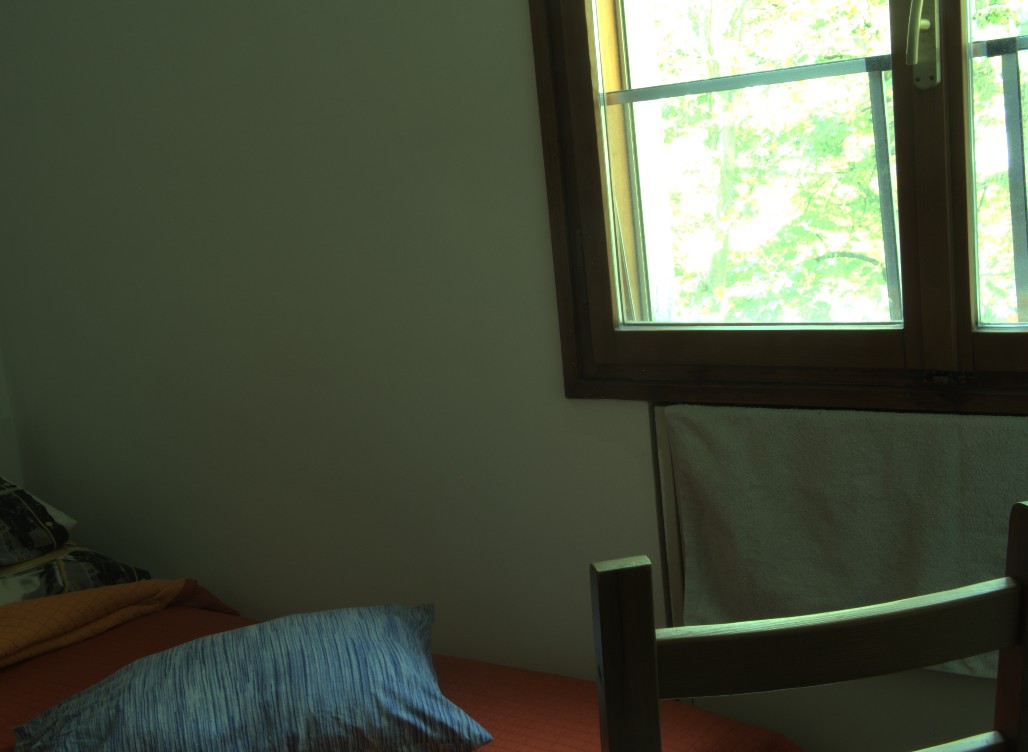} & 
        \includegraphics[width=0.1\textwidth]{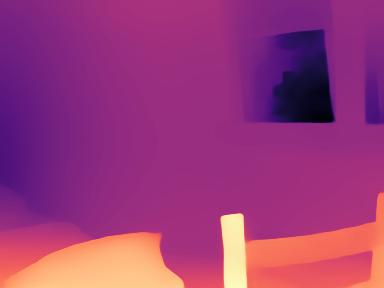} &
        \includegraphics[width=0.1\textwidth]{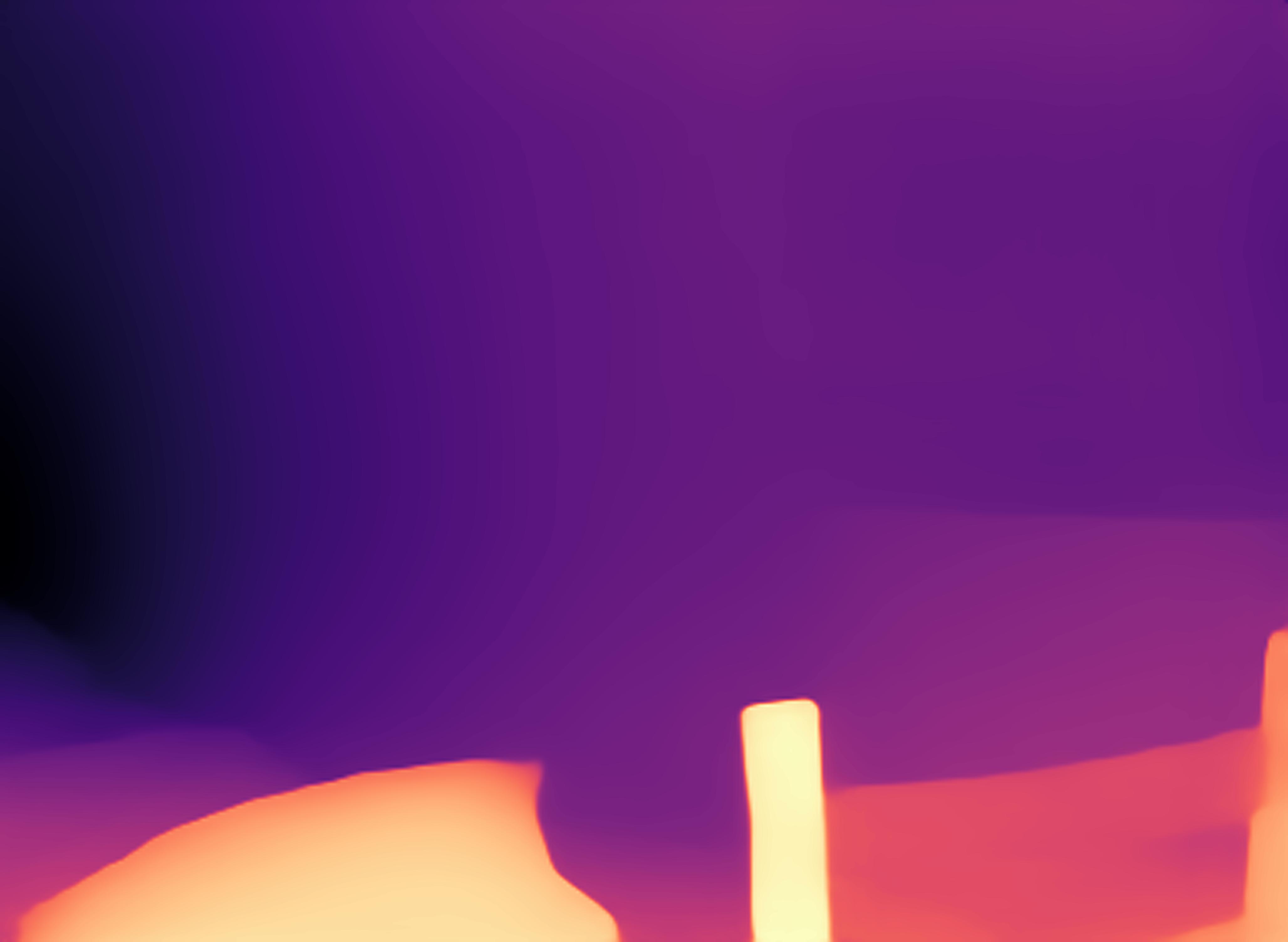} &
        \includegraphics[width=0.1\textwidth]{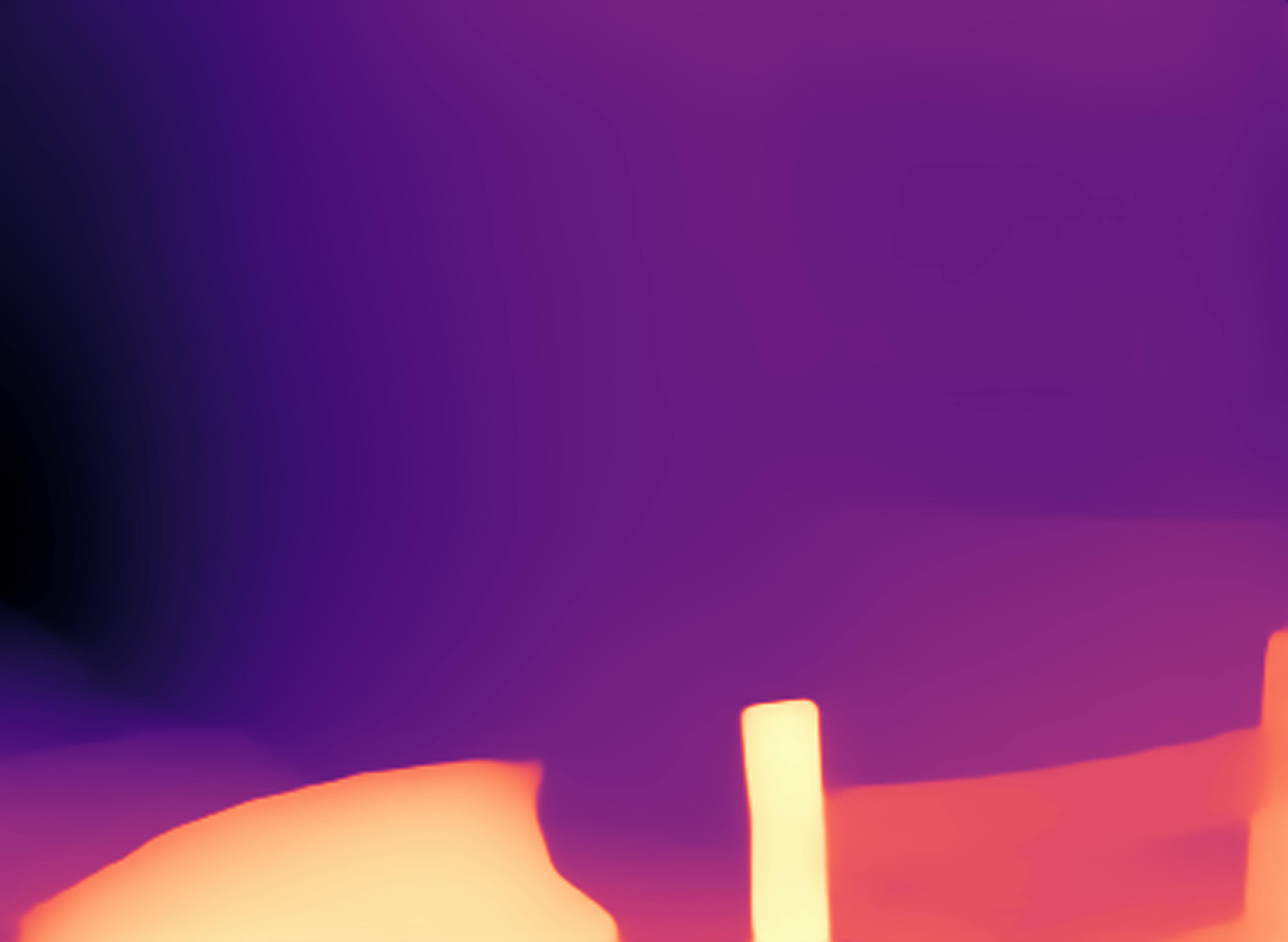} & 
        \includegraphics[width=0.1\textwidth]{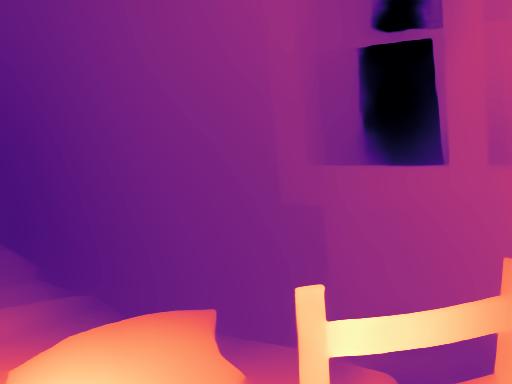} &
        \includegraphics[width=0.1\textwidth]{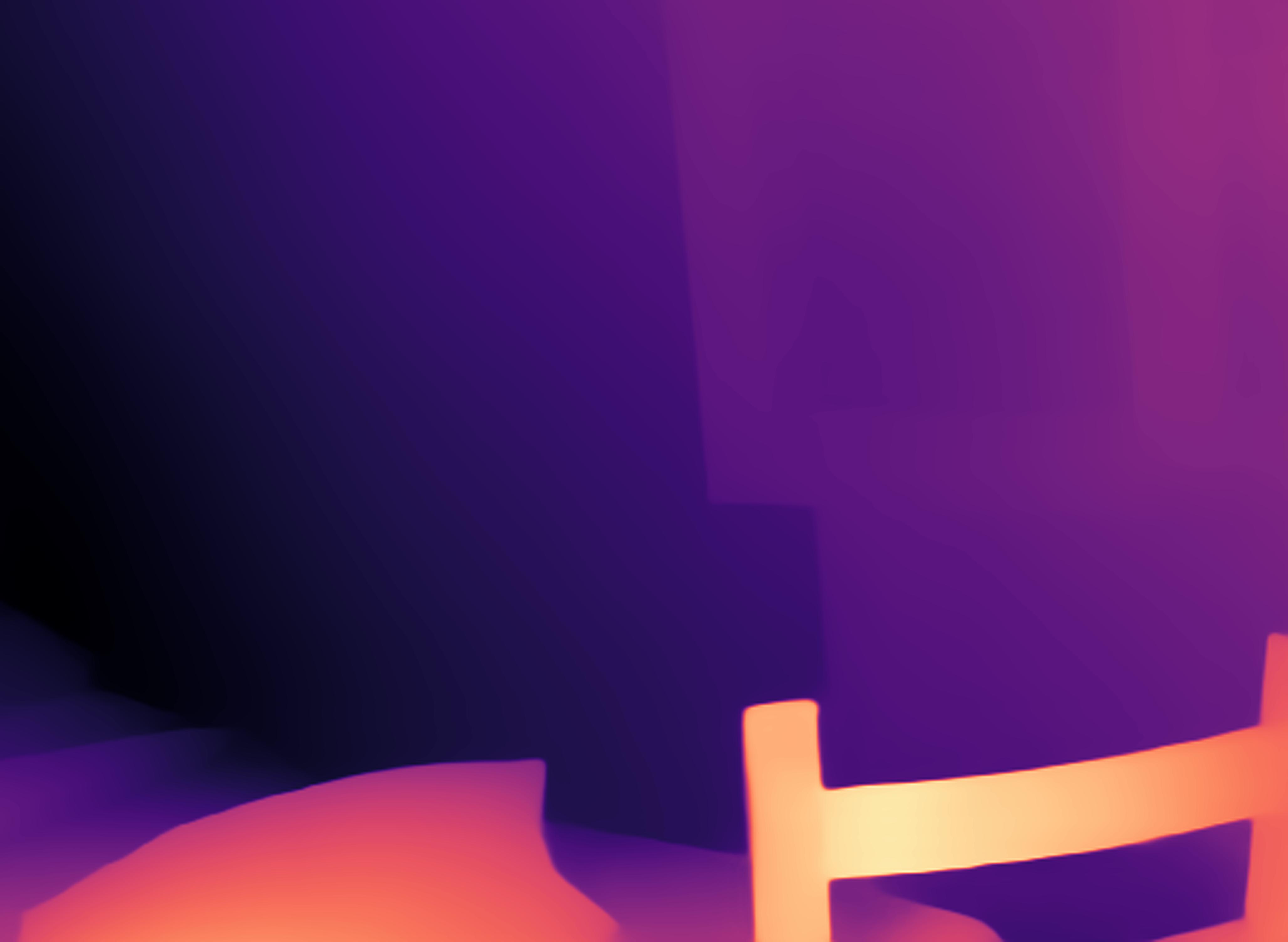} &
        \includegraphics[width=0.1\textwidth]{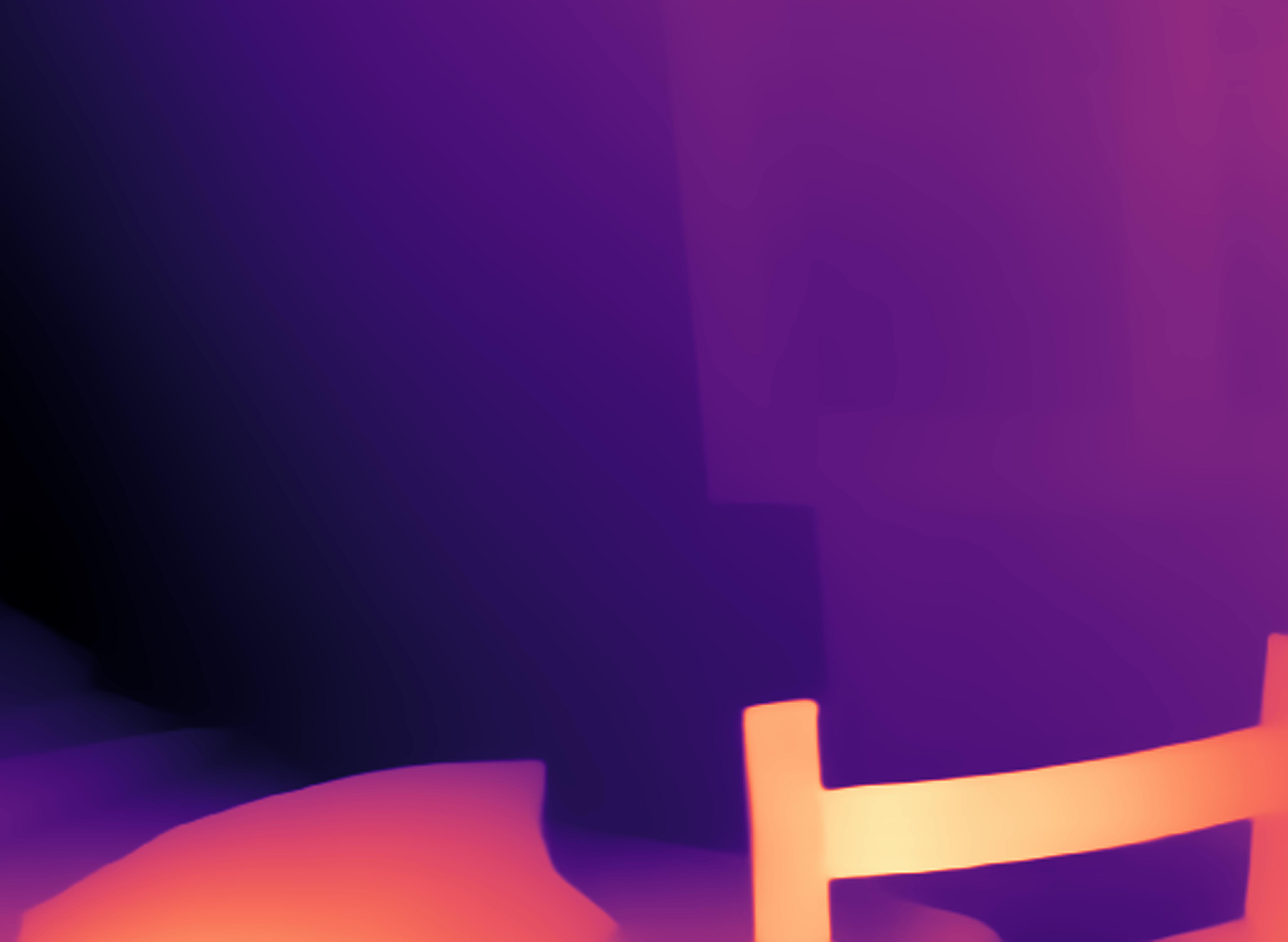} \\
        
        & \multicolumn{2}{c}{\scriptsize \textbf{\textit{RAFT-Stereo}}\cite{lipson2021raft}} & & 
        \multicolumn{2}{c}{\scriptsize \textbf{\textit{CREStereo}}\cite{li2022practical}} \\
        \scriptsize \textbf{\textit{RGB Left}} & 
        \scriptsize \textbf{\textit{Base}} & 
        \scriptsize \textbf{\textit{Ft (Proxy mask)}} & 
        \scriptsize \textbf{\textit{Ft (GT mask)}} & 
        \scriptsize \textbf{\textit{Base}} & 
        \scriptsize \textbf{\textit{Ft (Proxy mask)}} & 
        \scriptsize \textbf{\textit{Ft (GT mask)}} \\
        
        \includegraphics[width=0.1\textwidth]{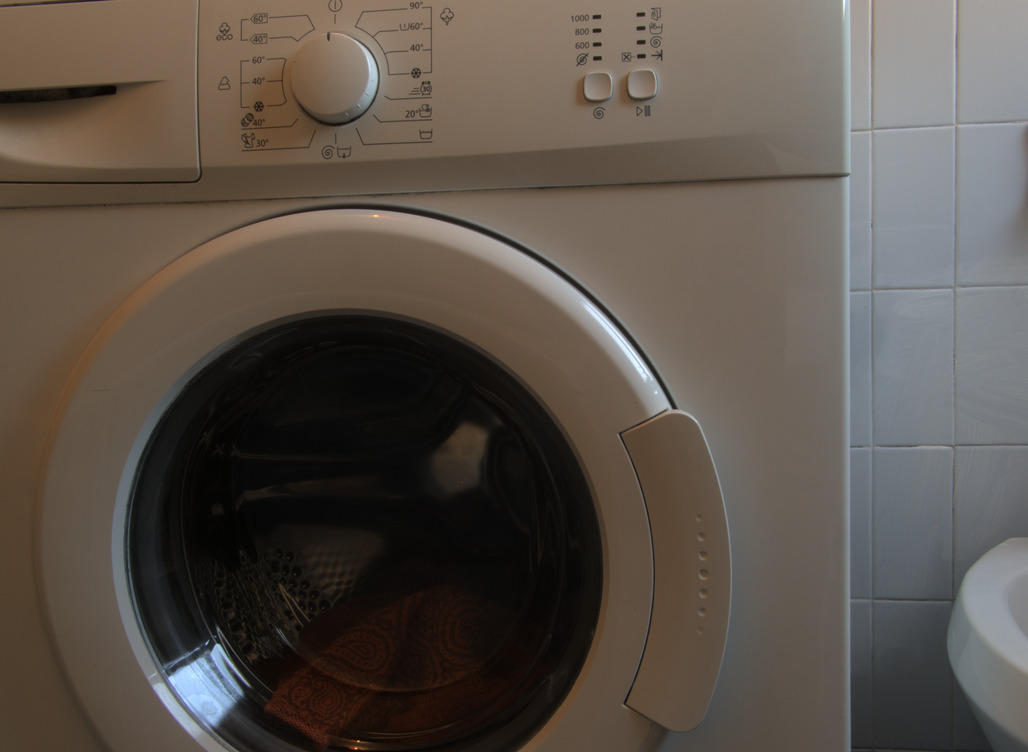} & 
        \includegraphics[width=0.1\textwidth]{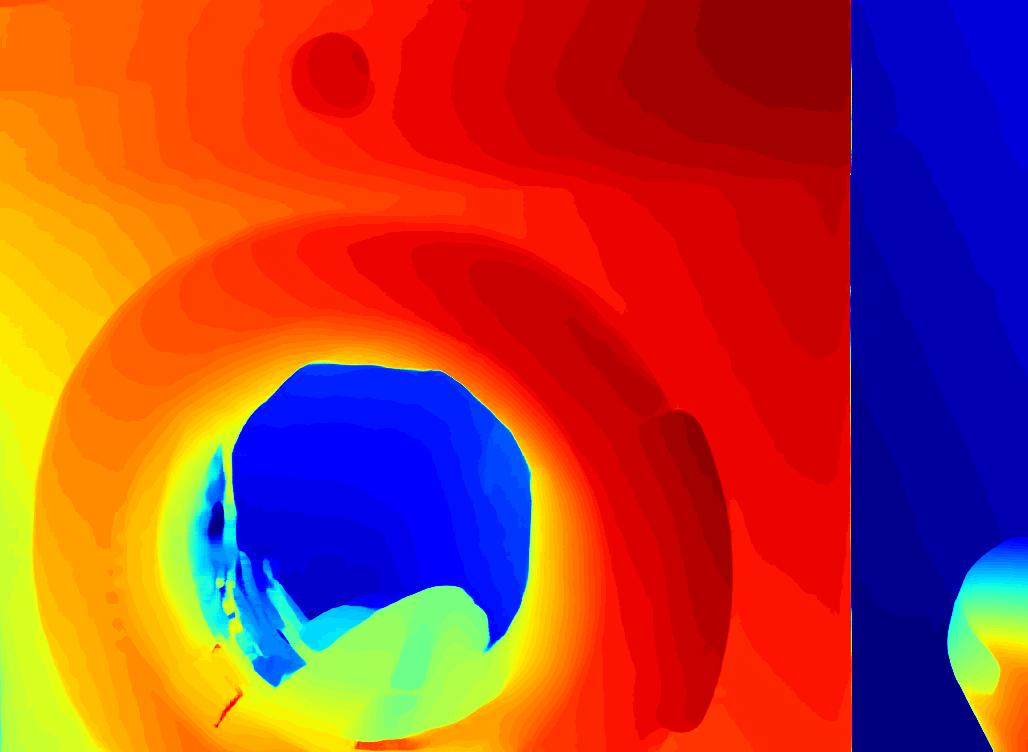} & 
        \includegraphics[width=0.1\textwidth]{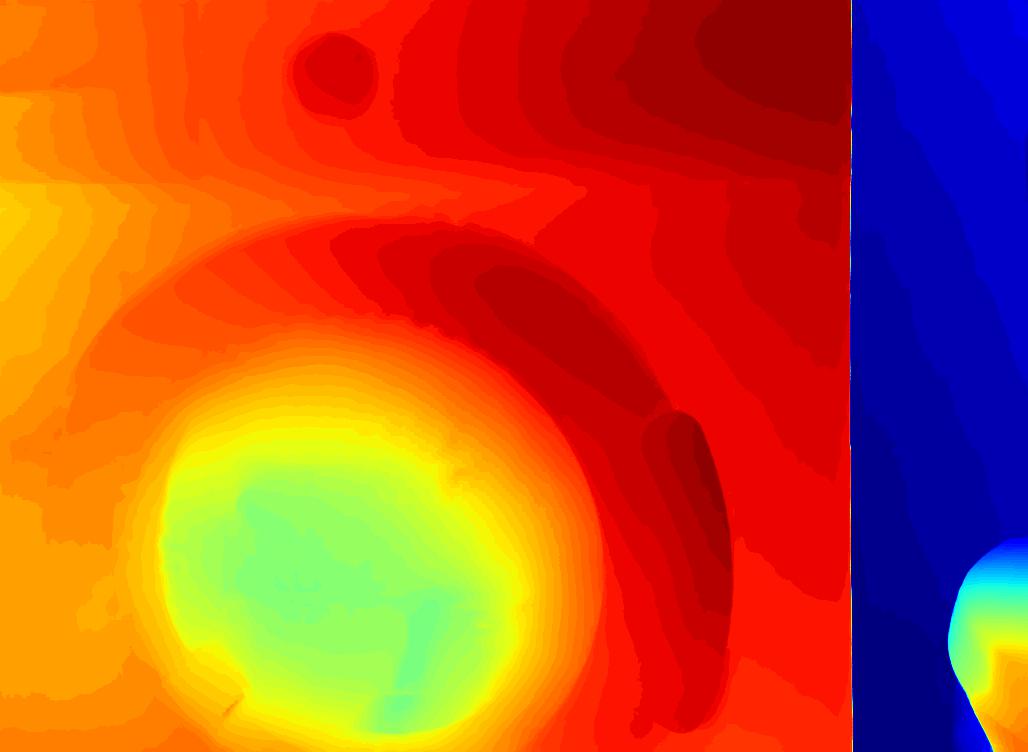} & 
        \includegraphics[width=0.1\textwidth]{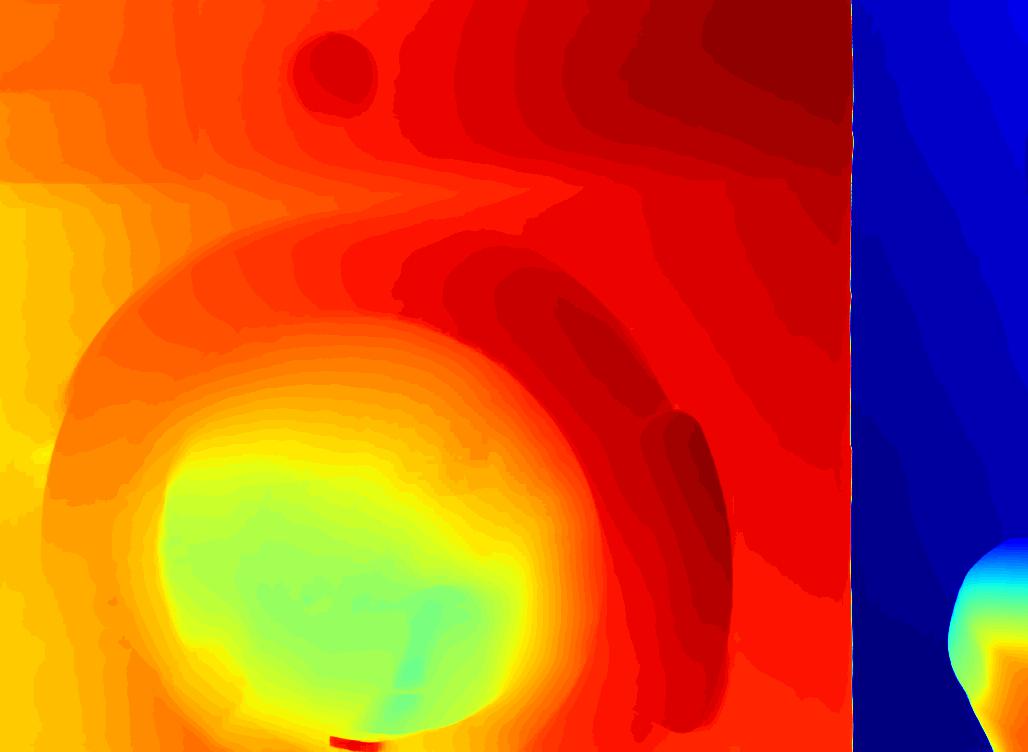} & 
        \includegraphics[width=0.1\textwidth]{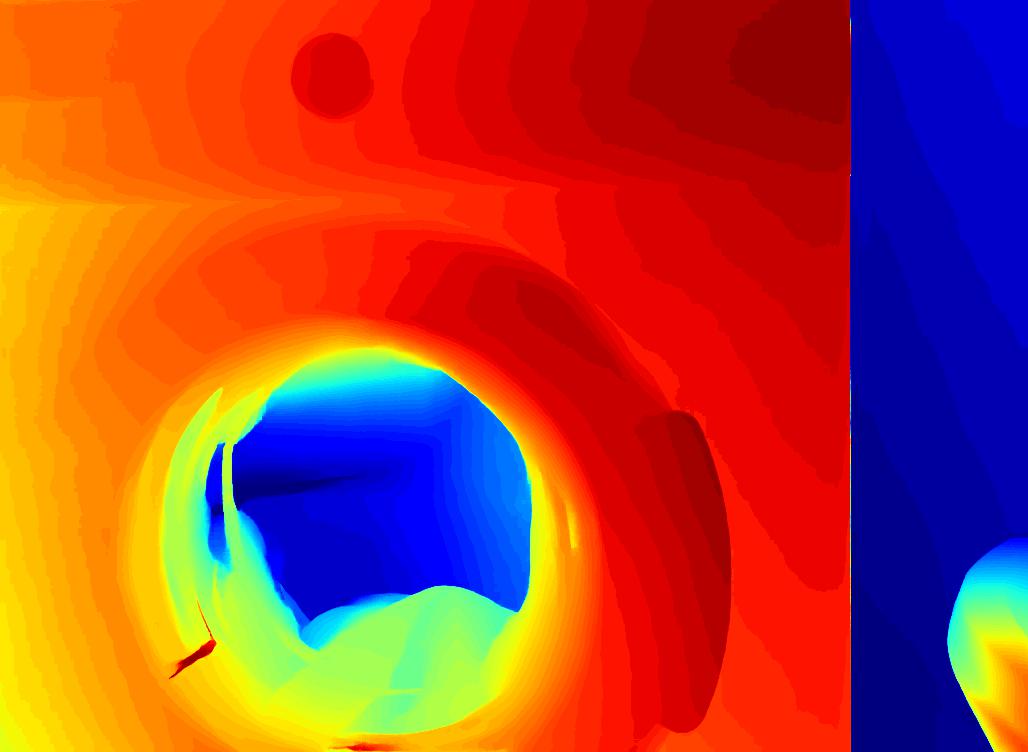} & 
        \includegraphics[width=0.1\textwidth]{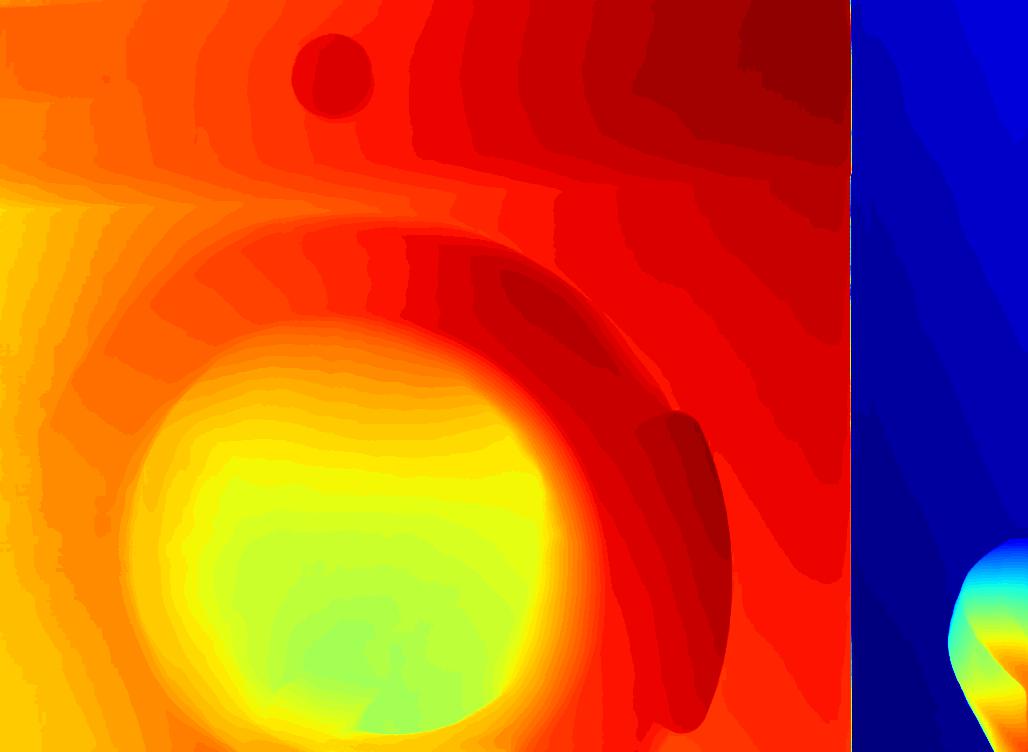} & 
        \includegraphics[width=0.1\textwidth]{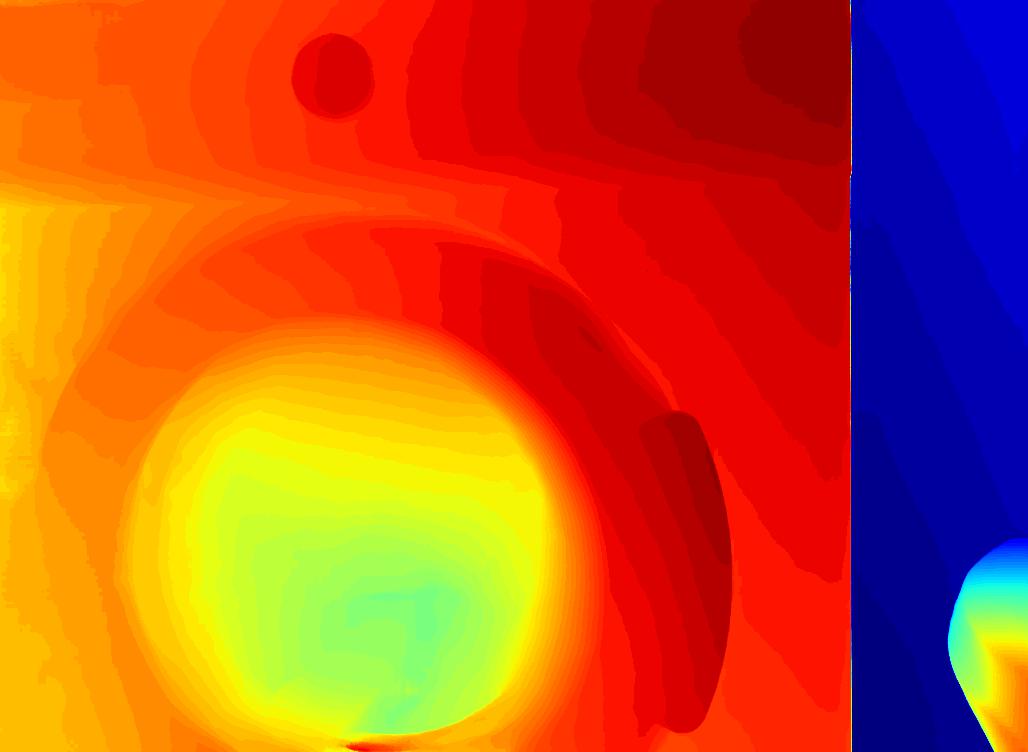} \\
        
        \includegraphics[width=0.1\textwidth]{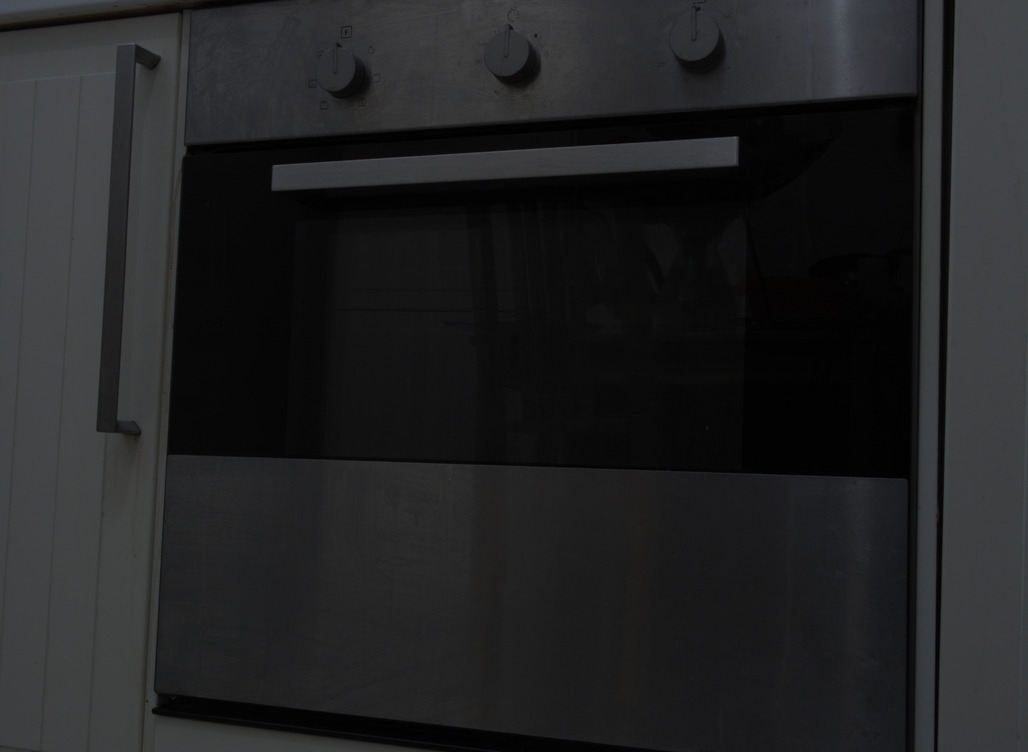} & 
        \includegraphics[width=0.1\textwidth]{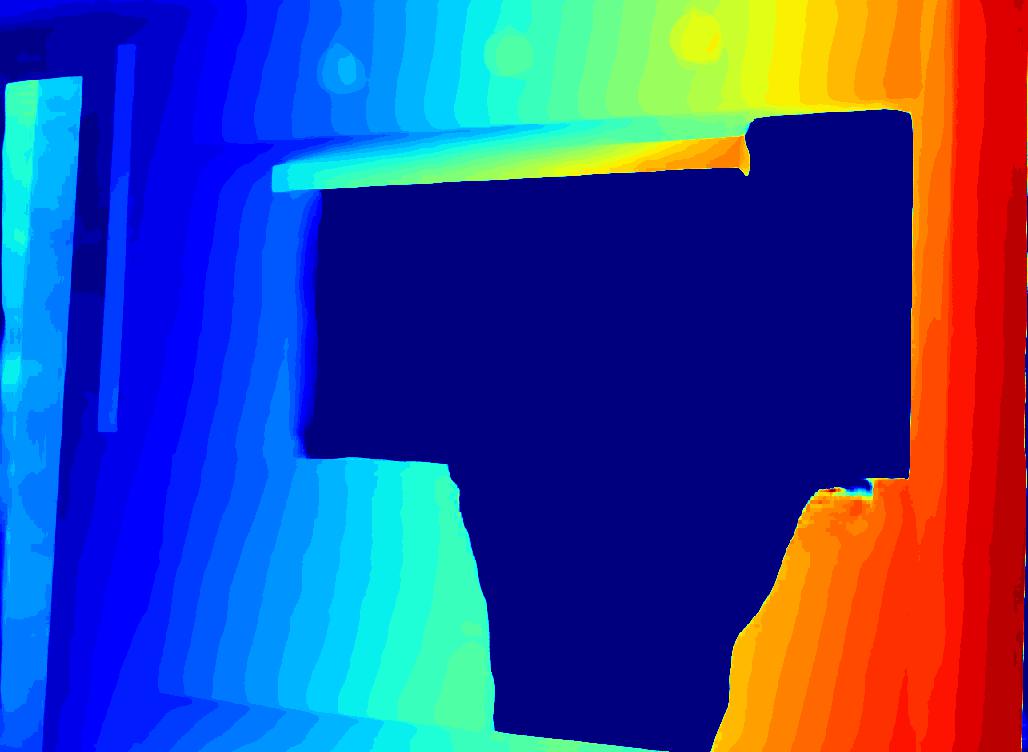} &
        \includegraphics[width=0.1\textwidth]{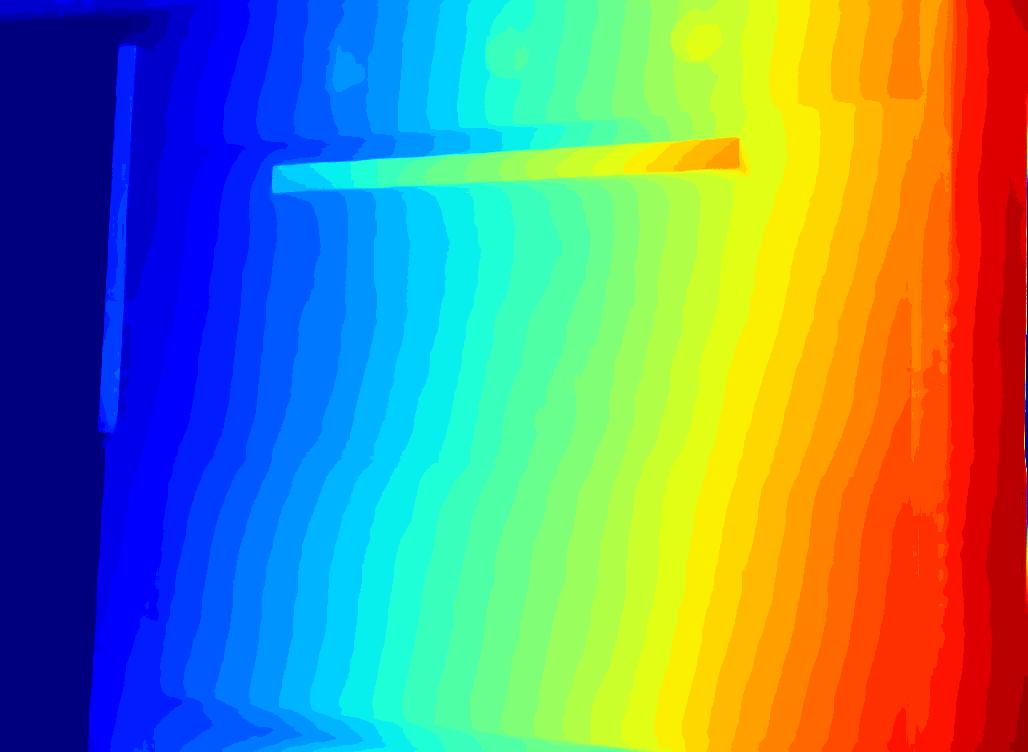} &
        \includegraphics[width=0.1\textwidth]{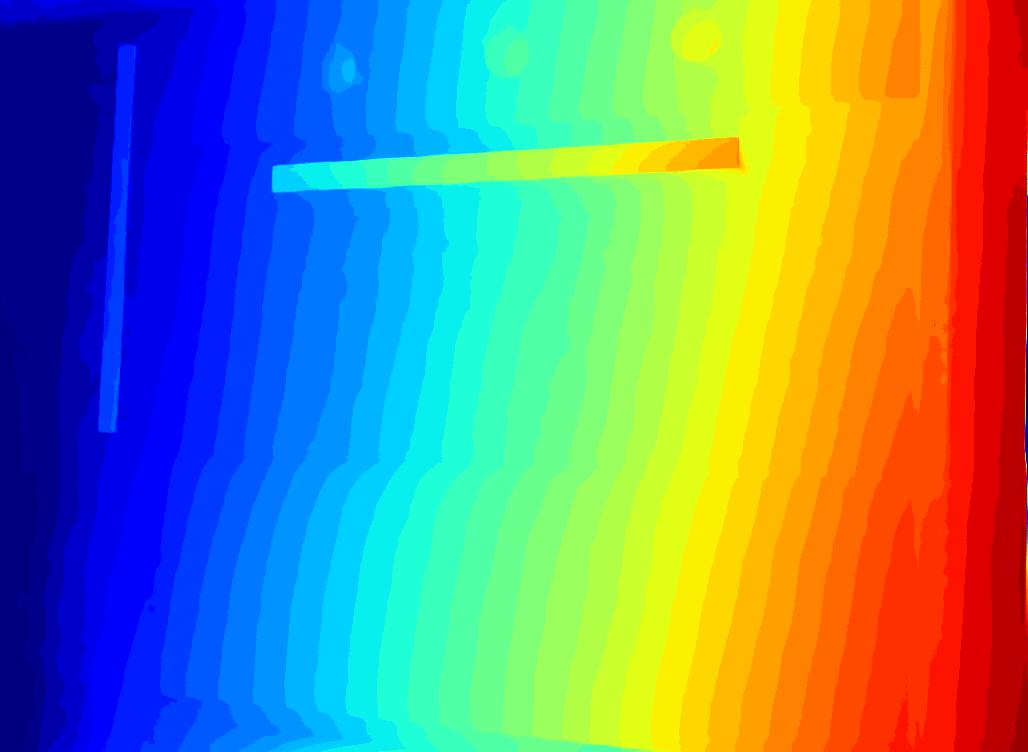} & 
        \includegraphics[width=0.1\textwidth]{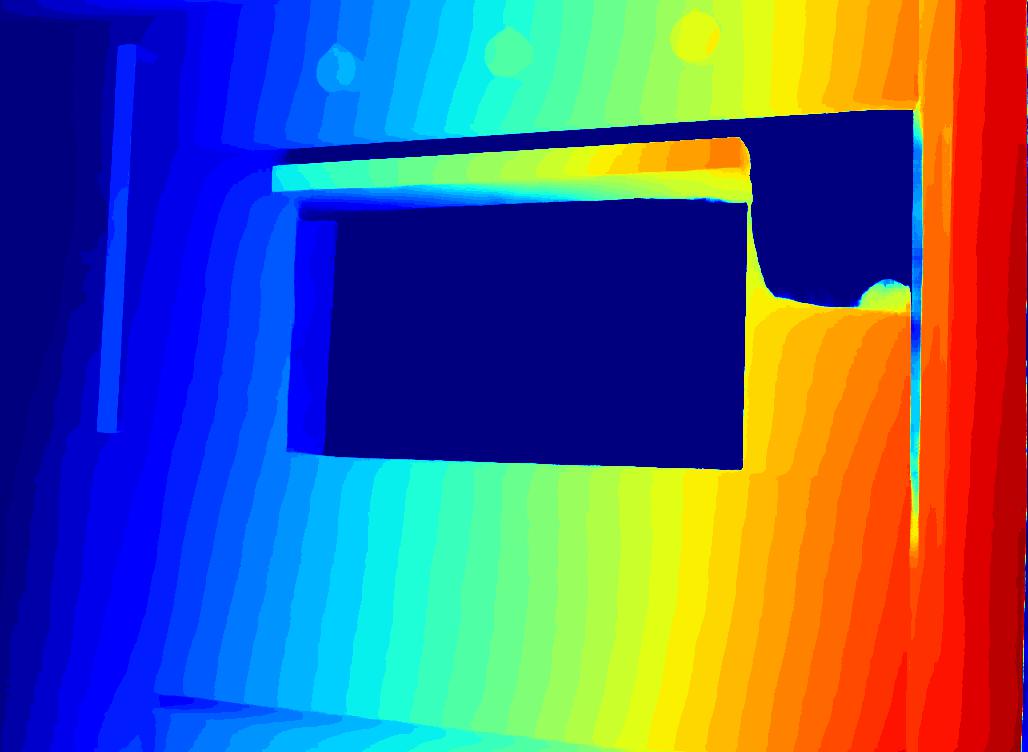} &
        \includegraphics[width=0.1\textwidth]{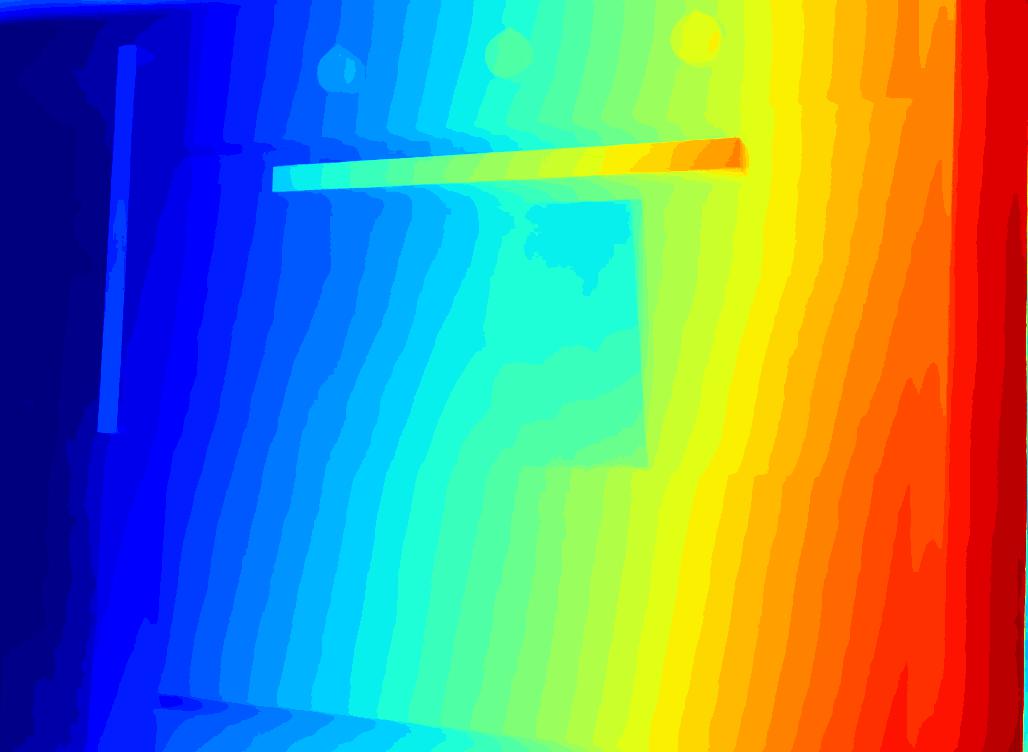} &
        \includegraphics[width=0.1\textwidth]{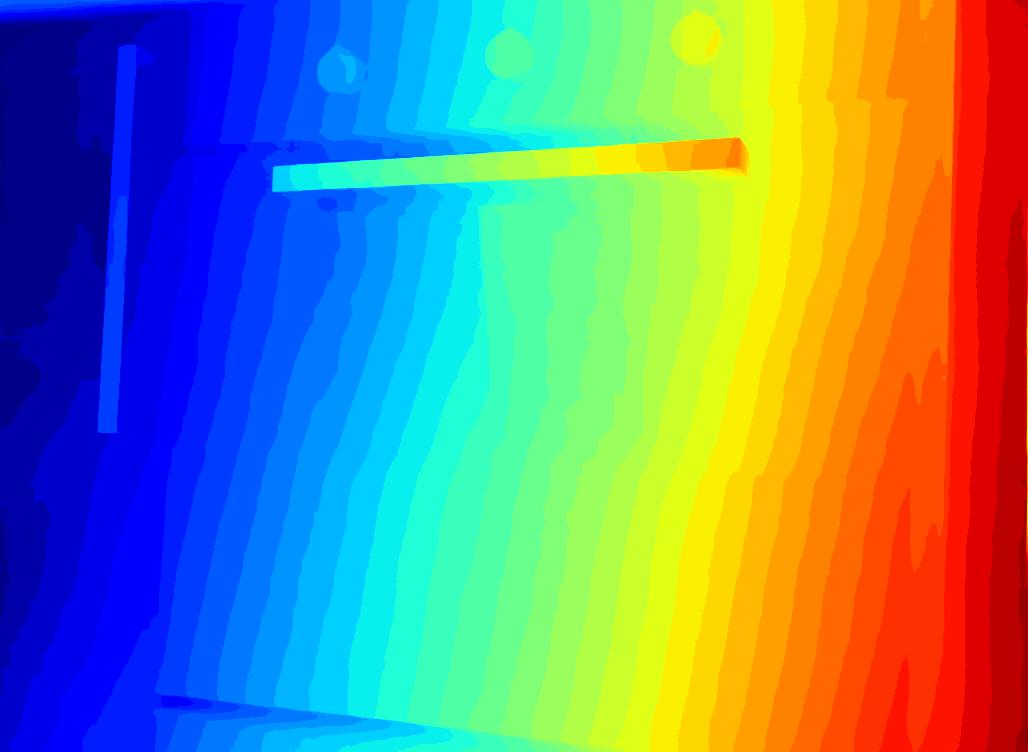} \\
    \end{tabular}
    \caption{\textbf{Qualitative post fine-tuning results.} Examples of predictions by MiDaS and DPT (top), RAFT-Stereo and CREStereo (bottom). For each model, we show results achieved by the original model and by fine-tuned instances using proxy or GT segmentation masks.}
    \label{fig:qual_stereo}
\end{figure*}

\subsection{Stereo Depth Estimation}


\textbf{Virtual Disparity Generation Alternatives.} We inquire about two main alternatives to generate virtual disparities: i) \textit{Virtual Disparity}: masking both left and right images according to material segmentation masks -- as materials annotations are provided for the left image only, we warp it over the right image according to ground-truth disparity -- and then processing the two with the stereo network we are going to fine-tune similar to Monocular networks, ii) \textit{Merged}: merging disparity labels produced by the stereo model itself with those generated by original DPT weights \cite{Ranftl2021}, as detailed in Eq. \ref{eq:merging}.
Although the former might appear as the natural extension of our proposal from the monocular to the stereo case, we will demonstrate its ineffectiveness.

\textbf{Fine-tuning Results (GT Segmentation).} Tab. \ref{tab:stereo} collects the results obtained by fine-tuning RAFT-Stereo and CREStereo through our technique. From top to bottom, we report the results achieved by the original models (\textit{Base}) as well as the instances fine-tuned on their own predictions (\textit{Ft. Base}) or on pseudo labels obtained according to the two strategies (\textit{Ft. Virtual Depth}, \textit{Ft. Merged}).

Not surprisingly, fine-tuning the networks on their own predictions is harmful (RAFT-Stereo) or scarcely effective (CREStereo). Applying the first of the two strategies sketched before yields just a negligible improvement over the original models on ToM classes. 
This evidence confirms that our pipeline designed for the monocular case cannot na\"ively be extended to the stereo case by in-painting the two images since masking ToM objects with constant colors does not ease matching -- on the contrary, it introduces textureless regions, which are  likely to be labeled as planar surfaces by stereo models. 
Conversely,  the second strategy consistently improves the predictions with both RAFT-Stereo and CREStereo. In particular, the former achieves 9.23, 13.83, 15.62, and 16.69\% absolute reductions on bad-2, bad-4, bad-6, and bad-8, respectively, as well 7.62 and 9.13 reductions on MAE and RMSE on ToM regions. CREStereo obtains 14.93, 17.13, 17.48, and 17.54\% on bad metrics, and 7.40 and 8.91 reductions on MAE and RMSE. Moreover, the accuracy over \textit{Other} pixels is also improved, although with minor margins.
Fig. \ref{fig:proxy_merge} provides a qualitative comparison between the labels obtained by the two strategies. The former produces a planar surface for the mirror completely misaligned with respect to the wall, whereas the latter combines the virtual depth labels from DPT on masked images with disparity labels at best.

\textbf{Fine-tuning Results (Proxy Segmentation).} Finally, we replace the manually annotated segmentation masks with those predicted by Trans2Seg and MirrorNet and then distill virtual disparities for fine-tuning both stereo networks. As pointed out before, both Trans2Seg and MirrorNet have not been trained on Booster. Thus, \textit{Merging} produces significant differences with respect to the use of manually annotated masks, as shown in Fig. \ref{fig:proxy_stereo}. Nevertheless, 
Table \ref{tab:stereo_proxy} shows that our pipeline improves the performance of both RAFT-Stereo and CREStereo on ToM objects, even in the case of extremely noisy proxy semantic annotations.
More precisely, CREStereo seems to benefit more from the Proxy segmentation configuration than RAFT-Stereo. Indeed, on the one hand, we can notice how RAFT-Stereo improves on ToM regions at the expense of accuracy on other pixels when using Proxy segmentations. This yields, on All pixels, an increase in the bad-2 and bad-4 error rates, whereas bad-6, bad-8, MAE, and RMSE remain lower.
On the other hand, CREStereo seems capable of exploiting fine-tuning much better, yielding more accurate results on any metric with both Proxy or GT masks.
This outcome proves that  our pipeline is effective for fine-tuning stereo models even without manually annotated masks. Nonetheless, segmenting images through human labeling unleashes its full potential, whose cost is much lower compared to that that would be required to annotate depths.

\subsection{Qualitative Results} 

To conclude, Fig. \ref{fig:qual_stereo} shows the effect of the fine-tuning carried out according to our proposal, with two examples for monocular (top) and stereo (bottom) networks from the Booster train and test sets respectively.
We highlight how MiDaS, DPT, RAFT-Stereo, and CREStereo learn to deal with ToM surfaces either when relying on proxy segmentation masks provided by neural networks or accurately annotated by humans. More qualitatives in the supplement.


\section{Conclusion}
We have proposed an effective methodology for training depth estimation networks to deal with transparent and mirror surfaces. By in-painting these surfaces on RGB images, we can quickly annotate a dataset with virtual depth labels, that can be used to fine-tune both monocular and stereo networks, with outstanding results.
A promising future direction would be to extend our technique to instance segmentation masks to get better virtual depth maps in the presence of multiple ToM objects in the same scene.


{\small
\bibliographystyle{ieee_fullname}
\bibliography{egbib}
}

\newpage\phantom{Supplementary}
\multido{\i=1+1}{9}{\includepdf[pages={\i}]{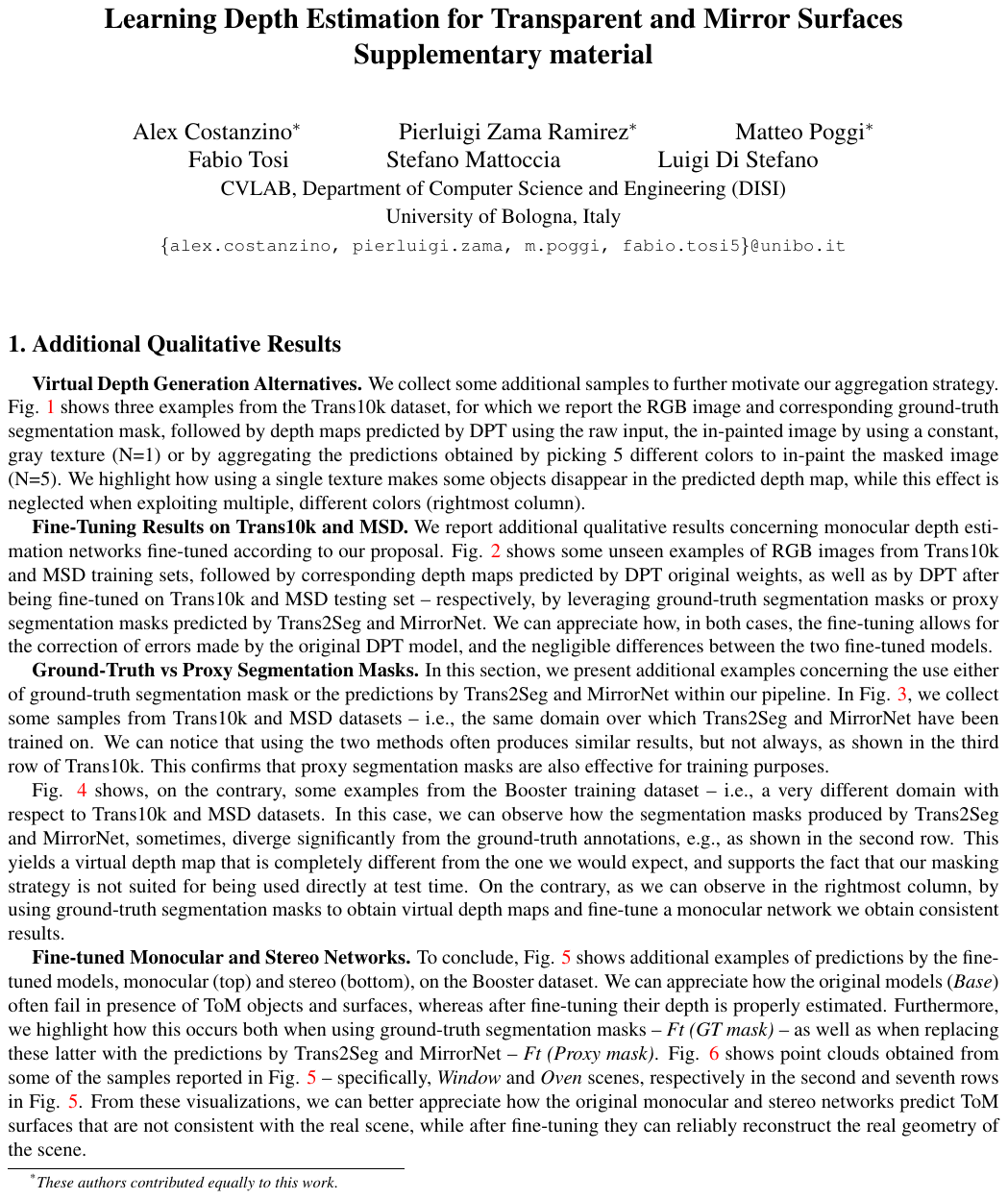}}

\end{document}